\documentclass{article}

 \usepackage[main, preprint]{neurips_2026}

\usepackage[utf8]{inputenc}
\usepackage[T1]{fontenc}
\usepackage{hyperref}
\usepackage{url}
\usepackage{booktabs}
\usepackage{amsmath,amssymb,amsfonts}
\usepackage{nicefrac}
\usepackage{microtype}
\usepackage{xcolor}
\usepackage{wrapfig}
\usepackage{enumitem}
\setlist{leftmargin=*}
\usepackage{makecell}
\usepackage{pifont}
\newcommand{\cmark}{\ding{51}}
\newcommand{\xmark}{\ding{55}}
\DeclareMathOperator{\Span}{span}
\DeclareMathOperator{\Cov}{Cov}

\newcommand\calM{{\mathcal{M}}}
\newcommand\calN{{\mathcal{N}}}

\newcommand\ve{{\boldsymbol{e}}}

\newcommand\vr{{\boldsymbol{r}}}

\newcommand\vu{{\boldsymbol{u}}}
\newcommand\vv{{\boldsymbol{v}}}
\newcommand\vw{{\boldsymbol{w}}}
\newcommand\vx{{\boldsymbol{x}}}

\newcommand{\inner}[2]{\langle #1,#2\rangle}
\newcommand{\thoinact}{\boldsymbol{\theta}^{\mathrm{inact}}}
\newcommand{\thbar}{\bar{\theta}^{\parallel}}

\newcommand\sR{{\mathbb{R}}}

\newcommand{\myparagraph}[1]{\noindent\textbf{#1.}}

\newcommand{\RR}{\mathbb{R}}

\newtheorem{theorem}{Theorem}

\newcommand{\Loss}{\mathcal{L}}
\newcommand{\bth}{\boldsymbol{\theta}}
\newcommand{\thp}{\theta^\parallel}
\newcommand{\tho}{\boldsymbol{\theta}^\perp}

\newtheorem{lem}{Lemma}
\newtheorem{cor}{Corollary}[section]
\newtheorem{prop}{Proposition}

\newtheorem{asmp}{Assumption}
\newtheorem{defn}{Definition}[section]

\newtheorem{rem}{Remark}

\usepackage[most]{tcolorbox}
 \newtcolorbox{valleybox}[1][]{%
  enhanced,
   colback=blue!3,
  colframe=blue!40!black,
   coltitle=white,
   fonttitle=\bfseries,
  left=4pt, right=4pt, top=4pt, bottom=4pt,
   arc=2pt,
   boxrule=0.8pt,
   title={Assumption A\; (Valley Geometry)},
   #1
 }

 \newtcolorbox{decaybox}[1][]{%
   enhanced,
   colback=orange!3,
   colframe=orange!50!black,
   coltitle=white,
   fonttitle=\bfseries,
   left=6pt, right=6pt, top=6pt, bottom=6pt,
  arc=2pt,
   boxrule=0.8pt,
   title={Assumption B\; (Decay Profile)},
   #1
 }

\definecolor{theoremblue}{RGB}{235,243,252}
\definecolor{theoremborder}{RGB}{70,110,160}

\newtcolorbox{mainthmbox}{
  breakable,
  colback=blue!3,
  colframe=blue!3,
  boxrule=0pt,
  arc=1mm,
  left=1.2mm,
  right=1.2mm,
  top=1.2mm,
  bottom=1.2mm,
  before skip=6pt,
  after skip=6pt,
  width=\linewidth,
  enlarge left by=0mm,
  enlarge right by=0mm
}

\title{Classification with Large-Step Gradient Descent: Unifying Self-Stabilization and Convergence Rates}

\author{
  \begin{minipage}[t]{0.96\textwidth}
  \centering
  \normalfont
  \textbf{Salma Tarmoun}$^{1,4}$ \hspace{1em}
  \textbf{Lachlan Ewen MacDonald}$^{3,4}$ \hspace{1em}
  \textbf{Konstantinos Emmanouilidis}$^{2,4}$ \\[0.15em]
  \textbf{Ziqing Xu}$^{4,6}$ \hspace{1em}
  \textbf{Ren\'e Vidal}$^{2,3,4,5,6}$ \\[0.35em]
  {\small
  $^{1}$AMCS \quad
  $^{2}$CIS \quad
  $^{3}$ESE \quad
  $^{4}$IDEAS \quad
  $^{5}$Radiology \quad
  $^{6}$Statistics and Data Science \\
  University of Pennsylvania, Philadelphia, PA 19104 \\
  \texttt{starmoun@sas.upenn.edu}
  }
  \end{minipage}
}

\begin{document}

\maketitle

\begin{abstract}
  Neural networks are often trained with gradient descent (GD) using step sizes that exceed classical stability thresholds, a regime where standard convergence theory no longer applies. Despite exhibiting non-monotonic loss and oscillatory dynamics, this regime frequently leads to faster convergence and better generalization. Existing theoretical perspectives offer complementary but incomplete insights: one identifies a \textit{self-stabilization} mechanism that drives GD toward flat regions but does not yield convergence rates; another derives rigorous rates for specific losses but does not explain the underlying stabilization mechanism; and a third unifies these views for overparameterized least squares, but relies on the existence of a minimizer manifold, which is absent in classification settings. In this work, we extend this unified geometric perspective to exponential-tail losses, including logistic and cross-entropy losses in separable linear classification, where the loss infimum is not attained at any finite parameter. We identify a curvature-defined reference subspace that replaces the role of the minimizer manifold, yielding a coupled dynamical system in which orthogonal oscillations are damped by parallel progress along directions of decreasing sharpness. Under regularity conditions, which we verify for logistic and multiclass cross-entropy losses on symmetric data models, we derive explicit bounds for the iterates, not just the loss, covering both the unstable and stable regimes, and revealing a three-phase damping structure in the latter. Our analysis further uncovers a transient implicit bias induced by large step sizes: although GD converges asymptotically to the max-margin direction, large steps can leave persistent orthogonal residuals and, in certain geometries, amplify them, affecting finite-time alignment with the max-margin direction.
\end{abstract}

\section{Introduction}
Modern deep learning owes much of its success to optimization heuristics refined through extensive experimentation \cite{baLayerNormalization2016, ioffeBatchNormalizationAccelerating2015a, loshchilovDecoupledWeightDecay2019a}. Notably, state-of-the-art results often depend on surprisingly simple algorithms such as gradient descent (GD), whose practical effectiveness persistently outpaces our theoretical understanding. One particularly intriguing illustration of this gap is the common use of large learning rates. Classical optimization theory traditionally advocates for small learning rates to ensure stable monotonic convergence \cite{nesterov2013introductory, boyd2004convex}. Yet, practitioners routinely exceed these conservative bounds, achieving faster convergence and often better generalization~\cite{smithSuperConvergenceVeryFast2018, goyal2017accurate, sadrtdinov2024large, leclerc2020two, lewkowycz2020large}. This counterintuitive regime, known as the edge of stability (EoS), is characterized by non-monotonic loss and oscillatory dynamics \cite{cohen2021gradient, wu2018sgd}. The EoS can arise either from initializing training with a large learning rate, or through progressive sharpening, whereby the local curvature grows along the training trajectory until it reaches the stability boundary \cite{wu2018sgd,macdonald_progressive_2023, wang2022analyzing}.

These empirical successes and observations raise fundamental theoretical questions:
\begin{center}
    \textit{Why does operating near instability lead to faster convergence? Can we provide rigorous theoretical guarantees for convergence under these oscillatory dynamics?}
\end{center}

Several theoretical perspectives have emerged in recent years \cite{zhu2023understanding, liu2025minimalist, wang2023good, kreisler2023gradient, chen2024stability, even2023s, arora2022understanding, song2023trajectory, chen2023beyond, lee2023new, ghosh_learning_2025}. Among the most insightful is the \emph{self-stabilization} mechanism of \cite{damianself}, which shows that GD naturally adjusts to local curvature in response to large-step-size instabilities, preventing divergence. However, their assumptions are difficult to verify analytically and have only been confirmed numerically on specific architectures. Moreover, this framework does not establish convergence rates. Wu et al.~\cite{wu2024large, wu2025benefits} take a complementary approach: they derive convergence rates for logistic-type losses through a direct analysis of the gradient structure. However, their arguments do not reveal the mechanistic role of self-stabilization and do not extend to other loss families without a new specialized analysis for each. MacDonald et al.~\cite{macdonald2025} unify both perspectives for overparameterized least squares via a Taylor expansion around a manifold of minimizers, establishing a parallel-orthogonal decomposition that makes the self-stabilization mechanism explicit and simultaneously yields convergence rates. However, this approach fundamentally relies on the existence of a \emph{minimizer manifold}, which is absent in the classification losses we target.

In this work, we extend this unified approach to exponential-tail losses, a class that includes logistic and cross-entropy losses. The high-level takeaway is:

\begin{tcolorbox}[boxrule=0.5pt, boxsep=1pt, left=2pt, right=2pt]
\centering
We develop a unified framework for gradient descent on exponential-tail losses that makes self-stabilization quantitatively explicit and yields convergence rates for the iterates, while revealing a three-phase damping mechanism and a geometry-dependent transient
implicit bias.
\end{tcolorbox}

Specifically, we establish the following:

\begin{enumerate}[leftmargin=*]
\item We identify a \emph{subspace} defined by the curvature structure of the loss as the geometric anchor for analyzing GD in the exponential-tail setting, and show that decomposing the dynamics relative to it yields a two-equation system: one governs the oscillations orthogonal to the subspace, and the other captures parallel drift along the subspace, where a third-order coupling acts as implicit gradient descent on the curvature, a key feature of the self-stabilization mechanism.

\item Under certain regularity conditions, which we verify for logistic and cross-entropy under symmetric separable data, we provide a fine-grained convergence analysis with explicit rates for the iterates, not just the loss, revealing a three-phase damping mechanism for the oscillations.

\item We identify a transient implicit bias of large step sizes: certain orthogonal components can be amplified during the unstable phase and persist at any finite horizon despite asymptotic directional  convergence to the max-margin solution.

\end{enumerate}

The remainder of the paper is structured as follows: we begin in Section~\ref{sec:related_work} with a review of the related literature. Sections~\ref{sec:setup}--\ref{sec:general_conv} develop the general framework; Section~\ref{sec:applications} verifies it for logistic regression and multiclass classification with the cross-entropy loss; Section~\ref{sec:experiments} empirically validates the theory.

\myparagraph{Notation} Throughout, scalars are denoted by lowercase letters (e.g., \(\eta,\lambda\)) and vectors by lowercase bold letters (e.g., \(\mathbf{x},\mathbf{u}\)). We use \(\|\cdot\|\) for the Euclidean norm and \(\langle\cdot,\cdot\rangle\) for the standard Euclidean inner product. For a subspace \(S\subseteq\mathbb R^d\), \(S^\perp\) denotes its orthogonal complement, \(\Pi_S\) the orthogonal projector onto \(S\), and \(I_S\) the identity operator on \(S\). We write \(f\asymp g\) on a set \(S\) if there exist constants \(c,C>0\) such that \(c\,g(x)\le f(x)\le C\,g(x)\) for all \(x\in S\). We use \(f\lesssim g\) and \(f\gtrsim g\) for the corresponding one-sided inequalities up to constants.

\section{Related Work}\label{sec:related_work}

The EoS phenomenon has motivated a rapidly growing literature on large-step-size training, progressive sharpening, benign oscillations, catapult dynamics, and step-size-dependent implicit bias \cite{wu2018sgd,jastrzkebskirelation,lewkowycz2020large, jastrzebskibreak,xing2018walk,li2019towards,leclerc2020two, cohen2021gradient,macdonald_progressive_2023,wang2022analyzing, nacson2022implicit,even2023s,wang2023good,cai2024large, chen2023beyond,song2023trajectory,zhu2023understanding, kreisler2023gradient,liu2025minimalist,lu_benign_2023, chen2024stability,ghosh_learning_2025,cruaciun2024convergence}. We defer a broader discussion of this literature to Appendix~\ref{app:related_work} and focus here on the works closest to ours:

\begin{table}[t]
\centering
\caption{High-level comparison of related theoretical analyses of GD at the edge of stability.}
\label{tab:comparison}
\small
\begin{tabular}{lcccc}
\toprule
 & \textbf{\makecell{Self-stabilization\\mechanism?}} & \textbf{Rates} & \textbf{Loss setting} & \textbf{\makecell{Reference Geometry}} \\
\midrule
Damian et al.~\cite{damianself}
 & \cmark & \xmark & Abstract conditions$^\dagger$ & Stable Set \\
Wu et al.~\cite{wu2024large, wu2025benefits}
 & \xmark & \cmark & Logistic-like & Max-margin line \\
MacDonald et al.~\cite{macdonald2025}
 & \cmark & \cmark & Least squares & Minimizer manifold \\
\textbf{This work}
 & \cmark & \cmark & Exponential-tail & Curvature-defined subspace \\
\bottomrule
\end{tabular}

\begin{flushleft}
\footnotesize
$^\dagger$ Verified numerically but not analytically for any specific loss.
\end{flushleft}
\vspace{-3mm}
\end{table}

\myparagraph{General self-stabilization mechanisms}
This line of work seeks architecture-independent mechanisms explaining why GD remains stable when the local quadratic model predicts instability. The most influential is the self-stabilization framework of \cite{damianself}. The key analytical tool is a cubic Taylor expansion of the loss, centered at a constrained set defined by the sharpness threshold $2/\eta$, which shows that the third-order term induces a negative feedback on the sharpness, causing GD to behave like projected gradient descent under $\lambda_{\max}(\bth) \le 2/\eta$. While this provides a compelling explanation of EoS, it is not designed to yield end-to-end convergence rates and its hypotheses involve local, trajectory-dependent quantities that are difficult to verify analytically.

\myparagraph{Quantitative convergence analyses for exponential-tail losses} Wu et al.~\cite{wu2023implicit,wu2024large,wu2025benefits} show that, for logistic regression and related classification losses, large-step GD can preserve the max-margin directional bias while achieving accelerated loss decay, with extensions to regularized objectives, NTK-regime models, and online SGD. Crawshaw and Liu~\cite{crawshaw2026} sharpen this in two dimensions, giving tight unstable-to-stable transition-time estimates via a fine-grained analysis of orthogonal oscillations.
Our work takes a different perspective: rather than deriving rates from margin-based recursions, we connect convergence to the underlying self-stabilization mechanism. This connection enables a fine-grained analysis of the dynamics at the level of the iterates, not just the loss, while simultaneously providing a mechanistic interpretation of convergence. In particular, it reveals a new phenomenon: when the support vectors do not span the full data subspace, the dynamics can exhibit a transient implicit bias, with persistent, and in some cases amplified, orthogonal components.

\myparagraph{Relationship to \cite{macdonald2025}} Our work is closest in spirit to MacDonald et al.~\cite{macdonald2025}, who bridge the gap between mechanistic and quantitative approaches for overparameterized least squares. Using a Taylor expansion of the gradient around a manifold of minimizers, they establish a parallel-orthogonal decomposition that makes the self-stabilization mechanism transparent while simultaneously yielding convergence rates.
We extend this framework in two key directions. First, we replace the minimizer manifold with a curvature-defined reference subspace, allowing the analysis to apply in settings where no finite minimizer exists. This shift moves the theory beyond the overparameterized least-squares regime to exponential-tail losses such as logistic and cross-entropy. Second, our assumptions are formulated at the level of structural properties of the loss and its geometry, rather than tied to a specific objective, thereby enabling a unified treatment of a broader class of problems.

\section{Setup and Motivation}\label{sec:setup}

We study gradient descent $\bth_{t+1} = \bth_t - \eta\,\nabla\Loss(\bth_t)$ on a smooth loss $\Loss:\sR^d \to \sR$ with step size $\eta > 0$. Two classical facts motivate the regime of interest. First, if $\nabla\Loss$ is $L$-Lipschitz on a convex set containing $\bth_t$ and $\bth_{t+1}$, the descent lemma gives
\[
  \Loss(\bth_{t+1}) \le \Loss(\bth_t) - \eta\Bigl(1 - \tfrac{\eta L}{2}\Bigr)\|\nabla\Loss(\bth_t)\|^2,
\]
so the loss decreases for $0<\eta<2/L$ whenever the gradient is nonzero \cite{nesterov2013introductory}. Second, the Hessian governs the linearized GD map, whose Jacobian is $I - \eta\,\nabla^2\Loss(\bth)$: a positive-curvature eigenmode with eigenvalue $\lambda$ has multiplier $1-\eta\lambda$ and is expansive when $\eta\lambda > 2$. The largest eigenvalue $\lambda_{\max}(\nabla^2\Loss(\bth))$, the \textbf{sharpness} at $\bth$, therefore sets the local stability threshold $2/\eta$. Under large step sizes, such as in the EoS regime, the sharpness exceeds $2/\eta$: the top modes are locally expansive and the loss oscillates non-monotonically \cite{cohen2021gradient}. Our analysis establishes convergence in exactly this setting, where classical theory provides no convergence guarantee \cite{cohen2021gradient}. Yet, the losses we study share a geometric structure that makes this possible: the parameter space decomposes into a low-curvature \emph{parallel} direction along which the loss decreases, and high-curvature orthogonal directions along which the iterates oscillate. The interplay between these components produces a coupled dynamical system, whose structure we develop in this section and use to prove convergence guarantees in Section~\ref{sec:general_conv}.

\subsection{Warmup Example}\label{subsec:warmup_example}

\begin{figure}
    \centering
    \includegraphics[width=0.9\textwidth]{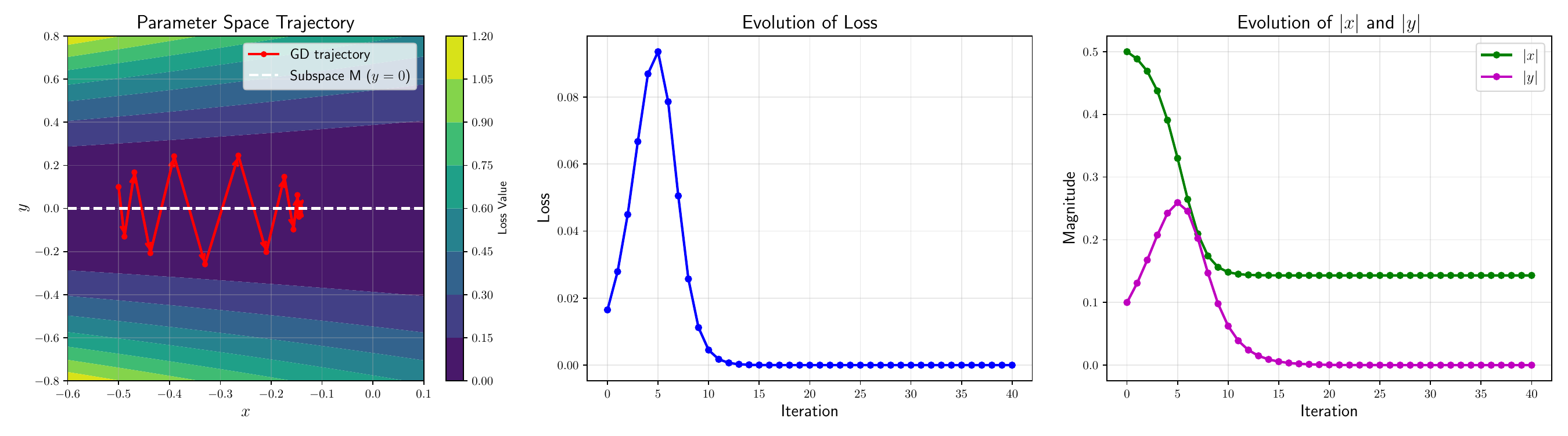}
    \caption{GD dynamics illustrating the flat subspace $\calM$, orthogonal oscillations, evolution of the loss, and parameter magnitudes.}
    \label{fig:gd_oscillations}
    \vspace{-2mm}
\end{figure}

Consider the 2D loss function: $\Loss(x, y) = e^{-x}y^2.$ The gradient and Hessian are given by:
\begin{align}
    \nabla \Loss(x, y) = e^{-x}\begin{pmatrix}
        -y^2\\ 2 y
    \end{pmatrix}, \quad \nabla^2 \Loss (x, y) = e^{-x}\begin{pmatrix}
        y^2 & -2 y \\  -2 y & 2
    \end{pmatrix}.
\end{align}

When $y = 0$, the curvature vanishes in the $x$ direction and the loss is identically zero along $x$. This suggests a natural decomposition of the
parameter space
into a low-curvature subspace $\mathcal{M}$, the $x$-axis, and its orthogonal complement, the $y$-axis. Using this decomposition, we analyze GD with learning~rate~$\eta$ and initialization $y_0 \neq 0$:
\begin{align}
x_{t+1} = x_{t} + \eta e^{-x_{t}} y_{t}^2, \qquad y_{t+1} = y_{t} (1 - 2\eta e^{-x_{t}}).
\end{align}

These coupled dynamics reveal the self-stabilization phenomenon. Define the \emph{stability threshold} $x_* := \log(\eta)$, so that
$2\eta e^{-x_*} = 2$ and let $x_0 < x_*$. The three regimes are then:
\begin{itemize}[leftmargin=*]
 \item \textbf{Unstable phase} ($2\eta e^{-x_t} > 2$, i.e.\ $x_t < x_*$). Since $|1 - 2\eta e^{-x_t}| > 1$, $|y_t|$ grows  while alternating sign at each step. However, this growth is self-limiting: the first equation can be interpreted as gradient descent on $e^{-x_t}$ with  $\eta y_t^2$ acting as an effective step size. A large $|y_t|$ therefore accelerates forward movement in the $x$ direction, which simultaneously decreases $e^{-x_t}$ and weakens the instability.

\item \textbf{Transition to stability} ($x_t$ crosses $x_*$). Once $x_t \ge x_*$, we have $2\eta e^{-x_t} \le 2$ and hence $|1 - 2\eta e^{-x_t}| \le 1$: the $y$-update becomes non-expanding. For $1 < 2\eta e^{-x_t} < 2$, the iterates still oscillate in sign but now contract. This stage can be described as an \emph{oscillatory stable regime}.

\item \textbf{Non-oscillatory contraction} ($2\eta e^{-x_t} < 1$, i.e.\ $x_t > x_* + \log 2$).  The multiplier $1 - 2\eta e^{-x_t} \in (0,1)$, so $y_t$ contracts without sign changes.
\end{itemize}

This illustrates the essential components of the unstable regime: oscillation in sharp directions, slow drift in flat directions, and an emergent coupling that stabilizes the dynamics. The subspace $\calM$ ($x$-axis), plays a central role in separating these behaviors.
Two structural features of this example carry over to the general analysis. First, the orthogonal update $y_{t+1} = (1 - 2\eta\, e^{-x_t})\, y_t$ is an \emph{exact multiplicative map}: the contraction/expansion factor depends on $x_t$ but not on $y_t$. Second, the sharpness on $\calM$ decays \emph{exponentially} along the parallel direction, converting the qualitative observation \emph{sharpness drops below $2/\eta$} into quantitative rates. One feature does \emph{not} carry over: the loss restricted to the parallel direction $\Loss(x, 0)$ vanishes identically, so forward progress comes  entirely from the curvature-reduction coupling. In the classification losses we analyze, the parallel gradient is nonvanishing and provides additional forward drift; the curvature-reduction term acts as an accelerant during the unstable phase, not the sole engine.

\subsection{Subspace Decomposition and Coupled Dynamics}\label{subsec:subspace_decomp}

We now extend the warmup’s two-component structure to a more general setting. The guiding principle is that, near a distinguished one-dimensional subspace, the GD dynamics decompose into a drift parallel to the subspace and oscillations orthogonal to it. Expanding the gradient in these two directions reveals that the reduced dynamics are governed by a reduced loss along the parallel direction and a curvature operator in the orthogonal directions.

\myparagraph{Subspace decomposition} Let $\vv\in\sR^d$ be a unit vector along which the curvature vanishes or decays and set $\calM:=\Span\{\vv\}$. The subspace $\calM$ is determined by the geometry of the objective: in the warmup example this was the $x$-axis; in the symmetric classification models studied here, it is the max-margin direction. The losses we study have the property that $\nabla\Loss(\bth)$ lies in a fixed subspace\footnote{For standard classification losses, $\calN$ is the span of the training data, so directions in $\calN^\perp$ are frozen at initialization.} $\calN \subseteq \sR^d$ containing $\calM$ for all $\bth \in \sR^d$. Without loss of generality, we assume the dynamics are initialized in $\calN$ and define $U := \calN \cap \calM^\perp$, so that $\calN=\calM\oplus U$ and every iterate can be written as
\begin{align}
    \bth=\thp\,\vv+\tho,
    \qquad
    \thp:=\langle \bth,\vv\rangle\in\mathbb{R},
    \quad
    \tho:=\Pi_U\bth\in U.
\end{align}
Similarly, the gradient decomposes into two components:
\(\nabla\Loss^\parallel:=\Pi_{\calM}\nabla\Loss\) and
\(\nabla\Loss^\perp:=\Pi_U\nabla\Loss\) denoting the orthogonal projections onto $\calM$ and $U$ respectively.

\myparagraph{Taylor expansion and coupled dynamics} Recall the GD update $\bth_{t+1} =\bth_t-\eta\nabla\Loss(\bth_t)$ and consider a second-order Taylor expansion of the gradient $\nabla\Loss(\bth_t)$ around the parallel direction point $\thp_t\,\vv$.\footnote{This differs from the third-order expansion of the \emph{loss} used in \cite{damianself} and from the expansion around a minimizer manifold in \cite{macdonald2025}; see Section~\ref{sec:related_work} for a detailed comparison.} Projecting onto $\calM$ and $U$ under regularity conditions formalized in the next section, we obtain a coupled two-equation system for the GD iterates:
\begin{align}
  \tho_{t+1}
  &= \bigl(I - \eta\,H(\thp_t)\bigr)\,\tho_t
    + \mathcal{O}(\|\tho_t\|^2),
  \label{eq:gd_perp} \\[3pt]
  \thp_{t+1}
  &= \thp_t - \eta\,\ell'(\thp_t)
    - \tfrac{\eta}{2}\,{\tho_t}^\top H'(\thp_t)\,\tho_t
    + \mathcal{O}(\|\tho_t\|^3),
  \label{eq:gd_par}
\end{align}

where the expansion produces two objects that encode the parallel geometry:
\begin{align}
  \ell(\thp) &:= \Loss(\thp\,\vv),
  &\quad&\text{the \emph{loss restricted to the parallel direction},} \\
  H(\thp) &:= \Pi_U \nabla^2\Loss(\thp\,\vv)\,\Pi_U,
  &\quad&\text{the \emph{Hessian restricted to the parallel direction},}
\end{align}

where $H'(\thp):=\tfrac{d}{d\thp}H(\thp)$ and $\ell'(\thp):=\tfrac{d}{d\thp}\ell(\thp)$. We regard $H(\thp)$ as an operator on $U$ and define
$\lambda_1(\thp):=\lambda_{\max}\!\left(H(\thp)|_U\right).$
All eigenvalues and Loewner inequalities involving $H$ below are
understood on $U$. We call $\lambda_1$ the restricted sharpness;
it need not equal the largest eigenvalue of the full Hessian
at the current iterate.

 The structure mirrors the warmup dynamics. Equation~\eqref{eq:gd_perp} governs the \emph{orthogonal dynamics}: if we neglect the higher order terms, then $\tho_t$ expands or contracts depending on whether the eigenvalues of $\eta H(\thp_t)$ exceed or fall below~$1$. Equation~\eqref{eq:gd_par} governs the \emph{parallel dynamics}: the parallel direction gradient $-\eta\,\ell'(\thp_t)$ provides one potential source of forward drift, while the quadratic coupling $-\tfrac{\eta}{2}\,{\tho_t}^\top H'(\thp_t)\,\tho_t$ converts orthogonal oscillation energy into additional forward progress.

\begin{tcolorbox}[boxrule=0.5pt, colback=blue!5!white, colframe=blue!40!black, boxsep=1pt, top=2pt, bottom=2pt,
  left=4pt, right=4pt]
The coupling term $-\tfrac{\eta}{2}\,{\theta_t^\perp}^{\!\top} H'(\theta_t^\parallel)\theta_t^\perp$ is the source of \textbf{self-stabilization}: this third-order derivative of the loss translates orthogonal oscillations into forward motion along $\calM$, which in turn reduces the curvature and damps the oscillations.
\end{tcolorbox}

The remainders in~\eqref{eq:gd_perp}--\eqref{eq:gd_par} are a potential obstacle if $\tho_t$ grows during the unstable phase. The next section introduces structural assumptions that control these remainders and convert the approximate coupled system into a form amenable to a rigorous convergence analysis.

\section{Convergence Analysis}\label{sec:general_conv}

\begin{table}[t]
\centering
\small
\caption{Summary of the structural assumptions used in the convergence analysis.}
\label{tab:assumptions}
\begin{tabular}{lll}
\toprule
{Category} & Assumption & Role \\
\midrule
parallel direction geometry
& 1.\; Subspace Alignment
& Parallel--orthogonal decoupling on $\calM$ \\
& 2.\; Relative Curvature Comparability
& Curvature control in neighborhood of $\calM$ \\
& 3.\; Loss--Curvature Compatibility
& Parallel drift lowers loss and curvature \\
\midrule
Decay profile
& 4.\; Local Curvature Decay
& Local curvature decay rate \\
& 5.\; Exponential Tail
& Tail gradient/sharpness profile \\
\bottomrule
\end{tabular}
\vspace{-3mm}
\end{table}
The warmup example and the subspace decomposition in Section~\ref{sec:setup} identified several ingredients that drive the self-stabilization mechanism. We distill them into five assumptions, organized in two categories (Table~\ref{tab:assumptions}). The first encodes the \emph{geometric structure} of the loss landscape: a clean subspace decomposition, where the curvature is well-approximated by its subspace value inside a tubular neighborhood, and a parallel direction along which both the loss and the sharpness decrease. The second category provides the \emph{quantitative decay conditions} needed to establish convergence rates: a local guarantee that the curvature cannot plateau in the unstable phase, and
an asymptotic decay condition for the stable phase.

\begin{valleybox}
\begin{enumerate}[label=\textup{\textbf{A.\arabic*.}},leftmargin=*,itemsep=4pt]

\item \textbf{Subspace Alignment.}\label{asmp:SA}
$\Loss$ is $C^4$. Moreover, for all $\thp \in \RR$,
\begin{align}
      \nabla\Loss(\thp \vv) \in \calM, \quad \textup{and} \quad
  \Pi_U\nabla^2\Loss(\thp \vv)\vv=0.
\end{align}

\item \textbf{Relative Curvature Comparability.}\label{asmp:RC}
There exists $\Tilde{\theta}^\parallel$ such that for every sufficiently small $r>0$, there exists $\delta_{r}\in[0,1)$ such that for all
$\thp \ge \Tilde{\theta}^\parallel$ and $\tho\in U$ with $\|\tho\|\le r$,
\begin{equation}\label{eq:RC}
  (1-\delta_{r})\,H(\thp)
  \;\preceq\;
  D_{\tho}\nabla\Loss^\perp(\thp v+\tho)
  \;\preceq\;
  (1+\delta_{r})\,H(\thp).
\end{equation}

\item \textbf{Loss--Curvature Compatibility.}\label{asmp:LCC}
There exists $\thbar$ such that for all $\thp \ge \thbar$:
\begin{enumerate}[label=\textup{(\alph*)},leftmargin=0.2em]
  \item $-\ell'(\thp) > 0$\; (the parallel direction gradient drives $\thp$ forward);
  \item $H(\thp) \succ 0$ and $H'(\thp) \preceq 0$\; (restricted hessian is positive definite and non-increasing on $\calM$).
\end{enumerate}
\end{enumerate}
\end{valleybox}

 \ref{asmp:SA} gives the correct expansion orders: linear in $\tho$ for the orthogonal update and quadratic for the parallel one. \ref{asmp:RC} upgrades the orthogonal dynamics to
the exact multiplicative form
\[
\tho_{t+1}=(I-\eta\widetilde H_t)\tho_t,
\qquad
(1-\delta_r)H(\thp_t)\preceq \widetilde H_t
\preceq (1+\delta_r)H(\thp_t),
\]
where $\delta_r$ measures the tubular neighborhood-curvature distortion. \ref{asmp:LCC} provides self-stabilization: forward motion in $\thp$ decreases the curvature until it falls below the threshold $2/\eta$. Together with the remainder control, local curvature decay, and tail assumptions below, these properties yield finite entry into a region of orthogonal
nonexpansion. In the warmup, $\delta_r=0$ and the loss restricted to the parallel direction vanishes, so forward progress comes solely from the curvature-reduction coupling.

\begin{decaybox}
\begin{enumerate}[label=\textup{\textbf{B.\arabic*.}},leftmargin=1.8em,itemsep=4pt]
\item \textbf{Local Curvature Decay.}\label{asmp:LCD}
For every bounded interval $J \subseteq [\thbar, \infty)$, there exists $\beta_J > 0$ such that $-H'(\thp) \succeq \beta_J\, H(\thp)$ for all $\thp \in J$.

\item \textbf{Exponential Tail.}\label{asmp:ET}
For every tubular neighborhood radius $r>0$, there exists an asymptotic rate $\beta_{\mathrm{tail}}(r)>0$ and a threshold $\thp_{\mathrm{tail}}(r)$ such that, uniformly for all
$\|\tho\|\le r$,
\begin{equation}\label{eq:ET_asymp}
  -\langle \nabla\Loss^{\parallel}(\bth), \vv \rangle
  \;\asymp_r\;
  e^{-\beta_{\mathrm{tail}}(r)(\thp-\thp_{\mathrm{tail}}(r))},
  \qquad
  \lambda_1(\thp)
  \;\asymp_r\;
  e^{-\beta_{\mathrm{tail}}(r)(\thp-\thp_{\mathrm{tail}}(r))}
\end{equation}
for all $\thp\ge \thp_{\mathrm{tail}}(r)$.
Here $f\asymp_r g$ means that there exist constants $0<c_r\le C_r<\infty$, depending only on $r$, such that
$c_r g\le f\le C_r g$ on the stated domain.

\end{enumerate}
\end{decaybox}

\smallskip
 \ref{asmp:LCD} strengthens \ref{asmp:LCC}(b) by requiring the curvature to decrease at a definite rate on every compact interval, ruling out plateaus before the stability threshold. \ref{asmp:ET} requires that the parallel gradient and sharpness decay exponentially in $\thp$ at the same rate $\beta_{\mathrm{tail}}$.

\begin{rem}[Uniform conditioning]
\label{rem:conditioning}
When $\dim U > 1$, contraction of the orthogonal component requires that the restricted Hessian satisfy for all $\thp$: $H(\thp) \succeq \rho\,\lambda_1(\thp)\,I_U$
for some $\rho \in (0,1]$, i.e., $\kappa(H(\thp)|_U) \le \rho^{-1}$.  This fails in geometrically interesting cases (Section~\ref{sec:applications}), so rather than imposing it as a blanket assumption, we state it as an explicit hypothesis whenever it is needed.
\end{rem}

\begin{wrapfigure}{r}{0.5\textwidth}
\vspace{-2mm}
  \centering
  \includegraphics[
    width=0.5\textwidth,
    trim=5 10 10 10
  ]{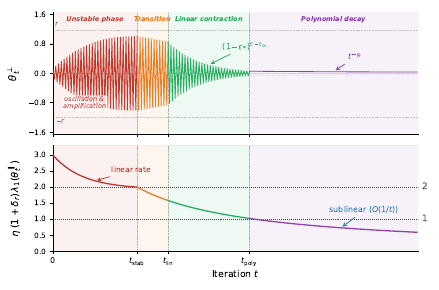}

  \caption{
  Roadmap of Theorem~\ref{thm:main}.
  }
  \label{fig:main_schematic}
  \vspace{-2mm}
\end{wrapfigure}

We now state the main convergence theorem, which gives a quantitative
version of the coupled dynamics \eqref{eq:gd_perp}--\eqref{eq:gd_par} in Section~\ref{sec:setup}. Equation~\eqref{eq:gd_par} implicitly performs gradient descent on the restricted curvature: forward motion in $\theta_t^\parallel$ decreases the sharpness and eventually the stability threshold is reached. Equation~\eqref{eq:gd_perp} then translates this change in the curvature into orthogonal dynamics, distinguishing contraction, expansion, and nonexpansion according to the spectrum of $I-\eta H(\theta_t^\parallel)$. Figure~\ref{fig:main_schematic} illustrates the different phases of the dynamics.

\myparagraph{Admissibility}
Fix a tubular neighborhood radius $r > 0$ and define the \emph{stability threshold}
\begin{align}\label{eq:corrected_anchor}
  \thp_*{(r, \eta)}
  := \inf\bigl\{\thp \ge \thp_0 :\;
     \eta(1+\delta_r)\lambda_1(\thp) \le 2\bigr\}.
\end{align}

 Since $r$ and $\eta$ are fixed throughout the analysis, we write simply $\thp_*$ hereafter. The triple $(\eta, r, \bth_0)$ is called \emph{admissible} if several requirements hold simultaneously: the initialization is unstable $\eta(1+\delta_r)\lambda_1(\thp_0) > 2$ and starts inside the tubular neighborhood; the radius $r$ is small enough that the Taylor remainder in equation~\eqref{eq:gd_par} does not overwhelm the curvature-reduction mechanism; and all the orthogonal oscillations fit inside the tubular neighborhood. These requirements impose a \emph{large but not too large} step  size: $\eta$ must induce oscillations while keeping them within the perturbative region where curvature comparability and  Taylor approximations hold. Precise constraints on $r$ and $\eta$ appear in Appendix~\ref{def:admissibility}.

\begin{theorem}[Convergence of GD]
\label{thm:main}
Suppose Assumptions A and B hold with $\thp_* \geq \thp_{\mathrm{tail}}(r)$ and that the triple $(\eta, r, \bth_0)$ is admissible with $\thp_0 \ge \max(\thbar, \tilde{\theta}^\parallel)$. The dynamics have two main phases, separated by the first time the
parallel coordinate reaches the stability threshold, $t_{\mathrm{stab}}
  := \inf\{t\ge 0:\thp_t\ge\thp_*\}.$
\noindent
\textbf{I.\; Unstable phase} $(0 \le t \le t_{\mathrm{stab}})$\textbf{.}\;
The iterates remain in the tubular neighborhood: $\|\tho_t\| \le r$ for all
$t \le t_{\mathrm{stab}}$.
The parallel coordinate grows at least linearly and the sharpness decays
exponentially:
\begin{equation}\label{eq:pre_schedules}
  \thp_t-\thp_0 \;\gtrsim\; t,
  \qquad
 \lambda_1(\thp_t)
  \;\lesssim\;
  \lambda_1(\thp_0)\,e^{-ct},
  \qquad
  0\le t<t_{\mathrm{stab}},
\end{equation}
for a constant $c>0$. In particular, $t_{\mathrm{stab}} < \infty$ and $\eta(1+\delta_r)\lambda_1(\thp_{t_{\mathrm{stab}}}) \le 2$.

\noindent
\textbf{II.\; Stable phase} $(t \ge t_{\mathrm{stab}})$\textbf{.} The parallel coordinate grows at least logarithmically and the sharpness decays sublinearly:
\begin{equation}\label{eq:post_schedules}
  \thp_t - \thp_*
  = \Omega(\log( t)),
  \qquad
  \lambda_1(\thp_t) = \mathcal{O}\left(1/t\right).
\end{equation}
Assume in addition that there exists $\rho_{\mathrm{tail}} \in (0,1]$ such that $H(\thp) \succeq \rho_{\mathrm{tail}}\,\lambda_1(\thp)\,I_U$ for all $\thp \ge \thp_*$. Then the stable phase can be further divided into three regimes according
to the corrected sharpness. Let
$\varepsilon_*:=\rho_{\mathrm{tail}}(1-\delta_r)/(1+\delta_r)$
be the contraction margin, and let $t_{\mathrm{lin}}$ and
$t_{\mathrm{poly}}$ denote the first times after stabilization
at which the corrected sharpness is at most $2-\varepsilon_*$
and $1$, respectively:
\[
\begin{aligned}
  t_{\mathrm{lin}}
  :=\inf\{t\ge t_{\mathrm{stab}}:
     \eta(1+\delta_r)\lambda_1(\thp_t)\le 2-\varepsilon_*\},\quad
  t_{\mathrm{poly}}
  :=\inf\{t\ge t_{\mathrm{stab}}:
     \eta(1+\delta_r)\lambda_1(\thp_t)\le 1\}.
\end{aligned}
\]
These times satisfy
$t_{\mathrm{stab}}\le t_{\mathrm{lin}}\le t_{\mathrm{poly}}<\infty$,
with equality possible at either boundary.
\begin{enumerate}[label=\textup{(\roman*)},leftmargin=1.4em]
  \item \textbf{Transition Phase} $t\in[t_{\mathrm{stab}},t_{\mathrm{lin}}]$:
    the oscillations are nonexpanding,
    $\|\tho_t\| \le \|\tho_{t_{\mathrm{stab}}}\|$.
  \item \textbf{Linear Contraction Phase}
    $t\in[t_{\mathrm{lin}},t_{\mathrm{poly}}]$:
    $\|\tho_t\| \le (1-\varepsilon_*)^{t - t_{\mathrm{lin}}}\,
    \|\tho_{t_{\mathrm{lin}}}\|$ with $\varepsilon_*
:=
\frac{\rho_{\mathrm{tail}}(1-\delta_r)}{1+\delta_r}$.
  \item \textbf{Polynomial Decay Phase} $(t \ge t_{\mathrm{poly}})$:
    $\|\tho_t\| = O\!\left((t-t_{\mathrm{poly}})^{-\gamma_r}\right)$ with $\gamma_r = \rho_{\mathrm{tail}}(1-\delta_r) C_{\mathrm{tail}}(r)$ and $C_{\mathrm{tail}}(r)$ a positive constant determined by the tail bounds and the tubular neighborhood tolerance.
\end{enumerate}
\end{theorem}

After the iterates stabilize, the orthogonal component initially
contracts linearly, but this rate is self-limiting: the
parallel motion that stabilizes the dynamics simultaneously drives
the restricted curvature to zero, yielding the polynomial tail
$t^{-\gamma_r}$. The phase boundaries are determined by the corrected sharpness $\eta(1+\delta_r)\lambda_1(\theta_t^\parallel)$.
During the unstable phase, this quantity exceeds $2$; the orthogonal dynamics can therefore be locally expansive. Self-stabilization drives $\theta_t^\parallel$ forward and decreases the sharpness until the threshold $2$ is crossed at $t_{\mathrm{stab}}$. After stabilization, crossing $2-\varepsilon_*$ at $t_{\mathrm{lin}}$ provides a uniform contraction margin, yielding geometric decay of the orthogonal component. Finally, when the corrected sharpness falls below $1$ at $t_{\mathrm{poly}}$, the per-step contraction becomes positive but progressively weaker as the sharpness tends to zero; the resulting product of contraction factors yields the polynomial tail. The conditioning constant $\rho_{\mathrm{tail}}$ separates two usually conflated effects: $\lambda_1$ determines \emph{when} the iterates stabilize, while $\rho_{\mathrm{tail}}$ determines \emph{whether} all orthogonal directions are damped comparably. This distinction is central in Section~\ref{sec:applications}. Finally, we note that the iterate schedules can be converted into convergence rates for the loss (see Appendix~\ref{app:loss_rates}).

\section{Instantiations of the Theory and Their Consequences}
\label{sec:applications}

This section instantiates our general framework on canonical models from machine learning. Specifically, we verify the structural assumptions of Section~\ref{sec:general_conv} for standard losses such as logistic and cross-entropy, and illustrate how the general theory yields sharper, model-specific insights, including geometry-dependent effects not visible at the level of the abstract theory.

\myparagraph{Verification for pair-symmetric logistic regression} Fix a unit vector $\vv \in \mathbb{R}^d$. Consider $N$ points $\vx_i^+ = a_i\,\vv + \vu_i$ with $a_i > 0$, $\vu_i \in \Span\{\vv\}^\perp \setminus \{0\}$, labeled $+1$, and their reflections across \(\Span\{\vv\}^{\perp}\), $\vx_i^- = -a_i\,\vv + \vu_i$, labeled $-1$. Writing $\bth = \thp\vv + \tho$ with $\tho \in U := \Span\{\vu_1,\dots,\vu_N\}$, the logistic loss is
\begin{equation}\label{eq:logistic_loss_hd}
  \Loss(\bth)
  = \frac{1}{2N} \sum_{i=1}^{N} \Bigl[
      \phi\!\bigl(a_i \thp + \langle \vu_i, \tho \rangle\bigr)
    + \phi\!\bigl(a_i \thp - \langle \vu_i, \tho \rangle\bigr)
  \Bigr],
  \qquad
  \phi(z) := \log(1 + e^{-z}),
\end{equation}
The pair symmetry $\tho \mapsto -\tho$ leaves each summand in~\eqref{eq:logistic_loss_hd} invariant, so the subspace $\calM = \Span\{\vv\}$ is precisely the max-margin line. The restricted Hessian matrix is
\begin{equation}
    H(\thp) = \tfrac{1}{N}\sum_i \phi''(a_i\thp)\,\vu_i \vu_i^\top.
\end{equation}
 Set $a_{\min} := \min_i a_i$ and $u_{\max} := \max_i \|\vu_i\|$. The set $I_* := \{i : a_i = a_{\min}\}$ is, by construction, the set of indices corresponding to support vectors. Although our numerical illustrations use $d=3$ for visualization, the pair-symmetric construction and the verification below hold in arbitrary ambient dimension; in particular, the active orthogonal subspace $U=\Span\{u_i\}$ may have arbitrary dimension.

\begin{prop}[Logistic Verification]
\label{prop:logistic_verification}
For every fixed $\bar{\theta}^{\parallel}>0$, the pair-symmetric logistic loss~\eqref{eq:logistic_loss_hd} satisfies Assumptions A and B with this choice of lower threshold in Assumptions \ref{asmp:LCC}--\ref{asmp:LCD}.
The following constants may be used:
\begin{enumerate}[label=\textup{(\roman*)},leftmargin=2.4em]
  \item \textbf{RC tolerance:}\;
    $\delta_r = \cosh(u_{\max}\, r) - 1 = \tfrac{1}{2}(u_{\max}\, r)^2
    + O(r^4)$, with $0 < r < \frac{\mathrm{arccosh(2)}}{u_{max}}$.
  \item \textbf{Exponential tail rate:}\;
    $\beta_{\mathrm{tail}} = a_{\min}$.
  \item \textbf{Uniform conditioning:}\;
    If the \emph{spanning condition}
    $U = \Span\{\vu_i : i \in I_*\}$ holds, then $H(\thp)
      \succeq \rho_{\mathrm{tail}}\, \lambda_1(\thp)\, I_U$ for all $\thp \ge 0$ with
    \[
      \rho_{\mathrm{tail}}
      = \frac{\lambda_{\min}(\Sigma_*|_U)}
             {\lambda_{\max}(\Sigma|_U)},
    \]
    where $\Sigma := \tfrac{1}{N}\sum_i \vu_i \vu_i^\top$,\; $\Sigma_* := \tfrac{1}{N}\sum_{i \in I_*} \vu_i \vu_i^\top$.
\end{enumerate}
Explicit expressions for all remaining constants are given in Appendix~\ref{app:logistic_verification}.
\end{prop}

\myparagraph{Extension to multiclass cross-entropy}
In the multiclass model, the two binary class directions $\pm\vv$ are
replaced by the vertices $\vr_1,\ldots,\vr_K$ of a centered regular
simplex. For each $i$, the examples $\vx_{c,i}=b_i\vr_c+\vu_i$ share
the same off-axis component $\vu_i$ across all classes. Writing
$R=[\vr_1,\ldots,\vr_K]$, the parallel direction is now the line
$\mathcal M=\Span\{R\}$ in weight-matrix space. Its positive ray is
generated by the Crammer--Singer hard-margin solution
\cite{crammer2001algorithmic,ravi2024implicit}.

 By symmetry, the gradient on the line is parallel
to $R$, so the parallel dynamics again reduce to a single scalar. The orthogonal dynamics are richer: even when $\vu_i=0$ for every $i$, perturbations within the simplex span can change the relative orientations of the class weights. Their curvature is governed by the softmax covariance of the class-logit perturbations, which plays the role of the scalar logistic curvature in the binary analysis. Full statements and proofs are in Appendix~\ref{app:cross_entropy}.

\begin{wrapfigure}{r}{0.5\textwidth}
\vspace{-4mm}
  \centering
    \includegraphics[
    width=\linewidth,
    trim=10 10 10 10
  ]{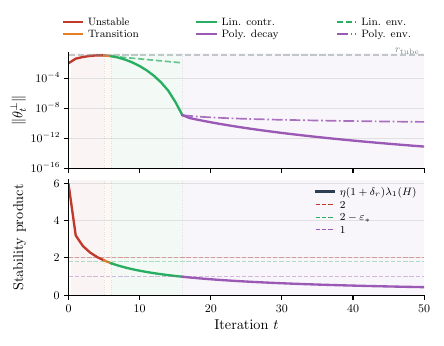}
    \caption{
Unstable and stable phases in a spanning logistic model
}
    \label{fig:rate_verification}
    \vspace{-3mm}
\end{wrapfigure}

\myparagraph{Effects of spanning geometry on the dynamics}
The convergence behavior of the orthogonal component depends on whether the orthogonal components of the support vectors span the orthogonal subspace~$U$. To quantify this, define the
\emph{conditioning ratio}
$\rho(\theta^\parallel)
:=
\lambda_{\min}(H(\theta^\parallel))
/\lambda_{\max}(H(\theta^\parallel))
\in(0,1].$
In the \emph{spanning case}, $\rho$ stays bounded away from zero and
the top eigenvalue $\lambda_1$ governs contraction of the full
orthogonal component. In the \emph{non-spanning case}, $\rho \to 0$
as $\thp \to \infty$: different orthogonal directions contract at
different rates (Figure~\ref{fig:nonspanning}), and the slowest may receive only a finite cumulative contraction budget. Both pair-symmetric logistic regression and its
multiclass cross-entropy analogue exhibit this dichotomy; we present
the binary derivation and defer the multiclass case to
Appendix~\ref{app:cross_entropy}.

Let $\alpha_1 < \cdots < \alpha_L$ be the distinct values among \(a_1,\dots,a_N\), and define the nested subspaces
\begin{equation*}
    U^{(1)} \subseteq U^{(2)} \subseteq \cdots \subseteq U^{(L)} = U, \quad U^{(j)} := \Span\{\vu_i : a_i\le \alpha_j\}
\end{equation*}
Let $j_* := \min\{j : U^{(j)} = U\}$. Recalling that the support
vectors of the max-margin classifier are exactly the
examples $\{\vx_i^\pm : i \in I_*\}$, we have $j_* = 1$ if and only if the orthogonal components of the support vectors span $U$.

\begin{theorem}[Spanning vs. non-spanning geometry]
\label{thm:rho_order_compact}
In the stable phase, the following holds:
\begin{enumerate}[label=\textup{(\roman*)},leftmargin=1.2em]
\item If \(j_*=1\), then \(\rho(\theta_t^\parallel)\asymp 1\) and the three-phase
structure of Theorem~\ref{thm:main}\textup{(II)} applies;
\item If \(j_*>1\), then $\rho(\theta_t^\parallel) \asymp t^{-(\alpha_{j_*}-\alpha_1)/\alpha_1}.$ Under an additional simultaneous diagonalizability assumption, explicit componentwise rates can be derived (see Appendix~\ref{app:componentwise_rates}).
\end{enumerate}
\end{theorem}

\myparagraph{Transient implicit bias induced by non-spanning geometry}  A key implication of our analysis is the emergence of a transient implicit bias: although GD converges asymptotically to the max-margin direction, the iterates can exhibit persistent deviations due to orthogonal components that are amplified during the unstable phase and decay only partially in the stable phase. This phenomenon arises specifically in the non-spanning case ($j_* > 1$): Theorem~\ref{thm:rho_order_compact} shows that the
conditioning ratio $\rho(\thp_t)$ decays fast enough that the
cumulative contraction budget $\sum_t \rho(\thp_t)\,\lambda_1(\thp_t)$ remains finite, in contrast to the spanning case where it diverges and drives
$\|\bth_t^\perp\| \to 0$. As a result, GD may temporarily
favor directions outside the max-margin subspace, leading to a
geometry-dependent bias controlled by the
step size~$\eta$: two competing effects --- oscillatory amplification and
early stable-phase contraction --- can make $\|\bth_t^\perp\|$
increasing, decreasing, or non-monotone in~$\eta$ depending on the
data geometry and initialization. We exhibit the increasing case in
Section~\ref{sec:experiments} and defer the others to
Appendix~\ref{app:additional_experiments}.

\section{Empirical Validation}\label{sec:experiments}

\myparagraph{Trajectory structure and three-phase decay}
We illustrate the trajectory structure predicted by Theorem~\ref{thm:main} on a spanning logistic model in $\RR^3$
with $N = 24$ data pairs and $\dim U = 2$ (setup details in
Appendix~\ref{app:experiment_setup}). Figure~\ref{fig:rate_verification}
confirms the full predicted trajectory: oscillatory growth inside the tubular neighborhood, followed by linear contraction at rate $(1-\varepsilon_*)^t$ and
polynomial decay with exponent $\gamma_r = 0.84$. The linear bound is
tight initially but loosens as $\eta\lambda_1$ shrinks below the
worst-case threshold.

\myparagraph{Non-spanning geometry and transient implicit bias}
We next illustrate the geometry analyzed in Theorem~\ref{thm:rho_order_compact} on a non-spanning dataset where the support vectors do not span $U$ (setup in Appendix~\ref{app:experiment_setup}). Figure~\ref{fig:nonspanning} reveals a componentwise split: the support-vector component $|c_{\mathrm{sv}}|$ contracts through the three-phase mechanism and decays to numerical precision, while the weak component $|c_{\mathrm{wk}}|$ plateaus at a nonzero value. Crucially, larger $\eta$ amplifies $|c_{\mathrm{wk}}|$ during the unstable phase, and the non-spanning tail dynamics cannot remove this amplification — larger steps leave a larger weak residual. This residual is invisible from the loss alone but manifests as a larger finite-time angle to the max-margin direction, confirming the transient implicit bias predicted by the theory. We leave a systematic investigation of connections to generalization and robustness to future work. Our experiments are intended as controlled tests of the predicted dynamics rather than large-scale validation on deep networks. Exact Hessian formation is not needed by GD itself, and dominant eigenvalues can be estimated using Hessian–vector products; however, identifying and verifying the appropriate reference subspace in modern networks remains an open problem.

\begin{figure}[t]
    \centering

    \includegraphics[width=0.9\linewidth,
      trim=0 5pt 0 5pt, clip]{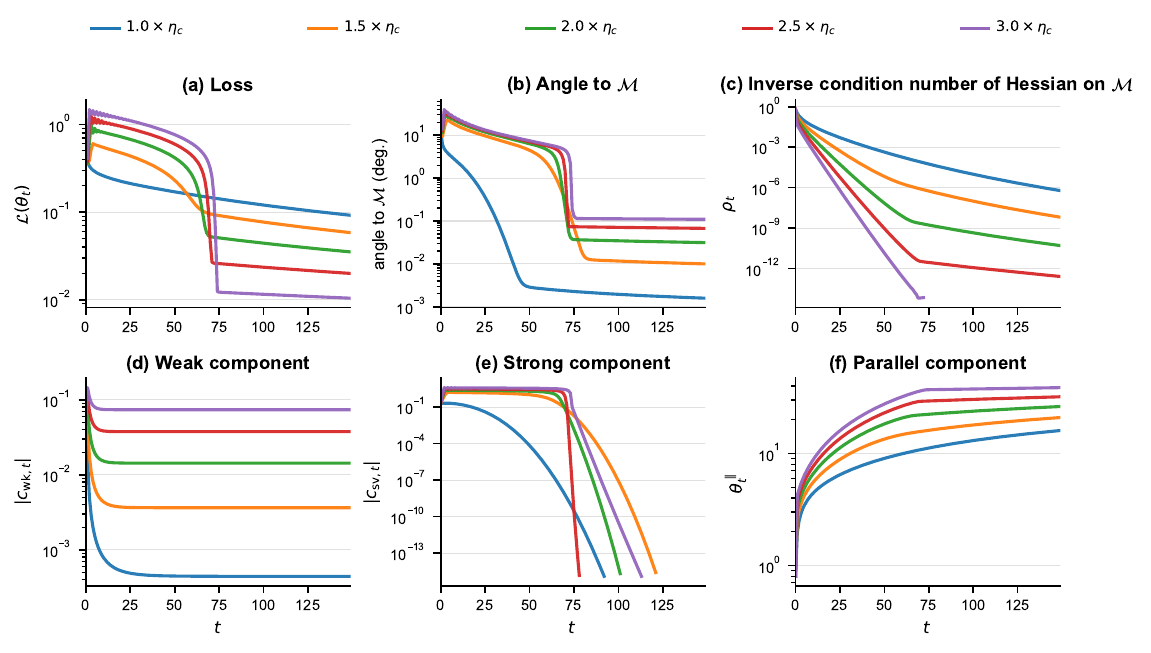}
    \caption{Non-spanning logistic regression at step sizes
      $\eta/\eta_c \in \{1, 1.5, 2, 2.5, 3\}$, where
      $\eta_c := 2/\lambda_{\max}(\nabla^2\Loss(\bth_0))$.
      \textbf{(a)}~Loss. \textbf{(b)}~Angle to max-margin
      line $\calM$. \textbf{(c)}~Conditioning ratio $\rho_t$.
      \textbf{(d)}~Weak component $|c_{\mathrm{wk}}|$.
      \textbf{(e)}~Strong component $|c_{\mathrm{sv}}|$.
      \textbf{(f)}~Parallel coordinate.}
    \label{fig:nonspanning}
    \vspace{-3mm}
\end{figure}

\section{Conclusion}\label{sec:conclusion}
We introduced a unified framework for analyzing gradient descent at the edge of stability that converts the qualitative self-stabilization picture into quantitative convergence guarantees for exponential tail losses. The framework not only recovers prior results for logistic regression and cross-entropy classification, but also extends them by providing a unified mechanism that explains and quantifies their large-step dynamics. Crucially, it reveals a phenomenon invisible to prior analyses of GD: under certain data geometries, the step size induces a transient implicit bias that affects the alignment with the max-margin direction. We leave the characterization of this dependence and its role in generalization and robustness to future work, as well as extensions to asymmetric data or deeper architectures that could further broaden the scope of the framework.

\section*{Acknowledgments and Disclosure of Funding}

This research was supported by NSF 2031985 and Simons Foundation 814201 (Theorinet), ONR MURI 503405-78051 and University of Pennsylvania Startup Funds.

\bibliographystyle{unsrt}
\bibliography{references}

\newpage

\appendix

\section*{Appendix Roadmap}
\begin{itemize}[leftmargin=*]

\item Appendix~\ref{app:related_work}: additional related work.
\item Appendix~\ref{app:full_main_theorem}: full proof of Theorem~\ref{thm:main}.
\item Appendix~\ref{app:logistic_verification}: verification for pair-symmetric logistic regression.
\item Appendix~\ref{app:cross_entropy}: multiclass cross-entropy extension.
\item Appendix~\ref{app:additional_experiments}: additional experiments and setup details.
\item Appendix~\ref{app:approximate_symmetry}: approximate symmetry.
\end{itemize}

\section{Additional Related Work on Large Step Size Gradient Descent}\label{app:related_work}

\myparagraph{Edge of Stability (EoS)}
The empirical success of training neural networks with learning rates near or above classical stability limits has generated substantial interest in understanding GD beyond the regime predicted by standard smoothness-based analysis. Earlier work had already connected step size, local curvature, and sharpness in neural-network training~\cite{wu2018sgd, jastrzkebskirelation}, related large-learning-rate phenomena such as the catapult mechanism~\cite{lewkowycz2020large}, and showed that training can transiently violate monotone descent while still recovering and converging~\cite{jastrzebskibreak, xing2018walk, li2019towards, leclerc2020two}. The seminal work of Cohen et al.~\cite{cohen2021gradient} conducted a systematic empirical study of this behavior, showing that GD on deep networks routinely operates near the classical threshold of \(2/\lambda_{\max}\), with sharpness rising until it hovers around the stability boundary; they termed this phenomenon the \emph{edge of stability}. Along with subsequent work ~\cite{macdonald_progressive_2023,wang2022analyzing}, they also documented progressive sharpening, whereby local curvature increases along the training trajectory and pushes the system toward instability even at a fixed step size.

\myparagraph{Model-specific analyses and implicit biases} A substantial body of work analyzes the EoS in specific architectures or simplified settings. Studies of diagonal linear and related homogeneous models have shown that large but stable step sizes can change the implicit bias of GD/SGD~\cite{nacson2022implicit,even2023s,wang2023good}, while analyses of two-layer non-homogeneous networks reveal additional step-size-dependent biases~\cite{cai2024large}. Chen and Bruna~\cite{chen2023beyond} analyze convergence to period-2 orbits through two-step gradient updates in one-dimensional functions, single-neuron models, and matrix factorization, focusing on local mechanisms rather than end-to-end convergence rates. Other works have used minimalist examples or bifurcation theory to probe the dynamics near the stability boundary~\cite{song2023trajectory, zhu2023understanding, kreisler2023gradient, liu2025minimalist}, and analyses of SGD have explored benign oscillations and step-size-dependent selection of global minima~\cite{lu_benign_2023}. Complex phenomena including cycles and chaotic behavior have been documented beyond the EoS~\cite{chen2024stability, ghosh_learning_2025}, and \cite{cruaciun2024convergence} provides impossibility results for global convergence from arbitrary initializations with very large step sizes, while identifying specific conditions under which convergence still occurs. These analyses yield rich insights into particular models but rely on architectural details that make it difficult to extract a unified mechanism.
\newpage
\section{Proof of Theorem~\ref{thm:main}}
\label{app:full_main_theorem}

This section gives the full quantitative version of Theorem~\ref{thm:main} and its proof. Section~\ref{app:setup_admissibility} collects the derived constants and states the admissibility conditions that the tubular neighborhood radius~$r$, the step size $\eta$ and the initialization $\mathbf{\theta}_0$ must satisfy.
Section~\ref{app:eos_phase} establishes tubular neighborhood invariance, monotone drift, and finite entry to the corrected stability regime (Theorem~\ref{thm:eos_phase}). Section~\ref{app:stable_phase} derives the stable regime  schedules and the three-phase orthogonal decay (Theorem~\ref{thm:stable_phase}). Together, these two theorems imply Theorem~\ref{thm:main} as explained in Remark~\ref{rem:main_relationship}.

\subsection{Setup and admissibility}
\label{app:setup_admissibility}
Throughout this appendix, Assumptions A and B are in force.
Fix a step size $\eta > 0$, an initial condition
$\bth_0 = \thp_0\, v + \tho_0$ with $\tho_0 \in U$ and
$\thp_0 \ge \bar{\thp}$, and a tubular neighborhood radius $r > 0$.

\begin{defn}[Stability threshold and oscillation slab]
\label{def:anchor_slab}
The \emph{stability threshold} is
\begin{equation}\label{eq:anchor}
  \thp_*
  := \inf\bigl\{\thp \ge \thp_0 :\;
     \eta(1+\delta_r)\lambda_1(\thp) \le 2\bigr\},
\end{equation}
and the \emph{oscillation slab} is
\[
  \mathcal{T}_{\mathrm{pre}}
  := \bigl\{\thp\, v + \tho :\;
       \thp \in [\thp_0, \thp_*],\;
       \tho \in U,\;
       \|\tho\| \le r\bigr\}.
\]
\end{defn}
 We assume throughout that \(\thp_*<\infty\). This is automatic under the
tail sharpness assumption, since \(\lambda_1(\thp)\to0\) exponentially.
Since $\mathcal{T}_{\mathrm{pre}}$ is compact and $\Loss$ is $C^4$
by Assumption~\ref{asmp:SA}, the following constants are finite and determined
by $\Loss$, $\eta$, $r$, and $\thp_0$:
\begin{alignat}{2}
  m_\ell
  &:= \inf_{\thp \in [\thp_0, \thp_*]}
       \bigl(-\ell'(\thp)\bigr)
  &&> 0,
  \label{eq:m_ell}
  \\[4pt]
  G_\parallel
  &:= \sup_{\mathcal{T}_{\mathrm{pre}}}
       \langle -\nabla\Loss^\parallel(\bth), \vv \rangle
  \quad&&< \infty,
  \label{eq:Gpar}
  \\[4pt]
  \beta
  &:= \beta_{[\thp_0,\thp_*]},
  \qquad
  &&\text{the log-derivative constant from Assumption~\ref{asmp:LCD}}.
  \label{eq:beta_pre}
\end{alignat}
The positivity of $m_\ell$ follows from~\ref{asmp:LCC} and the
compactness of $[\thp_0, \thp_*]$.
Let $c_{\parallel,r} > 0$ be a constant such that for all
$(\thp,\tho) \in [\thp_0,\thp_*] \times U$ with
$\|\tho\| \le r$,
\begin{equation}\label{eq:taylor_remainder}
  \langle \nabla\Loss(\thp\,\vv+\tho),\vv\rangle
  =
  \ell'(\thp)
  +\tfrac12\langle H'(\thp)\tho,\tho\rangle
  +R^\parallel(\thp,\tho),
  \qquad
  |R^\parallel(\thp,\tho)|
  \le c_{\parallel,r}\|\tho\|^3.
\end{equation}
Such a constant exists by the $C^4$ regularity of $\Loss$ and
compactness of $\mathcal{T}_{\mathrm{pre}}$.

\myparagraph{Uniform conditioning in the unstable phase}
By Assumption~\ref{asmp:LCC}(b), $H(\thp)$ is positive
definite on $U$. Continuity and compactness therefore give
\[
  \rho
  :=
  \min_{\thp\in[\thp_0,\thp_*]}
  \frac{\lambda_{\min}(H(\thp)|_U)}
       {\lambda_1(\thp)}
  >0.
\]
Consequently,
\begin{equation}\label{eq:pre_cond}
  H(\thp)\succeq \rho\,\lambda_1(\thp)I_U
  \qquad
  \forall\,\thp\in[\thp_0,\thp_*].
\end{equation}
If $\dim U=1$, then $\rho=1$.
Unlike uniform conditioning on an unbounded tail, this
compact-slab bound follows directly from positive definiteness.

\begin{defn}[Admissibility]
\label{def:admissibility}
Define the \emph{remainder-smallness parameter}
\begin{equation}\label{eq:chi}
  \chi
  := \beta
     - \frac{\eta(1+\delta_r)\,c_{\parallel,r}\,r}
            {\rho},
\end{equation}
and the \emph{initial instability surplus}
\begin{equation}\label{eq:Cunst}
  C_{\mathrm{unst}}
  := \eta(1+\delta_r)
     \bigl[\eta(1+\delta_r)\lambda_1(\thp_0) - 2\bigr].
\end{equation}
The triple $(\eta, r, \bth_0)$ is \emph{admissible} if:
\begin{enumerate}[label=\textup{(A\arabic*)},leftmargin=2.6em]
  \item \label{cond:unstable}
    $\eta(1+\delta_r)\lambda_1(\thp_0) > 2$
    and $\|\tho_0\| \le r$,
    \hfill\emph{(corrected instability)}
  \item \label{cond:chi}
    $\chi > 0$,
    \hfill\emph{(remainder smallness)}
  \item \label{cond:tubular neighborhood}
    $\displaystyle
    \|\tho_0\|^2
    + \frac{2\,C_{\mathrm{unst}}}
           {\eta\,\chi}
      \bigl(\thp_* - \thp_0 + \eta\, G_\parallel\bigr)
    \le r^2$.
    \hfill\emph{(tubular neighborhood-closing budget)}
\end{enumerate}
\end{defn}

\subsection{Non-emptiness of the admissible window}\label{app:admissibility_non_emptiness}

Admissibility carves out a \emph{moderately unstable} regime: large
enough to trigger oscillatory dynamics, but not so large that the local Taylor and curvature-comparability estimates break down. The three conditions are mutually compatible.  Conditions~\ref{cond:unstable}
and~\ref{cond:chi} together require
\begin{equation}\label{eq:eta_window}
  \frac{2}{\lambda_1(\thp_0)}
  \;<\;
  \eta(1+\delta_r)
  \;<\;
  \frac{\rho\beta}{c_{\parallel,r}r},
\end{equation}
a non-empty open interval whenever
\[
  r < \frac{\rho\beta\lambda_1(\thp_0)}{2c_{\parallel,r}}.
\]
In terms of \(\eta\) itself, this corresponds to
\[
  \eta\in
  \Bigl(
    \frac{2}{(1+\delta_r)\lambda_1(\thp_0)},\;
    \frac{\rho\beta}{(1+\delta_r)c_{\parallel,r}r}
  \Bigr).
\]
The constant \(c_{\parallel,r}\) in~\eqref{eq:taylor_remainder} depends
a priori on the slab \([\thp_0,\thp_*]\), hence on \(\eta\) through
\(\thp_*\). Fix $\thp_0\ge\max\{\thbar,\tilde{\theta}^{\parallel}\}$
and a reference interval
\[
  J=[\thp_0,\thp_0+\Delta],
  \qquad \Delta>0.
\]
Choose a reference radius $r_0>0$ on which relative curvature
comparability is available. On the fixed reference slab-tubular neighborhood,
choose constants $\beta_J>0$, $\rho_J>0$, $c_J<\infty$, and
$G_J<\infty$ such that
\[
  -H'(\thp)\succeq\beta_J H(\thp),
  \qquad
  H(\thp)\succeq\rho_J\lambda_1(\thp)I_U
  \qquad(\thp\in J),
\]
and the parallel Taylor remainder and drift satisfy
\[
  |R^\parallel(\thp,\tho)|\le c_J\|\tho\|^3,
  \qquad
  -\langle\nabla\Loss(\thp\vv+\tho),\vv\rangle\le G_J
\]
for $\thp\in J$ and $\|\tho\|\le r_0$.

Write $\lambda_0:=\lambda_1(\thp_0)$. Choose an
RC-valid radius $r\in(0,r_0]$ with $\delta_r<1$ and
\[
  \chi_c
  :=
  \beta_J-\frac{2c_Jr}{\rho_J\lambda_0}>0.
\]
Define
\[
  \eta_c(r):=\frac{2}{(1+\delta_r)\lambda_0}.
\]
For $\eta>\eta_c(r)$ sufficiently close to $\eta_c(r)$,
local curvature decay gives
\[
  \lambda_1(\thp)
  \le \lambda_0 e^{-\beta_J(\thp-\thp_0)}
  \qquad(\thp\in J).
\]
It follows that
\[
  0<\thp_*-\thp_0
  \le
  \frac{1}{\beta_J}
  \log\!\left(\frac{\eta}{\eta_c(r)}\right)
  <\Delta.
\]
Thus the true unstable-phase slab lies inside $J$, and
the reference constants can be used in the admissibility
conditions.

With these constants,
\[
  \chi(\eta)
  =
  \beta_J-\frac{\eta(1+\delta_r)c_Jr}{\rho_J}
  \longrightarrow \chi_c>0
  \qquad\text{as }\eta\downarrow\eta_c(r).
\]
Hence $\chi(\eta)\ge\chi_c/2$ for $\eta$ sufficiently
close to $\eta_c(r)$. Moreover,
$C_{\mathrm{unst}}(\eta)\to0$, while
$\thp_*-\thp_0+\eta G_J$ remains bounded. Therefore
\[
  \frac{2C_{\mathrm{unst}}(\eta)}
       {\eta\chi(\eta)}
  \bigl(\thp_*-\thp_0+\eta G_J\bigr)
  \longrightarrow0.
\]
For every initialization with $\|\tho_0\|\le r/2$,
the tubular neighborhood-closing budget consequently holds with strict slack
when $\eta>\eta_c(r)$ is sufficiently close to $\eta_c(r)$.

This proves that, for each such fixed radius, admissible step
sizes contain a nonempty interval immediately above $\eta_c(r)$.
If additionally $\thp_0\ge\thp_{\mathrm{tail}}(r)$, then
$\thp_*\ge\thp_0$ also guarantees the tail-entry condition
required by the stable-phase theorem.

\subsection{Unstable phase: tubular neighborhood invariance and finite time entry to the stable phase}
\label{app:eos_phase}

\begin{theorem}[Unstable phase]
\label{thm:eos_phase}
Suppose $(\eta, r, \bth_0)$ is admissible
\textup{(Definition~\ref{def:admissibility})}.
Let $t_{\mathrm{stab}} := \inf\{t \ge 0 : \thp_t \ge \thp_*\}$.
Then:
\begin{enumerate}[label=\textup{(\roman*)},leftmargin=2.4em]

\item \emph{Tubular neighborhood invariance:}\;
  $\|\tho_t\| \le r$ for all $0 \le t \le t_{\mathrm{stab}}$.

\item \emph{Monotone forward drift:}\;
  $\thp_{t+1} - \thp_t \ge \eta\, m_\ell > 0$ for all
  $0 \le t < t_{\mathrm{stab}}$.

  \item \emph{Exponential sharpness decay:}\;
  for all $0 \le t < t_{\mathrm{stab}}$,
  \[
    \lambda_1(\thp_t)
    \le
    \lambda_1(\thp_0)\,
    e^{-\beta\eta m_\ell t}.
  \]

\item \emph{Finite entry time:}\;
  $t_{\mathrm{stab}}
  \le \lceil(\thp_* - \thp_0)/(\eta\, m_\ell)\rceil < +\infty$.

 \item Amplitude at handoff: $\|\tho_{t_{\mathrm{stab}}}\| \leq r$

\item \emph{Overshoot control:}\;
  $0 \le \thp_{t_{\mathrm{stab}}} - \thp_*
  \le \eta\, G_\parallel$.

\end{enumerate}
\end{theorem}

\begin{proof} By admissibility, all assumptions needed below hold on the oscillation slab $ \mathcal T_{\mathrm{pre}}$ defined above.
In particular, Assumption~\ref{asmp:RC} gives an exact
multiplicative representation of the orthogonal update:
\[
  \tho_{t+1}=(I-\eta M_t)\tho_t,
\]
where
\begin{equation}\label{eq:proof_M_bounds}
  (1-\delta_r)H(\thp_t)
  \preceq M_t
  \preceq (1+\delta_r)H(\thp_t)
\end{equation}
whenever \(\theta_t\in\mathcal T_{\mathrm{pre}}\).

The proof is a bootstrap on the tubular neighborhood constraint. Let
\[
  \tau
  :=
  \inf\bigl\{
    t\ge0:
    \|\tho_t\|>r
    \ \text{or}\
    \thp_t<\thp_0
    \ \text{or}\
    \thp_t\ge\thp_*
  \bigr\}.
\]

We prove that the tubular neighborhood cannot be exited before the  stability threshold is reached.

\smallskip\noindent
\emph{Step 1: Orthogonal energy growth is controlled by curvature.}
Fix \(t<\tau\). Then \(\theta_t\in\mathcal T_{\mathrm{pre}}\) and
\(\thp_t<\thp_*\). Using the exact update,

\begin{align}
  \|\tho_{t+1}\|^2-\|\tho_t\|^2
  &=
  \|(I-\eta M_t)\tho_t\|^2-\|\tho_t\|^2  \\
  &=
  -2\eta\,{\tho_t}^{\top} M_t\tho_t
  +\eta^2\,{\tho_t}^{\top} M_t^2\tho_t .
\end{align}

Since \(M_t\succeq 0\) and
\begin{align}
    M_t
  \preceq (1+\delta_r)H(\thp_t)
  \preceq (1+\delta_r)\lambda_1(\thp_t) I_U,
\end{align}
we have
\begin{align}
    M_t^2
  \preceq
  (1+\delta_r)\lambda_1(\thp_t) M_t .
\end{align}
Therefore
\begin{align}
     \|\tho_{t+1}\|^2-\|\tho_t\|^2
  \le
  \eta\bigl(\eta(1+\delta_r)\lambda_1(\thp_t)-2\bigr)
  {\tho_t}^{\top} M_t\tho_t .
\end{align}
Using \(M_t\preceq (1+\delta_r)H(\thp_t)\), we obtain
\begin{align}
     \|\tho_{t+1}\|^2-\|\tho_t\|^2
  \le
  \eta(1+\delta_r)
  \bigl(\eta(1+\delta_r)\lambda_1(\thp_t)-2\bigr)
  Q_t .
\end{align}
Since \(\lambda_1\) is nonincreasing and \(\thp_t\ge \thp_0\),
\begin{align}
     \eta(1+\delta_r)\lambda_1(\thp_t)-2
  \le
  \eta(1+\delta_r)\lambda_1(\thp_0)-2.
\end{align}
Thus, with
\[
  C_{\mathrm{unst}}
  :=
  \eta(1+\delta_r)
  \bigl(\eta(1+\delta_r)\lambda_1(\thp_0)-2\bigr),
\]
we have the energy-growth bound
\begin{equation}\label{eq:proof_energy_growth}
  \|\tho_{t+1}\|^2-\|\tho_t\|^2
  \le
  C_{\mathrm{unst}}\,Q_t ,
\end{equation}
where $ Q_t := {\tho_t}^{\top} H(\thp_t)\tho_t.$

\smallskip\noindent
\emph{Step 2: Parallel drift pays for orthogonal growth.}
By the parallel Taylor expansion, we have
\begin{align}
    \langle \nabla\Loss^\parallel(\bth_t), \vv \rangle
  =
  \ell'(\thp_t)
  +\frac{1}{2} {\tho_t}^\top H'(\thp_t)\tho_t
  +R_t^\parallel,
\end{align}
with
\[
  |R_t^\parallel|
  \le c_{\parallel,r} \|\tho_t\|^3 .
\]
Hence the parallel GD update satisfies
\[
\begin{aligned}
  \thp_{t+1}-\thp_t
  &=
  -\eta \langle \nabla\Loss^\parallel(\bth_t), \vv \rangle  \\
  &=
  -\eta\ell'(\thp_t)
  -\frac{\eta}{2}{\tho_t}^{\top} H'(\thp_t)\tho_t
  -\eta R_t^\parallel .
\end{aligned}
\]
By Assumptions~\ref{asmp:LCC} and \ref{asmp:LCD},
\[
  -\ell'(\thp_t)\ge m_\ell,
  \qquad
  -H'(\thp_t)\succeq \beta H(\thp_t),
\]
 therefore
\begin{align}
     \thp_{t+1}-\thp_t
  \ge
  \eta m_\ell
  +\frac{\eta\beta}{2}Q_t
  -\eta c_{\parallel,r}\|\tho_t\|^3 .
\end{align}

In the Admissibility definition~\ref{def:admissibility}, the margin \(\chi>0\) is chosen precisely so that the cubic
remainder can be absorbed into the curvature-reduction term. More
explicitly, recall the compact-slab conditioning bound~\eqref{eq:pre_cond}
\[
  H(\thp_t)\succeq \rho\,\lambda_1(\thp_t) I_U
\]
on the oscillation slab. Since \(t<\tau\), we also have
\[
  \eta(1+\delta_r)\lambda_1(\thp_t)>2.
\]
Thus
\begin{align}
    Q_t
  =
  {\tho_t}^{\top} H(\thp_t)\tho_t
  \ge
  \rho\,\lambda_1(\thp_t)\|\tho_t\|^2
  \ge
  \frac{2\rho}{\eta(1+\delta_r)}\|\tho_t\|^2 .
\end{align}
Since \(\|\tho_t\|\le r\), this implies
\begin{align}
     \|\tho_t\|^3
  \le
  \frac{\eta(1+\delta_r)r}{2\rho}Q_t.
\end{align}
Consequently,
\begin{align}
  \thp_{t+1}-\thp_t
  &\ge
  \eta m_\ell
  +\frac{\eta}{2}
  \left(
    \beta-\frac{\eta(1+\delta_r)c_{\parallel,r} r}{\rho}
  \right)Q_t .
\end{align}

The quantity in parentheses is precisely \(\chi>0\). Therefore
\begin{equation}\label{eq:proof_drift_lower}
  \thp_{t+1}-\thp_t
  \ge
  \eta m_\ell
  +\frac{\eta\chi}{2}Q_t
  \ge
  \eta m_\ell>0 .
\end{equation}
This proves the monotone forward drift as long as the bootstrap
hypotheses hold.

Moreover, rearranging \eqref{eq:proof_drift_lower} gives
\begin{equation}\label{eq:proof_Q_paid}
  Q_t
  \le
  \frac{2}{\eta\chi}
  (\thp_{t+1}-\thp_t).
\end{equation}
Combining \eqref{eq:proof_energy_growth} and
\eqref{eq:proof_Q_paid}, we obtain the key self-stabilization inequality
\begin{equation}\label{eq:proof_paid_by_drift}
  \|\tho_{t+1}\|^2-\|\tho_t\|^2
  \le
  \frac{2C_{\mathrm{unst}}}{\eta\chi}
  (\thp_{t+1}-\thp_t).
\end{equation}
\smallskip\noindent
\emph{Step 3: Closure at the stopping time.}
For every finite integer $t\le\tau$, summing
\eqref{eq:proof_paid_by_drift} gives
\begin{equation}\label{eq:proof_energy_telescope}
  \|\tho_t\|^2
  \le
  \|\tho_0\|^2
  +
  \frac{2C_{\mathrm{unst}}}{\eta\chi}
  (\thp_t-\thp_0).
\end{equation}
Also, \eqref{eq:proof_drift_lower} gives
\[
  \thp_t\ge\thp_0+t\eta m_\ell
  \qquad
  \text{for every finite }t\le\tau.
\]
If $\tau=\infty$, this contradicts $\thp_t<\thp_*$
for all $t$. Hence $\tau<\infty$.

Admissibility implies $\tau\ge1$. Since the source iterate
$\bth_{\tau-1}$ belongs to $\mathcal T_{\mathrm{pre}}$,
\[
  \thp_\tau
  \le \thp_{\tau-1}+\eta G_\parallel
  <\thp_*+\eta G_\parallel.
\]
Consequently,
\[
  \|\tho_\tau\|^2
  \le
  \|\tho_0\|^2
  +
  \frac{2C_{\mathrm{unst}}}{\eta\chi}
  (\thp_*-\thp_0+\eta G_\parallel)
  \le r^2.
\]
Furthermore, positive forward drift gives
$\thp_\tau\ge\thp_0$. Thus neither tubular neighborhood exit nor a decrease
below $\thp_0$ can trigger $\tau$. Necessarily
$\thp_\tau\ge\thp_*$, so $\tau=t_{\mathrm{stab}}$.

This proves tubular neighborhood invariance through the handoff, finite entry,
and the overshoot bound
\[
  0\le
  \thp_{t_{\mathrm{stab}}}-\thp_*
  \le\eta G_\parallel.
\]

\smallskip\noindent
\emph{Step 4: Exponential sharpness decay.}
By the local curvature decay assumption on the oscillation slab,
\[
  -H'(\thp)\succeq \beta H(\thp)
  \qquad
  \forall\,\thp\in[\thp_0,\thp_*].
\]
Equivalently,
\[
  H'(\thp)+\beta H(\thp)\preceq 0.
\]
Therefore
\[
  \frac{d}{d\thp}\left(e^{\beta\thp}H(\thp)\right)
  =
  e^{\beta\thp}\bigl(H'(\thp)+\beta H(\thp)\bigr)
  \preceq 0.
\]
Hence, for every $\thp\in[\thp_0,\thp_*]$,
\[
  e^{\beta\thp}H(\thp)
  \preceq
  e^{\beta\thp_0}H(\thp_0),
\]
or equivalently,
\[
  H(\thp)
  \preceq
  e^{-\beta(\thp-\thp_0)}H(\thp_0).
\]
Taking largest eigenvalues gives
\[
  \lambda_1(\thp)
  \le
  e^{-\beta(\thp-\thp_0)}
  \lambda_1(\thp_0).
\]
Applying this with $\thp=\thp_t$ and using the monotone drift bound
\[
  \thp_t-\thp_0
  =
  \sum_{s=0}^{t-1}(\thp_{s+1}-\thp_s)
  \ge
  \eta m_\ell t
\]
for $0\le t<t_{\mathrm{stab}}$, we obtain
\[
  \lambda_1(\thp_t)
  \le
  \lambda_1(\thp_0)e^{-\beta\eta m_\ell t}.
\]
\smallskip\noindent
\emph{Step 5: Finite entry time.}
By \eqref{eq:proof_drift_lower}, for every \(0\le t<t_{\mathrm{stab}}\),
\[
  \thp_{t+1}-\thp_t\ge \eta m_\ell .
\]
Thus, for every \(n<t_{\mathrm{stab}}\),
\[
  \thp_n
  \ge
  \thp_0+n\eta m_\ell .
\]
Consequently, once
\[
  n\ge \frac{\thp_*-\thp_0}{\eta m_\ell},
\]
we must have \(\thp_n\ge \thp_*\). Therefore
\[
  t_{\mathrm{stab}}
  \le
  \left\lceil
  \frac{\thp_*-\thp_0}{\eta m_\ell}
  \right\rceil .
\]

\smallskip\noindent
\emph{Step 6: Orthogonal amplitude at handoff.}
The stopping-time argument already establishes
\[
  \|\tho_{t_{\mathrm{stab}}}\|^2
  \le
  \|\tho_0\|^2
  +
  \frac{2C_{\mathrm{unst}}}{\eta\chi}
  (\thp_*-\thp_0+\eta G_\parallel)
  \le r^2.
\]

\smallskip\noindent
\emph{Step 7: Corrected stability at handoff.}
By definition,
\[
  \thp_*
  =
  \inf\{\thp\ge \thp_0:
  \eta(1+\delta_r)\lambda_1(\thp)\le 2\}.
\]
Since \(\lambda_1\) is nonincreasing and
\(\thp_{t_{\mathrm{stab}}}\ge \thp_*\), we conclude that
\[
  \eta(1+\delta_r)\lambda_1(\thp_{t_{\mathrm{stab}}})\le 2.
\]
This proves all claims.
\end{proof}

\subsection{Stable phase: schedules and three-phase damping}
\label{app:stable_phase}

We now analyze the trajectory for $t \ge t_{\mathrm{stab}}$. Throughout this subsection, assume
$\thp_*\ge\thp_{\mathrm{tail}}(r)$.
The re-anchored tail bounds below therefore hold for every
$\thp\ge\thp_*$.

\myparagraph{Re-anchored tail constants}
Assumption~\ref{asmp:ET} gives exponential-tail bounds centered at
$\thp_{\mathrm{tail}}(r)$.  In the stable-phase proof it is more convenient
to center the same bounds at the stability threshold $\thp_*$.  We
therefore use the following re-anchored constants.  If
\[
  A_\pm^{\mathrm{tail}},\qquad
  \lambda_{1,\pm}^{\mathrm{tail}},
  \qquad
  \beta_{\mathrm{tail}}
\]
denote constants for which
\[
  A_-^{\mathrm{tail}}
  e^{-\beta_{\mathrm{tail}}(\thp-\thp_{\mathrm{tail}})}
  \le
  -\langle \nabla\Loss^\parallel(\bth),v\rangle
  \le
  A_+^{\mathrm{tail}}
  e^{-\beta_{\mathrm{tail}}(\thp-\thp_{\mathrm{tail}})}
\]
and
\[
  \lambda_{1,-}^{\mathrm{tail}}
  e^{-\beta_{\mathrm{tail}}(\thp-\thp_{\mathrm{tail}})}
  \le
  \lambda_1(\thp)
  \le
  \lambda_{1,+}^{\mathrm{tail}}
  e^{-\beta_{\mathrm{tail}}(\thp-\thp_{\mathrm{tail}})}
\]
for $\thp\ge \thp_{\mathrm{tail}}(r)$ and $\|\tho\|\le r$, then define
\begin{equation}\label{eq:reanchored_tail_constants}
  A_\pm^*
  :=
  A_\pm^{\mathrm{tail}}\,
  e^{-\beta_{\mathrm{tail}}(\thp_*-\thp_{\mathrm{tail}})},
  \qquad
  \lambda_{1,\pm}^*
  :=
  \lambda_{1,\pm}^{\mathrm{tail}}\,
  e^{-\beta_{\mathrm{tail}}(\thp_*-\thp_{\mathrm{tail}})}.
\end{equation}
Equivalently, for all $\thp\ge \thp_*$ and $\|\tho\|\le r$,
\begin{align}
  A_-^* e^{-\beta_{\mathrm{tail}}(\thp-\thp_*)}
  &\le
  -\langle \nabla\Loss^\parallel(\bth),v\rangle
  \le
  A_+^* e^{-\beta_{\mathrm{tail}}(\thp-\thp_*)},
  \label{eq:anchored_tail_drift}
  \\
  \lambda_{1,-}^* e^{-\beta_{\mathrm{tail}}(\thp-\thp_*)}
  &\le
  \lambda_1(\thp)
  \le
  \lambda_{1,+}^* e^{-\beta_{\mathrm{tail}}(\thp-\thp_*)}.
  \label{eq:anchored_tail_sharpness}
\end{align}
Throughout the stable-phase proof, $A_\pm^*$ and
$\lambda_{1,\pm}^*$ refer to these re-anchored constants.
Define
\[
  \Psi(B)
  :=
  \begin{cases}
  \dfrac{e^B-1}{B}, & B>0,\\[2mm]
  1, & B=0,
  \end{cases}
  \qquad
  B_r
  :=
  \frac{2\beta_{\mathrm{tail}}}{1+\delta_r}
  \frac{A_+^*}{\lambda_{1,-}^*}.
\]
\[
  \bar{y}
  =
  e^{\beta_{\mathrm{tail}}(\thp_{t_{\mathrm{stab}}}-\thp_*)},
  \qquad
  \alpha_-
  =
  \beta_{\mathrm{tail}}\eta A_-^*,
  \qquad
  \alpha_+
  =
  \beta_{\mathrm{tail}}\eta A_+^*\,\Psi(B_r).
\]

\begin{theorem}[Stable phase]
\label{thm:stable_phase}
Under the hypotheses of Theorem~\ref{thm:eos_phase}, assume
additionally that
$\thp_* \ge \thp_{\mathrm{tail}}(r)$.
Then for all $t \ge t_{\mathrm{stab}}$:

\begin{enumerate}[label=\textup{(\roman*)},leftmargin=2.4em]

\item \emph{Nonexpansion:}\;
  $\|\tho_t\| \le \|\tho_{t_{\mathrm{stab}}}\| \le r$.

\item \emph{Logarithmic parallel growth:}\;
  $$\displaystyle
  \frac{1}{\beta_{\mathrm{tail}}}
    \log\bigl(\bar{y} + \alpha_-(t - t_{\mathrm{stab}})\bigr)
  \le \thp_t - \thp_*
  \le \frac{1}{\beta_{\mathrm{tail}}}
    \log\bigl(\bar{y} + \alpha_+(t - t_{\mathrm{stab}})\bigr).
  $$

\item \emph{Sharpness schedule:}\;
  $\displaystyle
  \frac{\lambda_{1,-}^*}
       {\bar{y} + \alpha_+(t - t_{\mathrm{stab}})}
  \le \lambda_1(\thp_t)
  \le \frac{\lambda_{1,+}^*}
       {\bar{y} + \alpha_-(t - t_{\mathrm{stab}})}.
  $

\end{enumerate}

\medskip
\noindent
Assume further the \emph{uniform conditioning}
$H(\thp) \succeq \rho_{\mathrm{tail}}\,\lambda_1(\thp)\,I_U$
for all $\thp \ge \thp_*$.
Fix $0 < \varepsilon \le \varepsilon_*
:= \rho_{\mathrm{tail}}(1-\delta_r)/(1+\delta_r)$,
and define
\begin{equation}\label{eq:transition_times}
  t_{\mathrm{lin}}
  := \inf\bigl\{t \ge t_{\mathrm{stab}} :
     \eta(1+\delta_r)\lambda_1(\thp_t) \le 2 - \varepsilon\bigr\},
  \qquad
  t_{\mathrm{poly}}
  := \inf\bigl\{t \ge t_{\mathrm{stab}} :
     \eta(1+\delta_r)\lambda_1(\thp_t) \le 1\bigr\}.
\end{equation}
Then $t_{\mathrm{stab}} \le t_{\mathrm{lin}}
\le t_{\mathrm{poly}} < \infty$, and:

\begin{enumerate}[label=\textup{(\roman*)},leftmargin=2.4em,
                  resume]

\item \emph{Transition phase}
  $(t_{\mathrm{stab}} \le t \le t_{\mathrm{lin}})$:\;
  $\|\tho_t\| \le \|\tho_{t_{\mathrm{stab}}}\|$.

\item \emph{Linear contraction phase}
  $(t_{\mathrm{lin}} \le t \le t_{\mathrm{poly}})$:\;
  $\|\tho_t\|
  \le (1 - \varepsilon)^{t - t_{\mathrm{lin}}}
      \,\|\tho_{t_{\mathrm{lin}}}\|$. At $t=t_{\mathrm{lin}}$, the bound is interpreted as equality, including when $\varepsilon=1$.

\item \emph{Polynomial decay phase}
  $(t \ge t_{\mathrm{poly}})$:\;
  with $A := \lceil \bar{y}/\alpha_+ \rceil + 1$ and
  \[
  \gamma_{r}
  :=
  \frac{
    \rho_{\mathrm{tail}}(1-\delta_r)
  }{
    \beta_{\mathrm{tail}}\Psi(B_r)
  }
  \frac{\lambda_{1,-}^*}{A_+^*},
\]
  \[
    \|\tho_t\|
    \le \|\tho_{t_{\mathrm{poly}}}\|
        \left(
          \frac{A + t_{\mathrm{poly}} - t_{\mathrm{stab}}}
               {A + t - t_{\mathrm{stab}}}
        \right)^{\!\gamma_r}.
  \]

\end{enumerate}
\end{theorem}

\begin{proof}Since \(\thp_{t_{\mathrm{stab}}}\ge \thp_{\mathrm{tail}}(r)\), the exponential-tail
assumptions apply from time \(t_{\mathrm{stab}}\) onward as long as the iterates remain
inside the tubular neighborhood. We first prove nonexpansion, which closes this bootstrap.

\smallskip\noindent
\emph{Step 1: Nonexpansion.}
By Assumptions~\ref{asmp:RC} and \ref{asmp:SA}, the
orthogonal update admits the exact multiplicative representation
\[
  \tho_{t+1}=(I-\eta M_t)\tho_t,
\]
where
\begin{equation}\label{eq:post_M_bounds}
  (1-\delta_r)H(\thp_t)
  \preceq M_t
  \preceq (1+\delta_r)H(\thp_t)
\end{equation}
whenever \(\|\tho_t\|\le r\).

We claim that
\[
  \|\tho_t\|\le \|\tho_{t_{\mathrm{stab}}}\|
  \qquad\forall\,t\ge t_{\mathrm{stab}}.
\]
Indeed, suppose inductively that \(\|\tho_t\|\le \|\tho_{t_{\mathrm{stab}}}\|\). Since the tail
parallel drift satisfies
\[
  -\langle \nabla\Loss^\parallel(\bth_t), \vv \rangle>0,
\]
we have
\[
  \thp_{t+1}-\thp_t
  =
  -\eta\langle \nabla\Loss^\parallel(\bth_t), \vv \rangle>0.
\]
Thus \(\thp_t\) is nondecreasing for \(t\ge t_{\mathrm{stab}}\). Since
\(\lambda_1\) is nonincreasing,
\[
  \eta(1+\delta_r)\lambda_1(\thp_t)
  \le
  \eta(1+\delta_r)\lambda_1(\thp_{t_{\mathrm{stab}}})
  \le 2 .
\]
Using \eqref{eq:post_M_bounds},
\[
  0\preceq M_t
  \preceq (1+\delta_r)\lambda_1(\thp_t)I_U,
\]
so every eigenvalue \(m\) of \(M_t\) satisfies
\[
  0\le \eta m\le 2.
\]
Therefore
\[
  \|I-\eta M_t\|_{\mathrm{op}}\le 1,
\]
and hence
\[
  \|\tho_{t+1}\|
  \le
  \|\tho_t\|.
\]
The induction allows us to conclude that
\[
  \|\tho_t\|
  \le
  \|\tho_{t_{\mathrm{stab}}}\|
  \qquad
  \forall\,t\ge t_{\mathrm{stab}}.
\]
Thus the tail assumptions apply globally for all \(t\ge t_{\mathrm{stab}}\).

\smallskip\noindent
\emph{Step 2: Parallel recursion.}
By the re-anchored exponential-tail drift bound
\eqref{eq:anchored_tail_drift} and the GD update,
\[
  \thp_{t+1}-\thp_t
  =
  -\eta\,\langle \nabla\Loss^\parallel(\bth_t), \vv \rangle,
\]
we have
\[
  \eta A_-^*e^{-\beta_{\mathrm{tail}}(\thp_t-\thp_*)}
  \le
  \thp_{t+1}-\thp_t
  \le
  \eta A_+^*e^{-\beta_{\mathrm{tail}}(\thp_t-\thp_*)}.
\]
Equivalently, since
\[
  y_t:=e^{\beta_{\mathrm{tail}}(\thp_t-\thp_*)},
\]
\[
  \frac{\eta A_-^*}{y_t}
  \le
  \thp_{t+1}-\thp_t
  \le
  \frac{\eta A_+^*}{y_t}.
\]
We write
\[
  y_{t+1}
  =
  y_t
  e^{
    \beta_{\mathrm{tail}}(\thp_{t+1}-\thp_t)}
  .
\]
For the lower bound, using \(e^z\ge 1+z\),
\[
  y_{t+1}
  \ge
  y_t
  \left(
    1+\beta_{\mathrm{tail}}(\thp_{t+1}-\thp_t)
  \right)
  \ge
  y_t+\beta_{\mathrm{tail}}\eta A_-^*.
\]
Thus, with
\[
  \alpha_-:=\beta_{\mathrm{tail}}\eta A_-^*,
\]
we get
\[
  y_t\ge \bar y+\alpha_-(t-t_{\mathrm{stab}}).
\]
For the upper bound, set
\[
  z_t
  :=
  \beta_{\mathrm{tail}}(\thp_{t+1}-\thp_t).
\]
The re-centered tail drift bound gives
\[
  0\le z_t
  \le
  \frac{\beta_{\mathrm{tail}}\eta A_+^*}{y_t}.
\]
Since $y_t\ge 1$, it remains to bound
$\beta_{\mathrm{tail}}\eta A_+^*$. Evaluating the re-centered tail
sharpness lower bound at $\thp=\thp_*$ gives
\[
  \lambda_{1,-}^*\le \lambda_1(\thp_*).
\]
By the definition of the stability threshold,
\[
  \eta(1+\delta_r)\lambda_1(\thp_*)\le 2.
\]
Therefore
\[
  \beta_{\mathrm{tail}}\eta A_+^*
  =
  \beta_{\mathrm{tail}}\frac{A_+^*}{\lambda_{1,-}^*}
  \eta\lambda_{1,-}^*
  \le
  \frac{2\beta_{\mathrm{tail}}}{1+\delta_r}
  \frac{A_+^*}{\lambda_{1,-}^*}
  =
  B_r.
\]
Hence $z_t\in[0,B_r]$. By definition of $\Psi$,
\[
  e^{z_t}-1\le z_t\Psi(B_r).
\]
Thus
\[
\begin{aligned}
  y_{t+1}-y_t
  &=
  y_t(e^{z_t}-1)  \\
  &\le
  y_t z_t\Psi(B_r)  \\
  &\le
  \beta_{\mathrm{tail}}\eta A_+^*\,\Psi(B_r)
  =
  \alpha_+.
\end{aligned}
\]
Therefore
\[
  y_t\le \bar y+\alpha_+(t-t_{\mathrm{stab}}).
\]

Combining the two bounds gives
\begin{equation}\label{eq:post_y_schedule}
  \bar y+\alpha_-(t-t_{\mathrm{stab}})
  \le
  y_t
  \le
  \bar y+\alpha_+(t-t_{\mathrm{stab}}).
\end{equation}
Taking logarithms and dividing by \(\beta_{\mathrm{tail}}\) yields
\[
  \frac{1}{\beta_{\mathrm{tail}}}
    \log\bigl(\bar y+\alpha_-(t-t_{\mathrm{stab}})\bigr)
  \le
  \thp_t-\thp_*
  \le
  \frac{1}{\beta_{\mathrm{tail}}}
    \log\bigl(\bar y+\alpha_+(t-t_{\mathrm{stab}})\bigr).
\]
This proves the logarithmic parallel growth.

\smallskip\noindent
\emph{Step 3: sharpness schedule.}
The re-anchored exponential-tail sharpness bounds
\eqref{eq:anchored_tail_sharpness} give
\[
  \lambda_{1,-}^*e^{-\beta_{\mathrm{tail}}(\thp_t-\thp_*)}
  \le
  \lambda_1(\thp_t)
  \le
  \lambda_{1,+}^*e^{-\beta_{\mathrm{tail}}(\thp_t-\thp_*)}.
\]

Since \(y_t=e^{\beta_{\mathrm{tail}}(\thp_t-\thp_*)}\), this is
\[
  \frac{\lambda_{1,-}^*}{y_t}
  \le
  \lambda_1(\thp_t)
  \le
  \frac{\lambda_{1,+}^*}{y_t}.
\]
Using \eqref{eq:post_y_schedule}, we obtain
\[
  \frac{\lambda_{1,-}^*}
       {\bar y+\alpha_+(t-t_{\mathrm{stab}})}
  \le
  \lambda_1(\thp_t)
  \le
  \frac{\lambda_{1,+}^*}
       {\bar y+\alpha_-(t-t_{\mathrm{stab}})}.
\]
This proves the sharpness schedule.

\smallskip\noindent
\emph{Step 4: Transition times are finite and ordered.}
Because \(y_t\to\infty\) at least linearly by
\eqref{eq:post_y_schedule}, the sharpness upper bound implies
\[
  \lambda_1(\thp_t)
  \le
  \frac{\lambda_{1,+}^*}{\bar y+\alpha_-(t-t_{\mathrm{stab}})}
  \longrightarrow 0.
\]
Hence both thresholds
\[
  \eta(1+\delta_r)\lambda_1(\thp_t)\le 2-\varepsilon
  \qquad\text{and}\qquad
  \eta(1+\delta_r)\lambda_1(\thp_t)\le 1
\]
are reached in finite time. Since \(2-\varepsilon\ge 1\) for
\(0<\varepsilon\le 1\), we have
\[
  t_{\mathrm{stab}}\le t_{\mathrm{lin}}\le t_{\mathrm{poly}}<\infty.
\]

\smallskip\noindent
\emph{Step 5: Transition phase.}
For \(t_{\mathrm{stab}}\le t\le t_{\mathrm{lin}}\), nonexpansion from Step~1 already gives
\[
  \|\tho_t\|
  \le
  \|\tho_{t_{\mathrm{stab}}}\|.
\]
This proves the transition-phase bound.

\smallskip\noindent
\emph{Step 6: Linear contraction phase.}
Assume now that \(t_{\mathrm{lin}}\le t<t_{\mathrm{poly}}\). By definition
of \(t_{\mathrm{lin}}\),
\[
  \eta(1+\delta_r)\lambda_1(\thp_t)\le 2-\varepsilon.
\]
By definition of \(t_{\mathrm{poly}}\), for \(t<t_{\mathrm{poly}}\),
\[
  \eta(1+\delta_r)\lambda_1(\thp_t)>1.
\]
Using tail conditioning and relative curvature comparability,
\[
  M_t
  \succeq
  (1-\delta_r)H(\thp_t)
  \succeq
  \rho_{\mathrm{tail}}(1-\delta_r)\lambda_1(\thp_t)I_U.
\]
Therefore every eigenvalue \(m\) of \(M_t\) satisfies
\begin{align*}
    \eta m
  \ge
  \eta\rho_{\mathrm{tail}}(1-\delta_r)\lambda_1(\thp_t)
  &=
  \frac{\rho_{\mathrm{tail}}(1-\delta_r)}{1+\delta_r}
  \eta(1+\delta_r)\lambda_1(\thp_t) \\
  &> \frac{\rho_{\mathrm{tail}}(1-\delta_r)}{1+\delta_r}= \varepsilon_* .
\end{align*}
Since \(\varepsilon\le \varepsilon_*\), we have
\[
  \eta m\ge \varepsilon.
\]
The upper bound gives
\[
  \eta m
  \le
  \eta(1+\delta_r)\lambda_1(\thp_t)
  \le
  2-\varepsilon.
\]
Thus all eigenvalues of \(\eta M_t\) lie in
\[
  [\varepsilon,\,2-\varepsilon].
\]
Hence
\[
  \|I-\eta M_t\|_{\mathrm{op}}\le 1-\varepsilon.
\]
Therefore
\[
  \|\tho_{t+1}\|
  \le
  (1-\varepsilon)\|\tho_t\|.
\]
Iterating from \(t_{\mathrm{lin}}\) to \(t\) yields
\[
  \|\tho_t\|
  \le
  (1-\varepsilon)^{t-t_{\mathrm{lin}}}
  \|\tho_{t_{\mathrm{lin}}}\|
  \qquad
  (t_{\mathrm{lin}}\le t\le t_{\mathrm{poly}}).
\]

\smallskip\noindent
\emph{Step 7: Polynomial decay phase.}
For \(t\ge t_{\mathrm{poly}}\), the definition of \(t_{\mathrm{poly}}\)
gives
\[
  \eta(1+\delta_r)\lambda_1(\thp_t)\le 1.
\]
Hence
\[
  0\preceq \eta M_t\preceq I_U.
\]
Moreover, by tail conditioning and relative curvature comparability,
\[
  M_t
  \succeq
  \rho_{\mathrm{tail}}(1-\delta_r)\lambda_1(\thp_t)I_U.
\]
Thus
\[
  \|I-\eta M_t\|_{\mathrm{op}}
  \le
  1-\eta\rho_{\mathrm{tail}}(1-\delta_r)\lambda_1(\thp_t).
\]
Therefore, for \(s\ge t_{\mathrm{poly}}\),
\[
  \|\tho_{s+1}\|
  \le
  \left(
    1-\eta\rho_{\mathrm{tail}}(1-\delta_r)
    \lambda_1(\thp_s)
  \right)\|\tho_s\|.
\]
Iterating from \(s=t_{\mathrm{poly}}\) to \(s=t-1\), and using
\(1-a\le e^{-a}\), gives
\[
  \|\tho_t\|
  \le
  \|\tho_{t_{\mathrm{poly}}}\|
  \exp\!\left(
    -\eta\rho_{\mathrm{tail}}(1-\delta_r)
    \sum_{s=t_{\mathrm{poly}}}^{t-1}
    \lambda_1(\thp_s)
  \right).
\]
From the sharpness lower schedule,
\[
  \lambda_1(\thp_s)
  \ge
  \frac{\lambda_{1,-}^*}
       {\bar y+\alpha_+(s-t_{\mathrm{stab}})}.
\]
Define
\[
  A:=\left\lceil\frac{\bar y}{\alpha_+}\right\rceil+1.
\]
Then
\[
  \bar y\le \alpha_+(A-1),
\]
so
\[
  \bar y+\alpha_+(s-t_{\mathrm{stab}})
  \le
  \alpha_+(A+s-t_{\mathrm{stab}}).
\]
Consequently,
\[
  \lambda_1(\thp_s)
  \ge
  \frac{\lambda_{1,-}^*}
       {\alpha_+(A+s-t_{\mathrm{stab}})}.
\]
Therefore
\[
\begin{aligned}
  \sum_{s=t_{\mathrm{poly}}}^{t-1}\lambda_1(\thp_s)
  &\ge
  \frac{\lambda_{1,-}^*}{\alpha_+}
  \sum_{s=t_{\mathrm{poly}}}^{t-1}
  \frac{1}{A+s-t_{\mathrm{stab}}}  \\
  &\ge
  \frac{\lambda_{1,-}^*}{\alpha_+}
  \log\!\left(
    \frac{A+t-t_{\mathrm{stab}}}{A+t_{\mathrm{poly}}-t_{\mathrm{stab}}}
  \right).
\end{aligned}
\]
Substituting into the exponential estimate yields
\[
  \|\tho_t\|
  \le
  \|\tho_{t_{\mathrm{poly}}}\|
  \left(
    \frac{A+t_{\mathrm{poly}}-t_{\mathrm{stab}}}
         {A+t-t_{\mathrm{stab}}}
  \right)^{\gamma_r},
\]
where
\[
  \gamma_r
  :=
  \frac{
    \rho_{\mathrm{tail}}(1-\delta_r)\eta\lambda_{1,-}^*
  }{\alpha_+}.
\]
This is exactly the claimed
polynomial decay estimate.
\end{proof}

\begin{rem}[Recovering Theorem~\ref{thm:main}]
\label{rem:main_relationship}
Theorem~\ref{thm:main} in the main text is the union of
Theorems~\ref{thm:eos_phase} and~\ref{thm:stable_phase}.
\end{rem}

\begin{rem}[Relationship between $\varepsilon$ and $\varepsilon_*$]
\label{rem:eps_flexibility}
The parameter $\varepsilon$ in the linear contraction phase is free
within $(0, \varepsilon_*]$.  Choosing $\varepsilon = \varepsilon_*$
minimizes the contraction factor $1-\varepsilon$ but defines the latest
possible $t_{\mathrm{lin}}$.  Choosing $\varepsilon < \varepsilon_*$
gives a weaker per-step bound but an earlier onset of contraction,
which can produce a tighter overall bound depending on the sharpness
schedule.  The main theorem fixes $\varepsilon = \varepsilon_*$ for
simplicity.
\end{rem}

\subsection{Convergence rates for the loss}\label{app:loss_rates}

\begin{description}[style=nextline,leftmargin=2.8em,labelwidth=2.6em]
\item[\textbf{(LE)}]
\textbf{Exponential loss envelope.}
There exist constants \(B_{\Loss}>0\) such that, on the tubular neighborhood traversed by the
trajectory,
\[
\Loss(\bth)
\le
B_{\Loss}
e^{-\beta_{\mathrm{tail}}(\theta^\parallel-\theta_0^\parallel)}
\qquad
\text{for all }
\theta^\parallel \geq \theta_0^\parallel,
\quad
\|\tho\|\le r .
\]
\end{description}

Condition \textup{(LE)} is an additional assumption on the normalization and decay of the loss itself; it does not follow from derivative bounds alone. It holds for the normalized logistic and multiclass cross-entropy losses considered below.

\begin{cor}[Convergence of the Loss]
\label{cor:loss_decay_eos_and_stable}
Suppose Assumptions~A and~B hold, and suppose moreover that the
exponential loss envelope \textup{(LE)} holds on the slab-tubular neighborhood traversed
by the trajectory. Then the following bounds hold.

\begin{enumerate}[label=\textup{(\roman*)},leftmargin=2.2em]

\item \textbf{Averaged loss during the unstable phase.}
For every integer \(1\le T\le t_{\mathrm{stab}}+1\),
\begin{equation}\label{eq:eos_avg_loss_general}
\frac1T\sum_{t=0}^{T-1}\Loss(\theta_t)
\le
\frac{B_{\Loss}}{T}
\frac{1-e^{-\beta_{\mathrm{tail}}\eta m_\ell T}}
     {1-e^{-\beta_{\mathrm{tail}}\eta m_\ell}}
\le
\frac{B_{\Loss}}
     {T\bigl(1-e^{-\beta_{\mathrm{tail}}\eta m_\ell}\bigr)} .
\end{equation}
In particular,
\[
\frac1T\sum_{t=0}^{T-1}\Loss(\theta_t)
=
\mathcal O(T^{-1})
\qquad
\text{uniformly for }1\le T\le t_{\mathrm{stab}}+1 .
\]

\item \textbf{Last-iterate loss after stabilization.}
For every
\(t\ge t_{\mathrm{stab}}\),
\begin{equation}\label{eq:stable_last_iterate_loss}
\Loss(\theta_t)
\le
\frac{\widetilde B_{\Loss}}
     {\bar y+\alpha_-(t-t_{\mathrm{stab}})}
\le
\frac{\widetilde B_{\Loss}}
     {1+\beta_{\mathrm{tail}}\eta A_-^*(t-t_{\mathrm{stab}})},
\end{equation}
where
\[
\widetilde B_{\Loss}
:=
B_{\Loss}
e^{-\beta_{\mathrm{tail}}(\theta_{*}-\theta_0^\parallel)} ,
\]
and the last inequality uses \(\bar y\ge 1\). Consequently,
\[
\Loss(\theta_t)
=
\mathcal O\bigl((1+t-t_{\mathrm{stab}})^{-1}\bigr).
\]

\end{enumerate}
\end{cor}

\begin{proof}
For part \textup{(i)}, Theorem~\ref{thm:main} implies that throughout
the unstable phase the trajectory remains in the slab-tubular neighborhood and satisfies
\[
\theta_t^\parallel
\ge
\theta_0^\parallel+\eta m_\ell t
\qquad
(0\le t\le t_{\mathrm{stab}}).
\]
Applying the loss envelope \textup{(LE)} at \(\theta_t\) gives
\[
\Loss(\theta_t)
\le
B_{\Loss} e^{-\beta_{\mathrm{tail}}(\theta_t^\parallel-\theta_0^\parallel)}
\le
B_{\Loss} e^{-\beta_{\mathrm{tail}}\eta m_\ell t}.
\]
Therefore, for every \(1\le T\le t_{\mathrm{stab}}+1\),
\[
\frac1T\sum_{t=0}^{T-1}\Loss(\theta_t)
\le
\frac{B_{\Loss}}{T}
\sum_{t=0}^{T-1} e^{-\beta_{\mathrm{tail}}\eta m_\ell t}
=
\frac{B_{\Loss}}{T}
\frac{1-e^{-\beta_{\mathrm{tail}}\eta m_\ell T}}
     {1-e^{-\beta_{\mathrm{tail}}\eta m_\ell}} .
\]
Since \(1-e^{-\beta_{\mathrm{tail}}\eta m_\ell T}\le 1\), this proves
\eqref{eq:eos_avg_loss_general}.

For part \textup{(ii)}, rewrite the same loss envelope relative to
\(\theta_{*}\). For \(t\ge t_{\mathrm{stab}}\), we have
\[
\Loss(\theta_t)
\le
B_{\Loss}
e^{-\beta_{\mathrm{tail}}(\theta_t^\parallel-\theta_0^\parallel)}.
\]
Splitting the exponent gives
\[
\theta_t^\parallel-\theta_0^\parallel
=
(\theta_t^\parallel-\theta_{*})
+
(\theta_{*}-\theta_0^\parallel),
\]
and hence
\[
\Loss(\theta_t)
\le
B_{\Loss}
e^{-\beta_{\mathrm{tail}}(\theta_{*}-\theta_0^\parallel)}
e^{-\beta_{\mathrm{tail}}(\theta_t^\parallel-\theta_{*})}.
\]
By definition,
\[
\widetilde B_{\Loss}
:=
B_{\Loss}
e^{-\beta_{\mathrm{tail}}(\theta_{*}-\theta_0^\parallel)}
\]
and
\[
y_t
:=
e^{\beta_{\mathrm{tail}}(\theta_t^\parallel-\theta_{*})}.
\]
Therefore,
\[
\Loss(\theta_t)
\le
\frac{\widetilde B_{\Loss}}{y_t}.
\]
Using the previous inequality (from the stable phase proof)
\[
y_t
\ge
\bar y+\alpha_-(t-t_{\mathrm{stab}})
\qquad
(t\ge t_{\mathrm{stab}}),
\]
we obtain
\[
\Loss(\theta_t)
\le
\frac{\widetilde B_{\Loss}}
     {\bar y+\alpha_-(t-t_{\mathrm{stab}})}.
\]
In the stable phase \(\bar y\ge 1\), which further implies
\[
\Loss(\theta_t)
\le
\frac{\widetilde B_{\Loss}}
     {1+\alpha_-(t-t_{\mathrm{stab}})}.
\]
This proves \eqref{eq:stable_last_iterate_loss}.
\end{proof}

\begin{rem}[Comparison with sharper logistic loss-rate estimates]
\label{rem:wu_loss_rate_comparison}
The averaged unstable phase bound in Corollary~\ref{cor:loss_decay_eos_and_stable}
is intentionally stated under a minimal loss-envelope assumption. It only
uses the facts that the trajectory remains in the tubular neighborhood and that the
parallel coordinate makes uniform forward progress. Consequently, the
proof bounds the loss by a geometric envelope along the trajectory,
leading to
\[
\frac1T\sum_{t=0}^{T-1}\Loss(\theta_t)
\le
\frac{B_{\Loss}}
     {T\bigl(1-e^{-\beta_{\mathrm{tail}}\eta m_\ell}\bigr)}.
\]
This estimate is robust but not sharp in the large-step regime.

Sharper rates, such as the logarithmic averaged rates
obtained in logistic-tail analyses of \cite{wu2024large, crawshaw2026}, use additional structure: the loss
must be comparable to the parallel drift throughout the relevant part of
the oscillation tubular neighborhood. Namely, if on this portion of the tubular neighborhood one has
\[
\Loss(\theta)
\le
C_{\Loss}\bigl[-\langle \nabla \Loss(\theta),v\rangle\bigr],
\qquad
-\langle \nabla \Loss(\theta),v\rangle
\le
G_{\parallel}
e^{-\beta_{\mathrm{tail}}(\theta^\parallel-\theta_0^\parallel)},
\]
then the GD update gives the exact telescoping relation
\[
\Loss(\theta_t)
\le
\frac{C_{\Loss}}{\eta}
(\theta_{t+1}^\parallel-\theta_t^\parallel).
\]
Summing over \(t=0,\dots,T-1\) yields
\[
\frac1T\sum_{t=0}^{T-1}\Loss(\theta_t)
\le
\frac{C_{\Loss}}{\eta T}
(\theta_T^\parallel-\theta_0^\parallel).
\]
The second tail bound implies
\[
\theta_T^\parallel-\theta_0^\parallel
\lesssim
\frac{1}{\beta_{\mathrm{tail}}}
\log(1+\eta T),
\]
and hence
\[
\frac1T\sum_{t=0}^{T-1}\Loss(\theta_t)
\lesssim
\frac{\log(1+\eta T)}{\eta T}.
\]
Thus the corollary should be viewed as a general envelope estimate,
whereas the sharper Wu-type rate requires a loss--parallel-gradient
comparability condition on the unstable portion of the tubular neighborhood. Logistic and
cross-entropy losses satisfy this additional tail structure once the
trajectory lies in the corresponding exponential-tail region.
\end{rem}

\section{Logistic Regression}\label{app:logistic_verification}

\subsection{Preliminaries}

\begin{prop}[Full Hessian and its restriction to the parallel direction]
\label{prop:hd_full_Hessian}
Consider the high-dimensional symmetric logistic loss
\begin{equation}\label{eq:hd_loss_fullspace}
\Loss(\theta)
=
\frac{1}{2N}
\sum_{i=1}^N
\Bigl[
\phi\!\bigl(\inner{\theta}{x_i}\bigr)
+
\phi\!\bigl(-\inner{\theta}{\vx_i'}\bigr)
\Bigr],
\qquad
\vx_i=a_i v+\vu_i,\quad \vx_i'=-a_i v+\vu_i,
\end{equation}
where \(v\in\mathbb R^d\) is a unit vector,
\[
\vu_i\in v^\perp\setminus\{0\},
\qquad
a_i>0,
\qquad
i=1,\dots,N.
\]
Then the full Hessian of \(\Loss\) at an arbitrary point
\(\theta\in\mathbb R^d\) is
\begin{equation}\label{eq:hd_full_Hessian_formula}
\nabla^2\Loss(\theta)
=
\frac{1}{2N}
\sum_{i=1}^N
\Bigl[
\phi''\!\bigl(\inner{\theta}{\vx_i}\bigr)\,\vx_i \vx_i^\top
+
\phi''\!\bigl(-\inner{\theta}{\vx_i'}\bigr)\,\vx_i'(\vx_i')^\top
\Bigr].
\end{equation}
In particular, along the parallel direction \(\theta=\theta^\parallel v\),
\begin{equation}\label{eq:hd_Hessian_on_valley}
\nabla^2\Loss(\theta^\parallel v)
=
\frac{1}{N}
\sum_{i=1}^N
\phi''(a_i\theta^\parallel)
\bigl(
a_i^2\, vv^\top + \vu_i \vu_i^\top
\bigr).
\end{equation}
Therefore, the restriction of the Hessian to the active normal space
\[
U:=\Span\{u_1,\dots,u_N\}\subseteq v^\perp
\]
is
\begin{equation}\label{eq:hd_H_from_full_Hessian}
H(\theta^\parallel)
:=
\Pi_U \nabla^2\Loss(\theta^\parallel v)\Pi_U
=
\frac{1}{N}
\sum_{i=1}^N \phi''(a_i\theta^\parallel)\,\vu_i \vu_i^\top.
\end{equation}
\end{prop}

\begin{proof}
We first compute the gradient on the full space:
\[
\nabla \Loss(\theta)
=
\frac{1}{2N}
\sum_{i=1}^N
\Bigl[
\phi'\!\bigl(\inner{\theta}{\vx_i}\bigr)\vx_i
-
\phi'\!\bigl(-\inner{\theta}{\vx_i'}\bigr)\vx_i'
\Bigr].
\]
Since \(\phi''(-s)=\phi''(s)\), differentiating once more gives
\[
\nabla^2\Loss(\theta)
=
\frac{1}{2N}
\sum_{i=1}^N
\Bigl[
\phi''\!\bigl(\inner{\theta}{\vx_i}\bigr)\vx_i \vx_i^\top
+
\phi''\!\bigl(-\inner{\theta}{\vx_i'}\bigr)\vx_i'(\vx_i')^\top
\Bigr],
\]
which is exactly \eqref{eq:hd_full_Hessian_formula}.

Now evaluate at \(\theta=\theta^\parallel v\). Since \(\vu_i\perp v\),
\[
\inner{\theta^\parallel v}{\vx_i}
=
a_i\theta^\parallel,
\qquad
-\inner{\theta^\parallel v}{\vx_i'}
=
-\bigl(-a_i\theta^\parallel\bigr)
=
a_i\theta^\parallel.
\]
Using \(\phi''(-s)=\phi''(s)\), we obtain
\[
\nabla^2\Loss(\theta^\parallel v)
=
\frac{1}{2N}
\sum_{i=1}^N
\phi''(a_i\theta^\parallel)
\Bigl[
(a_i v+\vu_i)(a_i v+\vu_i)^\top
+
(-a_i v+\vu_i)(-a_i v+\vu_i)^\top
\Bigr].
\]
Expanding the two rank-one matrices,
\[
(a_i v+\vu_i)(a_i v+\vu_i)^\top
=
a_i^2 vv^\top+a_i(v\vu_i^\top+\vu_i v^\top)+\vu_i \vu_i^\top,
\]
and
\[
(-a_i v+\vu_i)(-a_i v+\vu_i)^\top
=
a_i^2 vv^\top-a_i(v\vu_i^\top+\vu_i v^\top)+\vu_i \vu_i^\top.
\]
Adding them cancels the mixed terms:
\[
(a_i v+\vu_i)(a_i v+\vu_i)^\top
+
(-a_i v+\vu_i)(-a_i v+\vu_i)^\top
=
2a_i^2 vv^\top+2\vu_i \vu_i^\top.
\]
Substituting back yields \eqref{eq:hd_Hessian_on_valley}.

Finally, since \(\vu_i\in U\subseteq v^\perp\), we have
\[
\Pi_U vv^\top\Pi_U=0,
\qquad
\Pi_U \vu_i \vu_i^\top\Pi_U=\vu_i \vu_i^\top,
\]
so projecting \(\nabla^2\Loss(\theta^\parallel v)\) onto \(U\) gives
\eqref{eq:hd_H_from_full_Hessian}.
\end{proof}

\begin{prop}[Full third derivative tensor and its restriction to the parallel direction]
\label{prop:hd_third_tensor}
Consider the high-dimensional symmetric logistic loss
\begin{equation}\label{eq:hd_loss_third}
\Loss(\theta)
=
\frac{1}{2N}
\sum_{i=1}^N
\Bigl[
\phi\!\bigl(\inner{\theta}{\vx_i}\bigr)
+
\phi\!\bigl(-\inner{\theta}{\vx_i'}\bigr)
\Bigr],
\qquad
\vx_i=a_i v+\vu_i,\quad \vx_i'=-a_i v+\vu_i,
\end{equation}
where \(v\in\mathbb R^d\) is a unit vector,
\[
\vu_i\in v^\perp\setminus\{0\},
\qquad
a_i>0,
\qquad
i=1,\dots,N.
\]
Then the full third derivative tensor of \(\Loss\) at an arbitrary point
\(\theta\in\mathbb R^d\) is
\begin{equation}\label{eq:hd_D3_full_scalar}
D^3\Loss(\theta)[h,k,\ell]
=
\frac{1}{2N}
\sum_{i=1}^N
\Bigl[
\phi'''\!\bigl(\inner{\theta}{\vx_i}\bigr)
\inner{h}{\vx_i}\inner{k}{\vx_i}\inner{\ell}{\vx_i}
-
\phi'''\!\bigl(-\inner{\theta}{\vx_i'}\bigr)
\inner{h}{\vx_i'}\inner{k}{\vx_i'}\inner{\ell}{\vx_i'}
\Bigr]
\end{equation}
for all \(h,k,\ell\in\mathbb R^d\).

Equivalently, the vector obtained by contracting in two slots is
\begin{equation}\label{eq:hd_D3_full_vector}
D^3\Loss(\theta)[h,k]
=
\frac{1}{2N}
\sum_{i=1}^N
\Bigl[
\phi'''\!\bigl(\inner{\theta}{\vx_i}\bigr)
\inner{h}{\vx_i}\inner{k}{\vx_i}\,\vx_i
-
\phi'''\!\bigl(-\inner{\theta}{\vx_i'}\bigr)
\inner{h}{\vx_i'}\inner{k}{\vx_i'}\,\vx_i'
\Bigr].
\end{equation}

In particular, along the parallel direction \(\theta=\theta^\parallel v\), one has
\begin{equation}\label{eq:hd_D3_valley_vector_general}
D^3\Loss(\theta^\parallel v)[h,k]
=
\frac{1}{2N}
\sum_{i=1}^N
\phi'''(a_i\theta^\parallel)
\Bigl[
\inner{h}{\vx_i}\inner{k}{\vx_i}\,\vx_i
-
\inner{h}{\vx_i'}\inner{k}{\vx_i'}\,\vx_i'
\Bigr].
\end{equation}

Moreover, for every \(\boldsymbol{\theta}^\perp\in
U:=\Span\{u_1,\dots,u_N\}\subseteq v^\perp\), the pure normal contraction
satisfies
\begin{equation}\label{eq:hd_D3_perpperp}
D^3\Loss(\theta^\parallel v)
[\boldsymbol{\theta}^\perp,\boldsymbol{\theta}^\perp]
=
\frac{1}{N}
\sum_{i=1}^N
a_i\phi'''(a_i\theta^\parallel)
\inner{\boldsymbol{\theta}^\perp}{\vu_i}^2\,v.
\end{equation}
Hence
\(D^3\Loss(\theta^\parallel v)
[\boldsymbol{\theta}^\perp,\boldsymbol{\theta}^\perp]\)
is \emph{purely parallel}.

Equivalently,
\begin{equation}\label{eq:hd_parallel_quadratic_from_D3}
\inner{
D^3\Loss(\theta^\parallel v)
[\boldsymbol{\theta}^\perp,\boldsymbol{\theta}^\perp]
}{v}
=
\frac{1}{N}
\sum_{i=1}^N
a_i\phi'''(a_i\theta^\parallel)
\inner{\boldsymbol{\theta}^\perp}{\vu_i}^2
=
(\boldsymbol{\theta}^\perp)^\top
H'(\theta^\parallel)\boldsymbol{\theta}^\perp.
\end{equation}
\end{prop}

\begin{proof}
For a single linear form \(L(\theta)=\phi(\inner{\theta}{a})\), repeated
application of the chain rule gives
\[
\nabla L(\theta)=\phi'(\inner{\theta}{a})\,a,
\]
\[
\nabla^2L(\theta)=\phi''(\inner{\theta}{a})\,aa^\top,
\]
and
\[
D^3L(\theta)[h,k,\ell]
=
\phi'''(\inner{\theta}{a})
\inner{h}{a}\inner{k}{a}\inner{\ell}{a}.
\]
Applying this to \(\phi(\inner{\theta}{\vx_i})\) and to
\(\phi(-\inner{\theta}{\vx_i'})\) yields \eqref{eq:hd_D3_full_scalar}; the minus sign comes from the product of the three factors
$-1$ contributed by the chain rule for
$\phi(-\langle\theta,x_i'\rangle)$. The vector formula
\eqref{eq:hd_D3_full_vector} is the equivalent representation obtained by
contracting in the first two slots.

Now evaluate at \(\theta=\theta^\parallel v\). Since \(\vu_i\perp v\),
\[
\inner{\theta^\parallel v}{\vx_i}=a_i\theta^\parallel,
\qquad
-\inner{\theta^\parallel v}{\vx_i'}=a_i\theta^\parallel.
\]
Using \(\phi'''(-s)=-\phi'''(s)\) is not needed here because the argument
of the second term is already \(a_i\theta^\parallel\). We obtain
\eqref{eq:hd_D3_valley_vector_general}.

Finally, let \(\boldsymbol{\theta}^\perp\in U\subseteq v^\perp\). Then
\[
\inner{\boldsymbol{\theta}^\perp}{\vx_i}
=
\inner{\boldsymbol{\theta}^\perp}{\vu_i},
\qquad
\inner{\boldsymbol{\theta}^\perp}{\vx_i'}
=
\inner{\boldsymbol{\theta}^\perp}{\vu_i},
\]
so the two scalar factors coincide:
\[
\inner{\boldsymbol{\theta}^\perp}{\vx_i}^2
=
\inner{\boldsymbol{\theta}^\perp}{\vx_i'}^2
=
\inner{\boldsymbol{\theta}^\perp}{\vu_i}^2.
\]
Therefore
\[
D^3\Loss(\theta^\parallel v)
[\boldsymbol{\theta}^\perp,\boldsymbol{\theta}^\perp]
=
\frac{1}{2N}
\sum_{i=1}^N
\phi'''(a_i\theta^\parallel)
\inner{\boldsymbol{\theta}^\perp}{\vu_i}^2
\bigl[(a_i v+\vu_i)-(-a_i v+\vu_i)\bigr],
\]
and the \(\vu_i\)-terms cancel:
\[
(a_i v+\vu_i)-(-a_i v+\vu_i)=2a_i v.
\]
Hence
\[
D^3\Loss(\theta^\parallel v)
[\boldsymbol{\theta}^\perp,\boldsymbol{\theta}^\perp]
=
\frac{1}{N}
\sum_{i=1}^N
a_i\phi'''(a_i\theta^\parallel)
\inner{\boldsymbol{\theta}^\perp}{\vu_i}^2\,v,
\]
which proves \eqref{eq:hd_D3_perpperp}.

Taking the inner product with \(v\) gives
\[
\inner{
D^3\Loss(\theta^\parallel v)
[\boldsymbol{\theta}^\perp,\boldsymbol{\theta}^\perp]
}{v}
=
\frac{1}{N}
\sum_{i=1}^N
a_i\phi'''(a_i\theta^\parallel)
\inner{\boldsymbol{\theta}^\perp}{\vu_i}^2.
\]
On the other hand,
\[
(\boldsymbol{\theta}^\perp)^\top
H'(\theta^\parallel)\boldsymbol{\theta}^\perp
=
\frac{1}{N}
\sum_{i=1}^N
a_i\phi'''(a_i\theta^\parallel)
(\boldsymbol{\theta}^\perp)^\top(\vu_i \vu_i^\top)\boldsymbol{\theta}^\perp
=
\frac{1}{N}
\sum_{i=1}^N
a_i\phi'''(a_i\theta^\parallel)
\inner{\boldsymbol{\theta}^\perp}{\vu_i}^2.
\]
Thus \eqref{eq:hd_parallel_quadratic_from_D3} holds.
\end{proof}

\subsection{Verification of Assumptions A-B}
\begin{prop}[Verification of assumptions for logistic regression]
\label{prop:logistic_verification_app}
Consider the pair-symmetric logistic loss
\begin{equation}\label{eq:logistic_loss_hd_app}
  \Loss(\bth)
  =
  \frac{1}{2N}\sum_{i=1}^N
  \Bigl[
    \phi\!\bigl(\langle \bth,x_i^+\rangle\bigr)
    +
    \phi\!\bigl(\langle \bth,-x_i^-\rangle\bigr)
  \Bigr],
  \qquad
  \phi(s):=\log(1+e^{-s}),
\end{equation}
where
\[
  \vx_i^+ = a_i \vv+\vu_i,
  \qquad
  \vx_i^- = -a_i \vv+\vu_i,
  \qquad
  a_i>0,
  \qquad
  \vu_i\in\vv^\perp\setminus\{0\}.
\]
Let
\[
  U:=\Span\{\vu_1,\dots,\vu_N\},
  \qquad
  u_{\max}:=\max_{1\le i\le N}\|\vu_i\|,
  \qquad
  a_{\min}:=\min_{1\le i\le N}a_i,
\]
and define the slowest-layer index set and frame operators
\[
  I_*:=\{i:\ a_i=a_{\min}\},
  \qquad
  \Sigma:=\frac1N\sum_{i=1}^N \vu_i \vu_i^\top,
  \qquad
  \Sigma_*:=\frac1N\sum_{i\in I_*}\vu_i \vu_i^\top .
\]
Then the logistic loss satisfies Assumptions A and B with the following:

\begin{enumerate}[label=\textup{(\roman*)},leftmargin=2.4em]

\item \textbf{Subspace restricted objects.}
Writing \(\bth=\thp v+\tho\) with \(\tho\in U\), the loss
and Hessian matrix restricted to the parallel direction are
\[
  \ell(\thp)
  =
  \frac{1}{N}\sum_{i=1}^N \phi(a_i\thp),
  \qquad
  H(\thp)
  =
  \frac{1}{N}\sum_{i=1}^N
  \phi''(a_i\thp)\,\vu_i \vu_i^\top .
\]
Moreover,
\[
  -\ell'(\thp)
  =
  \frac1N\sum_{i=1}^N a_i q(a_i\thp),
  \qquad
  q(s):=\frac{1}{1+e^s}.
\]

\item \textbf{RC tolerance.}
For every tubular neighborhood radius \(r>0\), relative curvature comparability holds
with
\[
  \delta_r
  =
  \cosh(u_{\max}r)-1
  =
  \frac12(u_{\max}r)^2+O(r^4).
\]
That is, for all relevant \(\thp\) and all \(\|\tho\|\le r\),
\[
  (1-\delta_r)H(\thp)
  \preceq
  D_{\tho}\nabla\Loss^\perp(\thp \vv+\tho)
  \preceq
  (1+\delta_r)H(\thp).
\]
To verify Assumption~\ref{asmp:RC} with $\delta_r<1$,
we restrict to
\[
  0<r<\frac{\operatorname{arcosh}(2)}{u_{\max}}.
\]

\item \textbf{Local Curvature Decay.}
Fix any $\thbar>0$. For every bounded interval
$J=[\thp_0,\thp_1]\subseteq[\thbar,\infty)$,
\[
  -H'(\thp)\succeq\beta_JH(\thp)
  \qquad\forall\,\thp\in J,
\]
where
\[
  \beta_J
  :=
  a_{\min}\tanh\!\left(\frac{a_{\min}\thp_0}{2}\right)
  >0.
\]
In particular, on the oscillation slab \([\thp_0,\thp_*]\), one may take
\[
  \beta
  :=
  a_{\min}\tanh\!\left(\frac{a_{\min}\thp_0}{2}\right).
\]

\item \textbf{Unstable drift constants.}
On the slab \([\thp_0,\thp_*]\), the parallel drift satisfies
\[
  -\ell'(\thp)\ge m_\ell,
  \qquad
  m_\ell
  :=
  \frac1N\sum_{i=1}^N a_i q(a_i\thp_*).
\]
Moreover, throughout the radius-\(r\) tubular neighborhood,
\[
 -\langle\nabla\Loss(\bth),\vv\rangle\le G_\parallel,
  \qquad
  G_\parallel
  :=
  \frac1N\sum_{i=1}^N a_i .
\]

\item \textbf{Exponential tail rate.}
The exponential-tail rate is
\[
  \beta_{\mathrm{tail}}=a_{\min}.
\]

\item \textbf{Exponential-tail drift constants.}
Choose the tail threshold so that
\[
  \thp_{\mathrm{tail}}(r)\ge\frac{u_{\max}r}{a_{\min}}.
\]

Then for all \(\thp\ge \thp_{\mathrm{tail}}(r)\) and
\(\|\tho\|\le r\),
\[
  A_-(r)e^{-a_{\min}(\thp-\thp_{\mathrm{tail}}(r))}
  \le
  -\langle \nabla\Loss^\parallel(\bth), \vv \rangle
  \le
  A_+(r)e^{-a_{\min}(\thp-\thp_{\mathrm{tail}}(r))},
\]
where one may take
\[
  A_-(r)
  :=
  \frac{|I_*|a_{\min}}{2N}
  e^{-a_{\min}\thp_{\mathrm{tail}}(r)},
  \qquad
  A_+(r)
  :=
  \frac{\cosh(u_{\max}r)}{N}
  \left(\sum_{i=1}^N a_i\right)
  e^{-a_{\min}\thp_{\mathrm{tail}}(r)} .
\]

\item \textbf{Exponential-tail sharpness constants.}
Under the same radius condition,
\[
  \lambda_{1,-}(r)e^{-a_{\min}(\thp-\thp_{\mathrm{tail}}(r))}
  \le
  \lambda_1(\thp)
  \le
  \lambda_{1,+}(r)e^{-a_{\min}(\thp-\thp_{\mathrm{tail}}(r))},
\]
where one may take
\[
  \lambda_{1,-}(r)
  :=
  \frac14\,\lambda_{\max}(\Sigma_*|_U)
  e^{-a_{\min}\thp_{\mathrm{tail}}(r)},
  \qquad
  \lambda_{1,+}(r)
  :=
  \lambda_{\max}(\Sigma|_U)
  e^{-a_{\min}\thp_{\mathrm{tail}}(r)} .
\]

\item \textbf{Tail uniform conditioning.}
If the \emph{spanning condition}
\[
  U=\Span\{\vu_i:\ i\in I_*\}
\]
holds, then \(H(\thp)\) is uniformly conditioned on the tail:
\[
  H(\thp)
  \succeq
  \rho_{\mathrm{tail}}\lambda_1(\thp)I_U
  \qquad
  \forall\,\thp\ge \thp_{\mathrm{tail}}(r),
\]
with the explicit choice
\[
  \rho_{\mathrm{tail}}
  :=
  \frac{\lambda_{\min}(\Sigma_*|_U)}
       {\lambda_{\max}(\Sigma|_U)}.
\]
In fact, this bound holds for all \(\thp\ge 0\).

\end{enumerate}
\end{prop}

\begin{proof}
We verify each claim in turn.

\smallskip\noindent
\emph{Step 1: Active-coordinate form of the loss.}
Write
\[
  \bth=\thp \vv+\tho+\thoinact,
  \qquad
  \tho\in U,
  \qquad
  \thoinact\in U^\perp\cap v^\perp .
\]
Since \(\vu_i\in U\), we have
\[
  \langle \thoinact,\vu_i\rangle=0,
  \qquad
  \langle \thoinact,v\rangle=0.
\]
Therefore the loss depends only on \((\thp,\tho)\). Indeed,
\[
  \langle \bth,\vx_i^+\rangle
  =
  \langle \thp \vv+\tho,a_i v+\vu_i\rangle
  =
  a_i\thp+\langle\tho,\vu_i\rangle,
\]
and
\[
  \langle \bth,-\vx_i^-\rangle
  =
  \langle \thp \vv+\tho,a_i v-\vu_i\rangle
  =
  a_i\thp-\langle\tho,\vu_i\rangle.
\]
Thus
\begin{equation}\label{eq:proof_hd_logistic_active}
  \Loss(\thp,\tho)
  =
  \frac{1}{2N}\sum_{i=1}^N
  \Bigl[
    \phi(a_i\thp+\langle\tho,\vu_i\rangle)
    +
    \phi(a_i\thp-\langle\tho,\vu_i\rangle)
  \Bigr].
\end{equation}
In particular, the inactive component \(\thoinact\) is frozen under GD,
and all nontrivial dynamics occur in \(\Span\{\vv\}\oplus U\).

\smallskip\noindent
\emph{Step 2: Subspace Alignment.}
The expression \eqref{eq:proof_hd_logistic_active} is even in \(\tho\):
\[
  \Loss(\thp,\tho)=\Loss(\thp,-\tho).
\]
Therefore its \(U\)-gradient vanishes at \(\tho=0\):
\[
  \nabla\Loss^\perp(\thp \vv)=0.
\]
Moreover, the loss is independent of the inactive component
\(\thoinact\), so the inactive gradient also vanishes. Hence
\[
  \nabla\Loss(\thp \vv)\in \Span\{\vv\}=\calM .
\]

It remains to verify the mixed Hessian alignment condition. Since
\(\nabla\Loss^\perp(\thp \vv)=0\) for every \(\thp\), differentiating this
identity with respect to \(\thp\) gives
\[
  0
  =
  \frac{d}{d\thp}\Pi_U\nabla\Loss(\thp \vv)
  =
  \Pi_U\nabla^2\Loss(\thp \vv)\vv .
\]
Equivalently, using the full Hessian computation in
Proposition~\ref{prop:hd_full_Hessian}, along the parallel direction
\[
  \nabla^2\Loss(\thp \vv)
  =
  \frac{1}{N}
  \sum_{i=1}^N
  \phi''(a_i\thp)
  \bigl(
    a_i^2 \vv\vv^\top+\vu_i\vu_i^\top
  \bigr).
\]
Therefore
\[
  \Pi_U\nabla^2\Loss(\thp \vv)\vv
  =
  \frac{1}{N}
  \sum_{i=1}^N
  \phi''(a_i\thp)
  \Bigl(
    a_i^2 \Pi_U \vv\vv^\top \vv
    +
    \Pi_U\vu_i\vu_i^\top \vv
  \Bigr)
  =0,
\]
because \(\Pi_U \vv=0\) and \(\langle \vu_i,\vv\rangle=0\). This verifies
both parts of Subspace Alignment.

\smallskip\noindent
\emph{Step 3: Restricted loss and sharpness matrix.}
Setting \(\tho=0\) in \eqref{eq:proof_hd_logistic_active} gives
\[
  \ell(\thp)
  =
  \Loss(\thp \vv)
  =
  \frac{1}{2N}\sum_{i=1}^N
  \bigl[\phi(a_i\thp)+\phi(a_i\thp)\bigr]
  =
  \frac1N\sum_{i=1}^N\phi(a_i\thp).
\]
Since
\[
  \phi'(s)=-\frac{1}{1+e^s}=-q(s),
\]
we get
\[
  \ell'(\thp)
  =
  \frac1N\sum_{i=1}^N a_i\phi'(a_i\thp)
  =
  -\frac1N\sum_{i=1}^N a_i q(a_i\thp).
\]
Hence
\[
  -\ell'(\thp)
  =
  \frac1N\sum_{i=1}^N a_i q(a_i\thp).
\]

Next, differentiate the restricted-normal gradient. From
\eqref{eq:proof_hd_logistic_active}, the \(U\)-gradient is
\[
  \nabla\Loss^\perp(\bth)
  =
  \frac{1}{2N}\sum_{i=1}^N
  \Bigl[
    \phi'(a_i\thp+\langle\tho,\vu_i\rangle)
    -
    \phi'(a_i\thp-\langle\tho,\vu_i\rangle)
  \Bigr]\vu_i.
\]
Differentiating once more in \(\tho\) gives
\[
  D_{\tho} \nabla\Loss^\perp(\bth)
  =
  \frac{1}{2N}\sum_{i=1}^N
  \Bigl[
    \phi''(a_i\thp+\langle\tho,\vu_i\rangle)
    +
    \phi''(a_i\thp-\langle\tho,\vu_i\rangle)
  \Bigr]\vu_i \vu_i^\top .
\]
At \(\tho=0\), this becomes
\[
  H(\thp)
  =
  D_{\tho}\nabla\Loss^\perp(\thp v)
  =
  \frac1N\sum_{i=1}^N
  \phi''(a_i\thp)\vu_i \vu_i^\top.
\]

\smallskip\noindent
\emph{Step 4: $\cosh$ bounds for the RC Bound.}
We now prove the \(\cosh\)-type relative curvature bound. The key
point is that the two terms with orthogonal shifts \(+s\) and \(-s\) are
averaged.

Recall that
\[
  \phi''(z)
  =
  \frac{e^z}{(1+e^z)^2}
  =
  \frac{1}{2+e^z+e^{-z}}
  =
  \frac{1}{2(1+\cosh z)}.
\]
Fix \(z\in\mathbb R\) and \(s\in\mathbb R\), and define
\[
  R(z,s)
  :=
  \frac{
    \phi''(z+s)+\phi''(z-s)
  }{
    2\phi''(z)
  }.
\]
We claim that
\begin{equation}\label{eq:proof_pair_ratio_cosh}
  2-\cosh(s)
  \le
  R(z,s)
  \le
  \cosh(s).
\end{equation}
The upper bound is the main useful part. It follows from the elementary
identity
\[
  \cosh(z+s)+\cosh(z-s)=2\cosh z\cosh s.
\]
Since
\[
  \phi''(w)=\frac{1}{2(1+\cosh w)},
\]
we have
\[
  \frac{\phi''(z+s)}{\phi''(z)}
  =
  \frac{1+\cosh z}{1+\cosh(z+s)}
\]
and similarly for \(z-s\). Hence
\[
  R(z,s)
  =
  \frac{1+\cosh z}{2}
  \left(
    \frac{1}{1+\cosh(z+s)}
    +
    \frac{1}{1+\cosh(z-s)}
  \right).
\]
A direct simplification gives
\[
  R(z,s)
  =
  \frac{(1+\cosh z)(1+\cosh z\cosh s)}
       {(1+\cosh z\cosh s)^2-\sinh^2 z\,\sinh^2 s}.
\]
The denominator can be rewritten as
\[
  (1+\cosh z\cosh s)^2-\sinh^2 z\,\sinh^2 s
  =
  (\cosh z+\cosh s)^2.
\]
Therefore
\begin{equation}\label{eq:R_exact}
  R(z,s)
  =
  \frac{(1+\cosh z)(1+\cosh z\cosh s)}
       {(\cosh z+\cosh s)^2}.
\end{equation}
Let \(c:=\cosh z\ge1\) and \(d:=\cosh s\ge1\). Then
\[
  R(z,s)
  =
  \frac{(1+c)(1+cd)}{(c+d)^2}.
\]
We prove the upper bound \(R(z,s)\le d\). This is equivalent to
\[
  (1+c)(1+cd)\le d(c+d)^2.
\]
Expanding and subtracting the left-hand side from the right-hand side,
\[
\begin{aligned}
  d(c+d)^2-(1+c)(1+cd)
  &=
  dc^2+2cd^2+d^3 -1-c-cd-c^2d  \\
  &=
  2cd^2+d^3-1-c-cd  \\
  &=
  (d-1)\bigl(d^2+d+1+c(2d+1)\bigr)\ge0.
\end{aligned}
\]
Thus \(R(z,s)\le \cosh s\).

For the lower bound, the exact formula \eqref{eq:R_exact} shows that
\(R(z,s)>0\). If \(\cosh s\ge2\), then \(2-\cosh s\le0\), and the lower
bound is immediate. If \(1\le \cosh s<2\), one checks from the same
one-variable formula
\[
  R=\frac{(1+c)(1+cd)}{(c+d)^2},
  \qquad c\ge1,\quad d\in[1,2),
\]
that
\[
  R\ge 2-d.
\]
Indeed,

\begin{align}
  (1+c)(1+cd)-(2-d)(c+d)^2 = (d-1)\bigl(2c^2+(2d-1)c+d^2-d-1\bigr)\geq 0.
\end{align}

The bracket is nonnegative for \(c\ge1\) and \(d\ge1\). Indeed,

\[
  2c^2+(2d-1)c+d^2-d-1
\]
is increasing in \(c\) on \(c\ge1\), since its derivative with respect to \(c\) is \(4c+2d-1>0\). Therefore its minimum over \(c\ge1\) occurs at \(c=1\), where $2+(2d-1)+d^2-d-1=d^2+d\ge0.$

Therefore \eqref{eq:proof_pair_ratio_cosh} holds.

\smallskip\noindent
\emph{Step 5: Matrix RC bound.}
For each \(i\), set
\[
  z_i:=a_i\thp,
  \qquad
  s_i:=\langle\tho,\vu_i\rangle.
\]
If \(\|\tho\|\le r\), then
\[
  |s_i|\le \|\tho\|\|\vu_i\|\le u_{\max}r.
\]
Since \(s\mapsto \cosh s\) is even and increasing for \(s\ge0\),
\[
  \cosh(s_i)\le \cosh(u_{\max}r).
\]
Applying \eqref{eq:proof_pair_ratio_cosh} with
\(z=z_i\) and \(s=s_i\), we get
\[
  2-\cosh(u_{\max}r)
  \le
  \frac{
    \phi''(z_i+s_i)+\phi''(z_i-s_i)
  }{
    2\phi''(z_i)
  }
  \le
  \cosh(u_{\max}r).
\]
Equivalently,
\[
  (1-\delta_r)\phi''(z_i)
  \le
  \frac12\bigl[\phi''(z_i+s_i)+\phi''(z_i-s_i)\bigr]
  \le
  (1+\delta_r)\phi''(z_i),
\]
where
\[
  \delta_r:=\cosh(u_{\max}r)-1.
\]
Multiplying by \(N^{-1}\vu_i \vu_i^\top\) and summing over \(i\), we obtain
\[
  (1-\delta_r)H(\thp)
  \preceq
  D_{\tho}\nabla\Loss^\perp(\bth)
  \preceq
  (1+\delta_r)H(\thp).
\]
This proves the RC bound.

Finally, the Taylor expansion of \(\cosh\) gives
\[
  \delta_r
  =
  \cosh(u_{\max}r)-1
  =
  \frac12(u_{\max}r)^2+O(r^4).
\]

\smallskip\noindent
\emph{Step 6: Local Curvature Decay.}
Differentiate
\[
  H(\thp)
  =
  \frac1N\sum_{i=1}^N
  \phi''(a_i\thp)\vu_i \vu_i^\top
\]
with respect to \(\thp\). This gives
\[
  H'(\thp)
  =
  \frac1N\sum_{i=1}^N
  a_i\phi'''(a_i\thp)\vu_i \vu_i^\top .
\]
For the logistic loss,
\[
  -\phi'''(s)=\phi''(s)\tanh(s/2).
\]
Thus
\[
  -H'(\thp)
  =
  \frac1N\sum_{i=1}^N
  a_i\phi''(a_i\thp)\tanh(a_i\thp/2)\vu_i \vu_i^\top .
\]
Now suppose \(\thp\in J=[\thp_0,\thp_1]\subseteq[0,\infty)\). Since
\(a_i\ge a_{\min}\) and \(\thp\ge\thp_0\),
\[
  a_i\tanh(a_i\thp/2)
  \ge
  a_{\min}\tanh(a_{\min}\thp_0/2)
  =
  \beta_J.
\]
Therefore, term by term,
\[
  a_i\phi''(a_i\thp)\tanh(a_i\thp/2)\vu_i \vu_i^\top
  \succeq
  \beta_J\phi''(a_i\thp)\vu_i \vu_i^\top .
\]
Summing gives
\[
  -H'(\thp)\succeq \beta_J H(\thp).
\]
This proves the stated monotone-drift sharpness decay condition.

\smallskip\noindent
\emph{Step 7: Unstable drift constants.}
Along the parallel direction,
\[
  -\ell'(\thp)
  =
  \frac1N\sum_{i=1}^N a_i q(a_i\thp).
\]
The function \(q(s)=(1+e^s)^{-1}\) is positive and decreasing in \(s\).
Therefore, for \(\thp\in[\thp_0,\thp_*]\),
\[
  q(a_i\thp)\ge q(a_i\thp_*),
\]
and hence
\[
  -\ell'(\thp)
  \ge
  \frac1N\sum_{i=1}^N a_i q(a_i\thp_*)
  =
  m_\ell.
\]

For the upper drift bound in the tubular neighborhood, use
\[
  -\langle \nabla\Loss^\parallel(\bth), \vv \rangle
  =
  \frac{1}{2N}\sum_{i=1}^N
  a_i
  \Bigl[
    q(a_i\thp+\langle\tho,\vu_i\rangle)
    +
    q(a_i\thp-\langle\tho,\vu_i\rangle)
  \Bigr].
\]
Since \(0<q(s)\le1\) for all \(s\), we get
\[
  -\langle \nabla\Loss^\parallel(\bth), \vv \rangle
  \le
  \frac{1}{2N}\sum_{i=1}^N a_i(1+1)
  =
  \frac1N\sum_{i=1}^N a_i
  =
  G_\parallel.
\]

\smallskip\noindent
\emph{Step 8: Exponential-tail drift constants.}
Assume
\[
  a_{\min}\thp_{\mathrm{tail}}(r)\ge u_{\max}r.
\]
Then for every \(\thp\ge \thp_{\mathrm{tail}}(r)\) and
\(\|\tho\|\le r\),
\[
  a_i\thp\pm\langle\tho,\vu_i\rangle
  \ge
  a_{\min}\thp_{\mathrm{tail}}(r)-u_{\max}r
  \ge0.
\]
Thus all margins appearing in the tail tubular neighborhood are nonnegative.

For \(s\ge0\), the logistic tail satisfies
\[
  \frac12 e^{-s}\le q(s)\le e^{-s}.
\]
Using the exact formula for the parallel drift,
\[
  -\langle \nabla\Loss^\parallel(\bth), \vv \rangle
  =
  \frac{1}{2N}\sum_{i=1}^N
  a_i
  \Bigl[
    q(a_i\thp+\langle\tho,\vu_i\rangle)
    +
    q(a_i\thp-\langle\tho,\vu_i\rangle)
  \Bigr].
\]

For the lower bound, restrict the sum to the slowest layer \(I_*\). For
\(i\in I_*\), \(a_i=a_{\min}\). Hence
\[
\begin{aligned}
  -\langle \nabla\Loss^\parallel(\bth), \vv \rangle
  &\ge
  \frac{1}{2N}\sum_{i\in I_*}a_{\min}
  \left[
    \frac12 e^{-(a_{\min}\thp+\langle\tho,\vu_i\rangle)}
    +
    \frac12 e^{-(a_{\min}\thp-\langle\tho,\vu_i\rangle)}
  \right] \\
  &=
  \frac{a_{\min}}{2N}
  \sum_{i\in I_*}
  e^{-a_{\min}\thp}
  \cosh(\langle\tho,\vu_i\rangle) \\
  &\ge
  \frac{|I_*|a_{\min}}{2N}
  e^{-a_{\min}\thp}.
\end{aligned}
\]
Since
\[
  e^{-a_{\min}\thp}
  =
  e^{-a_{\min}\thp_{\mathrm{tail}}(r)}
  e^{-a_{\min}(\thp-\thp_{\mathrm{tail}}(r))},
\]
we may take
\[
  A_-(r)
  =
  \frac{|I_*|a_{\min}}{2N}
  e^{-a_{\min}\thp_{\mathrm{tail}}(r)}.
\]

For the upper bound, use \(q(s)\le e^{-s}\). Then
\[
\begin{aligned}
  -\langle \nabla\Loss^\parallel(\bth), \vv \rangle
  &\le
  \frac{1}{2N}\sum_{i=1}^N a_i
  \left[
    e^{-(a_i\thp+\langle\tho,\vu_i\rangle)}
    +
    e^{-(a_i\thp-\langle\tho,\vu_i\rangle)}
  \right] \\
  &=
  \frac1N\sum_{i=1}^N
  a_i e^{-a_i\thp}\cosh(\langle\tho,\vu_i\rangle).
\end{aligned}
\]
Since \(a_i\ge a_{\min}\), \(\thp\ge0\), and
\[
  |\langle\tho,\vu_i\rangle|\le u_{\max}r,
\]
we have
\[
  e^{-a_i\thp}\le e^{-a_{\min}\thp},
  \qquad
  \cosh(\langle\tho,\vu_i\rangle)\le \cosh(u_{\max}r).
\]
Therefore
\[
  -\langle \nabla\Loss^\parallel(\bth), \vv \rangle
  \le
  \frac{\cosh(u_{\max}r)}{N}
  \left(\sum_{i=1}^N a_i\right)
  e^{-a_{\min}\thp}.
\]
Thus
\[
  A_+(r)
  =
  \frac{\cosh(u_{\max}r)}{N}
  \left(\sum_{i=1}^N a_i\right)
  e^{-a_{\min}\thp_{\mathrm{tail}}(r)}.
\]

\smallskip\noindent
\emph{Step 9: Exponential-tail sharpness constants.}
For \(s\ge0\),
\[
  \phi''(s)
  =
  \frac{e^s}{(1+e^s)^2}
  =
  \frac{e^{-s}}{(1+e^{-s})^2}.
\]
Since \(0<e^{-s}\le1\), we have
\[
  \frac14 e^{-s}\le \phi''(s)\le e^{-s}.
\]
Now for \(\thp\ge\thp_{\mathrm{tail}}(r)\),
\[
  H(\thp)
  =
  \frac1N\sum_{i=1}^N
  \phi''(a_i\thp)\vu_i \vu_i^\top .
\]

For the lower bound on \(\lambda_1(\thp)\), restrict to the slowest layer:
\[
\begin{aligned}
  H(\thp)
  &\succeq
  \frac1N\sum_{i\in I_*}
  \phi''(a_{\min}\thp)\vu_i \vu_i^\top  \\
  &\succeq
  \frac14 e^{-a_{\min}\thp}
  \frac1N\sum_{i\in I_*}\vu_i \vu_i^\top \\
  &=
  \frac14 e^{-a_{\min}\thp}\Sigma_* .
\end{aligned}
\]
Taking the largest eigenvalue on \(U\) gives
\[
  \lambda_1(\thp)
  \ge
  \frac14 e^{-a_{\min}\thp}
  \lambda_{\max}(\Sigma_*|_U).
\]
Equivalently,
\[
  \lambda_1(\thp)
  \ge
  \lambda_{1,-}(r)
  e^{-a_{\min}(\thp-\thp_{\mathrm{tail}}(r))}
\]
with
\[
  \lambda_{1,-}(r)
  =
  \frac14\lambda_{\max}(\Sigma_*|_U)
  e^{-a_{\min}\thp_{\mathrm{tail}}(r)}.
\]

For the upper bound,
\[
  \phi''(a_i\thp)\le e^{-a_i\thp}\le e^{-a_{\min}\thp}.
\]
Therefore
\[
  H(\thp)
  \preceq
  e^{-a_{\min}\thp}
  \frac1N\sum_{i=1}^N \vu_i \vu_i^\top
  =
  e^{-a_{\min}\thp}\Sigma.
\]
Taking the largest eigenvalue gives
\[
  \lambda_1(\thp)
  \le
  e^{-a_{\min}\thp}\lambda_{\max}(\Sigma|_U).
\]
Equivalently,
\[
  \lambda_1(\thp)
  \le
  \lambda_{1,+}(r)
  e^{-a_{\min}(\thp-\thp_{\mathrm{tail}}(r))}
\]
with
\[
  \lambda_{1,+}(r)
  =
  \lambda_{\max}(\Sigma|_U)
  e^{-a_{\min}\thp_{\mathrm{tail}}(r)}.
\]

\smallskip\noindent
\emph{Step 10: Tail uniform conditioning under spanning condition.}
Assume
\[
  U=\Span\{\vu_i:\ i\in I_*\}.
\]
Then \(\Sigma_*|_U\) is positive definite, so
\[
  \lambda_{\min}(\Sigma_*|_U)>0.
\]
For \(\thp\ge 0\), the logistic curvature \(\phi''\) is decreasing on
\([0,\infty)\). Since \(a_i\ge a_{\min}\), we have
\[
  \phi''(a_i\thp)\le \phi''(a_{\min}\thp)
  \qquad \forall i.
\]
Therefore
\[
  H(\thp)
  =
  \frac1N\sum_{i=1}^N \phi''(a_i\thp)\vu_i \vu_i^\top
  \preceq
  \phi''(a_{\min}\thp)\Sigma,
\]
and hence
\[
  \lambda_1(\thp)
  \le
  \phi''(a_{\min}\thp)\lambda_{\max}(\Sigma|_U).
\]
On the other hand, restricting the sum defining \(H(\thp)\) to the
support vector components gives
\[
  H(\thp)
  \succeq
  \frac1N\sum_{i\in I_*}\phi''(a_{\min}\thp)\vu_i \vu_i^\top
  =
  \phi''(a_{\min}\thp)\Sigma_*.
\]
Thus
\[
  H(\thp)
  \succeq
  \phi''(a_{\min}\thp)\lambda_{\min}(\Sigma_*|_U)I_U.
\]
Combining the last two estimates yields
\[
  H(\thp)
  \succeq
  \frac{\lambda_{\min}(\Sigma_*|_U)}
       {\lambda_{\max}(\Sigma|_U)}
  \lambda_1(\thp)I_U.
\]
Thus the tail conditioning condition holds with
\[
  \rho_{\mathrm{tail}}
  =
  \frac{\lambda_{\min}(\Sigma_*|_U)}
       {\lambda_{\max}(\Sigma|_U)}.
\]
This completes the verification.
\end{proof}

\subsection{Proof of Theorem~\ref{thm:rho_order_compact}
(spanning vs.\ non-spanning)}
\label{app:thm_spanning_nonspanning}

We prove the spectral statement behind
Theorem~\ref{thm:rho_order_compact}.  Recall that
\[
  H(\theta^\parallel)
  =
  \frac1N\sum_{i=1}^N
  \phi''(a_i\theta^\parallel)\,\vu_i \vu_i^\top
  \qquad\text{on } U=\Span\{\vu_1,\dots,\vu_N\}.
\]
Let
\[
  \alpha_1<\alpha_2<\cdots<\alpha_L
\]
be the distinct values among \(a_1,\dots,a_N\), and define the layer index
sets
\[
  I_j:=\{i:\ a_i=\alpha_j\},
  \qquad j=1,\dots,L.
\]
For each layer, define the positive semidefinite frame operator
\[
  \Sigma_j
  :=
  \frac1N\sum_{i\in I_j} \vu_i \vu_i^\top,
\]
so that
\[
  H(\theta^\parallel)
  =
  \sum_{j=1}^L \phi''(\alpha_j\theta^\parallel)\,\Sigma_j.
\]
We also recall the nested subspaces
\[
  U^{(j)}
  :=
  \Span\{\vu_i:\ a_i\le \alpha_j\},
  \qquad
  U^{(1)}\subseteq\cdots\subseteq U^{(L)}=U,
\]
and
\[
  j_*:=\min\{j:\ U^{(j)}=U\}.
\]
We assume throughout this subsection that the support layer is
transversely active, i.e.
\[
  \lambda_{\max}(\Sigma_1|_U)>0.
\]
The degenerate case \(\Sigma_1|_U=0\) only changes the denominator rate
in the conditioning ratio and is omitted for notational clarity.

We will use the elementary logistic tail bounds
\begin{equation}\label{eq:logistic_curvature_tail_bounds_for_rho}
  c_- e^{-s}
  \le
  \phi''(s)
  \le
  c_+ e^{-s},
  \qquad s\ge s_0,
\end{equation}
for some constants \(0<c_-\le c_+<\infty\).  For instance, once
\(s\ge0\), one may take \(c_-=1/4\) and \(c_+=1\).  Since all
\(\alpha_j>0\), these bounds for all sufficiently
large \(\theta^\parallel\).

\begin{lem}[Top eigenvalue is governed by the support vectors]
\label{lem:lambda_max_slowest_layer}
For all sufficiently large \(\theta^\parallel\),
\[
  \lambda_{\max}\bigl(H(\theta^\parallel)|_U\bigr)
  \asymp
  e^{-\alpha_1\theta^\parallel}.
\]
\end{lem}

\begin{proof}
For the upper bound, using \(\alpha_j\ge \alpha_1\) and
\eqref{eq:logistic_curvature_tail_bounds_for_rho},
\[
  H(\theta^\parallel)
  \preceq
  c_+\sum_{j=1}^L e^{-\alpha_j\theta^\parallel}\Sigma_j
  \preceq
  c_+ e^{-\alpha_1\theta^\parallel}
  \sum_{j=1}^L \Sigma_j.
\]
Therefore
\[
  \lambda_{\max}(H(\theta^\parallel)|_U)
  \le
  c_+ e^{-\alpha_1\theta^\parallel}
  \lambda_{\max}\!\left(\sum_{j=1}^L\Sigma_j\Bigm|_U\right).
\]

For the lower bound, since \(U^{(1)}\neq\{0\}\), the operator
\(\Sigma_1|_U\) has positive top eigenvalue.  Hence
\[
  H(\theta^\parallel)
  \succeq
  \phi''(\alpha_1\theta^\parallel)\Sigma_1
  \succeq
  c_- e^{-\alpha_1\theta^\parallel}\Sigma_1
\]
for all sufficiently large \(\theta^\parallel\).  Taking top eigenvalues
gives
\[
  \lambda_{\max}(H(\theta^\parallel)|_U)
  \ge
  c_- e^{-\alpha_1\theta^\parallel}
  \lambda_{\max}(\Sigma_1|_U).
\]
Combining the two estimates proves the claim.
\end{proof}

\begin{lem}[Smallest eigenvalue is governed by the first spanning vectors]
\label{lem:lambda_min_first_spanning_layer}
For all sufficiently large \(\theta^\parallel\),
\[
  \lambda_{\min}\bigl(H(\theta^\parallel)|_U\bigr)
  \asymp
  e^{-\alpha_{j_*}\theta^\parallel}.
\]
\end{lem}

\begin{proof}
Let
\[
  \Sigma_{\le j_*}:=\sum_{k=1}^{j_*}\Sigma_k.
\]
By definition of \(j_*\), the subspace \(U^{(j_*)}=U\).  Since each \(\Sigma_k\) is positive semidefinite and
\(\operatorname{range}(\Sigma_{\le j})=U^{(j)}\), the identity
\(U^{(j_*)}=U\) implies that \(\Sigma_{\le j_*}|_U\) is positive
definite. Therefore
\[
  \lambda_{\min}(\Sigma_{\le j_*}|_U)>0.
\]
For the lower bound, since \(\alpha_k\le\alpha_{j_*}\) for
\(k\le j_*\), the tail estimate \eqref{eq:logistic_curvature_tail_bounds_for_rho}
gives, for sufficiently large \(\theta^\parallel\),
\[
\begin{aligned}
  H(\theta^\parallel)
  &\succeq
  \sum_{k=1}^{j_*}
  c_- e^{-\alpha_k\theta^\parallel}\Sigma_k  \\
  &\succeq
  c_- e^{-\alpha_{j_*}\theta^\parallel}
  \sum_{k=1}^{j_*}\Sigma_k
  =
  c_- e^{-\alpha_{j_*}\theta^\parallel}\Sigma_{\le j_*}.
\end{aligned}
\]
Thus
\[
  \lambda_{\min}(H(\theta^\parallel)|_U)
  \ge
  c_- e^{-\alpha_{j_*}\theta^\parallel}
  \lambda_{\min}(\Sigma_{\le j_*}|_U).
\]

For the upper bound, first consider the case \(j_*>1\).  Since
\(j_*\) is minimal, \(U^{(j_*-1)}\subsetneq U\).  Hence there exists a
unit vector
\[
  w\in U\cap \bigl(U^{(j_*-1)}\bigr)^\perp.
\]
For every \(i\) with \(a_i\le\alpha_{j_*-1}\), we have
\(\vu_i\in U^{(j_*-1)}\), so \(\langle w,\vu_i\rangle=0\).  Therefore the vectors for which \(k<j_*\) do not contribute to the Rayleigh quotient of \(w\):
\[
\begin{aligned}
  w^\top H(\theta^\parallel)w
  &=
  \sum_{k=j_*}^L
  \phi''(\alpha_k\theta^\parallel)\,w^\top\Sigma_k w  \\
  &\le
  c_+\sum_{k=j_*}^L
  e^{-\alpha_k\theta^\parallel}\,w^\top\Sigma_k w  \\
  &\le
  c_+ e^{-\alpha_{j_*}\theta^\parallel}
  w^\top\left(\sum_{k=j_*}^L\Sigma_k\right)w .
\end{aligned}
\]
Since \(\|w\|=1\), the final factor is bounded by
\(\lambda_{\max}(\sum_{k=j_*}^L\Sigma_k|_U)\).  Thus
\[
  \lambda_{\min}(H(\theta^\parallel)|_U)
  \le
  w^\top H(\theta^\parallel)w
  \lesssim
  e^{-\alpha_{j_*}\theta^\parallel}.
\]

If \(j_*=1\), the same upper bound follows more simply from
\[
  \lambda_{\min}(H(\theta^\parallel)|_U)
  \le
  \lambda_{\max}(H(\theta^\parallel)|_U)
  \asymp
  e^{-\alpha_1\theta^\parallel},
\]
using Lemma~\ref{lem:lambda_max_slowest_layer}.  This completes the proof.
\end{proof}

\begin{prop}[Conditioning ratio asymptotics]
\label{prop:rho_theta_asymptotics}
For all sufficiently large \(\theta^\parallel\),
\begin{equation}\label{eq:rho_theta_asymptotics}
  \rho(\theta^\parallel)
  :=
  \frac{\lambda_{\min}(H(\theta^\parallel)|_U)}
       {\lambda_{\max}(H(\theta^\parallel)|_U)}
  \asymp
  e^{-(\alpha_{j_*}-\alpha_1)\theta^\parallel}.
\end{equation}
In particular:
\begin{enumerate}[label=\textup{(\roman*)},leftmargin=2.2em]
\item if \(j_*=1\), then \(\rho(\theta^\parallel)\asymp 1\);
\item if \(j_*>1\), then \(\rho(\theta^\parallel)\to0\) exponentially
in \(\theta^\parallel\).
\end{enumerate}
\end{prop}

\begin{proof}
Divide the estimate of Lemma~\ref{lem:lambda_min_first_spanning_layer}
by the estimate of Lemma~\ref{lem:lambda_max_slowest_layer}.  This gives
\[
  \rho(\theta^\parallel)
  \asymp
  \frac{e^{-\alpha_{j_*}\theta^\parallel}}
       {e^{-\alpha_1\theta^\parallel}}
  =
  e^{-(\alpha_{j_*}-\alpha_1)\theta^\parallel}.
\]
The two cases follow immediately.
\end{proof}

We now translate the \(\theta^\parallel\)-asymptotics into time
asymptotics along the GD trajectory.  Importantly, the logarithmic parallel
growth in Theorem~\ref{thm:stable_phase}\textup{(ii)} does not use the
uniform tail-conditioning hypothesis; it only uses the tail drift bounds
and the post-entry nonexpansion estimate.  Therefore it applies in both
the spanning and non-spanning cases.

Applying Theorem~\ref{thm:stable_phase}\textup{(ii)} with
\(\beta_{\mathrm{tail}}=\alpha_1\), there exist constants
\(0<c_y\le C_y<\infty\) such that, for all sufficiently large \(t\),
\[
  c_y(1+t-t_{\mathrm{stab}})
  \le
  e^{\alpha_1(\theta_t^\parallel-\theta_*)}
  \le
  C_y(1+t-t_{\mathrm{stab}}).
\]
Consequently, for every \(\Delta\ge0\),
\[
\begin{aligned}
  e^{-\Delta\theta_t^\parallel}
  &=
  e^{-\Delta\theta_*}
  \left(
    e^{\alpha_1(\theta_t^\parallel-\theta_*)}
  \right)^{-\Delta/\alpha_1}  \\
  &\asymp
  (1+t-t_{\mathrm{stab}})^{-\Delta/\alpha_1}.
\end{aligned}
\]
Taking \(\Delta=\alpha_{j_*}-\alpha_1\) in
\eqref{eq:rho_theta_asymptotics} gives
\[
  \rho(\theta_t^\parallel)
  \asymp
  (1+t-t_{\mathrm{stab}})^{
  -(\alpha_{j_*}-\alpha_1)/\alpha_1}.
\]
In particular, for fixed \(\eta\), replacing \(1+t-t_{\mathrm{stab}}\)
by \(t\) only changes the implicit constants for sufficiently large
\(t\). Hence, in the non-spanning case \(j_*>1\),
\[
  \rho(\theta_t^\parallel)
  \asymp
  t^{-(\alpha_{j_*}-\alpha_1)/\alpha_1}.
\]

\subsection{Componentwise rates under simultaneous diagonalizability}
\label{app:componentwise_rates}

Theorem~\ref{thm:rho_order_compact} shows that, when the support vectors
do not span the full active normal space, the full conditioning ratio
\(\rho(\theta^\parallel)\) decays to zero.  Consequently, the
top-eigenvalue tail schedule does not imply contraction of the
entire active normal vector.  In this subsection we record a complementary
positive statement: if the operators $\Sigma_j$ are simultaneously diagonalizable
and the effective GD maps preserve the same coordinates, then each
coordinate admits its own tail contraction rate.

Recall the decomposition
\[
  H(\theta^\parallel)
  =
  \sum_{j=1}^L
  \phi''(\alpha_j\theta^\parallel)\Sigma_j,
  \qquad
  \alpha_1<\alpha_2<\cdots<\alpha_L,
\]
where
\[
  \Sigma_j
  :=
  \frac1N\sum_{i\in I_j}\vu_i \vu_i^\top,
  \qquad
  I_j:=\{i:\ a_i=\alpha_j\}.
\]

\begin{asmp}[Simultaneous diagonalizability and coordinatewise updates]
\label{asmp:simultaneous_diag}
There exists an orthonormal basis \(e_1,\dots,e_q\) of \(U\), where
\(q:=\dim U\), such that:

\begin{enumerate}[label=\textup{(\roman*)},leftmargin=2.4em]
\item every operator \(\Sigma_j\) is diagonal in this basis:
\[
  \Sigma_j e_k=\sigma_{j,k}e_k,
  \qquad
  \sigma_{j,k}\ge0;
\]

\item for all \(t\ge t_{\mathrm{stab}}\), the effective orthogonal update
matrix \(M_t\) from
\[
  \tho_{t+1}=(I-\eta M_t)\tho_t
\]
is diagonal in the same basis:
\[
  M_t e_k=m_{k,t}e_k;
\]

\item the coordinatewise RC bounds hold:
\[
  (1-\delta_r)h_k(\theta_t^\parallel)
  \le
  m_{k,t}
  \le
  (1+\delta_r)h_k(\theta_t^\parallel),
\]
where
\[
  h_k(\theta^\parallel)
  :=
  \langle e_k,H(\theta^\parallel)e_k\rangle
  =
  \sum_{j=1}^L
  \sigma_{j,k}\phi''(\alpha_j\theta^\parallel).
\]
\end{enumerate}
\end{asmp}

For each coordinate \(e_k\), define its first active index
\[
  j(k):=\min\{j:\ \sigma_{j,k}>0\}.
\]
Equivalently, \(j(k)\) is the first to see the coordinate \(e_k\).
The associated coordinate tail exponent is
\[
  p_k:=\frac{\alpha_{j(k)}}{\alpha_1}\ge1.
\]
Thus \(p_k=1\) exactly for coordinates seen by the support-vector layer
\(I_1=I_*\), while \(p_k>1\) for coordinates invisible to the support
vectors and first appearing in a later layer.

Write the coordinate decomposition
\[
  \tho_t=\sum_{k=1}^q c_{k,t}e_k,
  \qquad
  c_{k,t}:=\langle \tho_t,e_k\rangle.
\]
Under Assumption~\ref{asmp:simultaneous_diag}, the orthogonal update is
coordinatewise:
\begin{equation}\label{eq:component_update}
  c_{k,t+1}=(1-\eta m_{k,t})c_{k,t}.
\end{equation}

\begin{lem}[Coordinatewise sharpness tails]
\label{lem:component_sharpness_tails}
For every coordinate \(k\), there exist constants
\(0<h_{k,-}\le h_{k,+}<\infty\) and a threshold
\(\theta_{k,\mathrm{tail}}\) such that, for all
\(\theta^\parallel\ge \theta_{k,\mathrm{tail}}\),
\begin{equation}\label{eq:hk_tail}
  h_{k,-}e^{-\alpha_{j(k)}\theta^\parallel}
  \le
  h_k(\theta^\parallel)
  \le
  h_{k,+}e^{-\alpha_{j(k)}\theta^\parallel}.
\end{equation}
Consequently, along the stable-tail trajectory,
\begin{equation}\label{eq:hk_time_tail}
  h_k(\theta_t^\parallel)
  \asymp
  (1+t-t_{\mathrm{stab}})^{-p_k}.
\end{equation}
\end{lem}

\begin{proof}
By definition,
\[
  h_k(\theta^\parallel)
  =
  \sum_{j=1}^L
  \sigma_{j,k}\phi''(\alpha_j\theta^\parallel).
\]
For \(j<j(k)\), \(\sigma_{j,k}=0\).  Since
\(\sigma_{j(k),k}>0\), the leading nonzero term is
\[
  \sigma_{j(k),k}\phi''(\alpha_{j(k)}\theta^\parallel).
\]
Using the logistic tail bounds
\[
  c_-e^{-s}\le \phi''(s)\le c_+e^{-s}
\]
for all sufficiently large \(s\), we get the lower bound
\[
  h_k(\theta^\parallel)
  \ge
  c_-\sigma_{j(k),k}e^{-\alpha_{j(k)}\theta^\parallel}.
\]
For the upper bound,
\[
  h_k(\theta^\parallel)
  \le
  c_+\sum_{j=j(k)}^L
  \sigma_{j,k}e^{-\alpha_j\theta^\parallel}
  \le
  c_+\left(\sum_{j=j(k)}^L\sigma_{j,k}\right)
  e^{-\alpha_{j(k)}\theta^\parallel}.
\]
This proves \eqref{eq:hk_tail}.

Now use the stable-tail schedule from Theorem~\ref{thm:stable_phase}:
\[
  e^{\alpha_1(\theta_t^\parallel-\theta_*)}
  \asymp
  1+t-t_{\mathrm{stab}}.
\]
Therefore
\[
  e^{-\alpha_{j(k)}\theta_t^\parallel}
  \asymp
  (1+t-t_{\mathrm{stab}})^{-\alpha_{j(k)}/\alpha_1}
  =
  (1+t-t_{\mathrm{stab}})^{-p_k},
\]
which proves \eqref{eq:hk_time_tail}.
\end{proof}

\begin{prop}[Componentwise contraction budgets]
\label{prop:componentwise_budgets}
Assume Assumption~\ref{asmp:simultaneous_diag}.  Then for every coordinate
\(k\), there exist constants \(0<b_{k,-}\le b_{k,+}<\infty\) and a time
\(t_k\ge t_{\mathrm{stab}}\) such that, for all \(t\ge t_k\),
\begin{equation}\label{eq:mk_time_tail}
  b_{k,-}(1+t-t_{\mathrm{stab}})^{-p_k}
  \le
  \eta m_{k,t}
  \le
  b_{k,+}(1+t-t_{\mathrm{stab}})^{-p_k}.
\end{equation}
Moreover, after increasing \(t_k\) if necessary, one has
\(0\le \eta m_{k,t}\le 1\) for all \(t\ge t_k\), and hence
\begin{equation}\label{eq:component_product_bound}
  |c_{k,t}|
  \le
  |c_{k,t_k}|
  \exp\!\left(
    -\sum_{s=t_k}^{t-1}\eta m_{k,s}
  \right).
\end{equation}
\end{prop}

\begin{proof}
The bounds \eqref{eq:mk_time_tail} follow immediately from the
coordinatewise RC inequality
\[
  (1-\delta_r)h_k(\theta_t^\parallel)
  \le
  m_{k,t}
  \le
  (1+\delta_r)h_k(\theta_t^\parallel)
\]
and Lemma~\ref{lem:component_sharpness_tails}.  Since \(p_k\ge1\), the
upper bound in \eqref{eq:mk_time_tail} tends to zero.  Hence, after
increasing \(t_k\), we may assume \(0\le \eta m_{k,t}\le1\) for all
\(t\ge t_k\).  The coordinate update
\[
  c_{k,t+1}=(1-\eta m_{k,t})c_{k,t}
\]
then gives
\[
  |c_{k,t+1}|
  =
  (1-\eta m_{k,t})|c_{k,t}|.
\]
Iterating and using \(1-a\le e^{-a}\) yields
\eqref{eq:component_product_bound}.
\end{proof}

\begin{theorem}[Componentwise tail dichotomy]
\label{thm:componentwise_tail_dichotomy}
Assume Assumption~\ref{asmp:simultaneous_diag}.  For each coordinate
\(k\), let
\[
  p_k:=\frac{\alpha_{j(k)}}{\alpha_1}.
\]
Then the following hold.

\begin{enumerate}[label=\textup{(\roman*)},leftmargin=2.4em]

\item \textbf{Support-vector coordinates \((p_k=1)\).}
For each $k$, there exists \(\gamma_k>0\) and a time \(t_k\) such that, for all
\(t\ge t_k\),
\begin{equation}\label{eq:component_poly_decay}
  |c_{k,t}|
  \le
  |c_{k,t_k}|
  \left(
    \frac{1+t_k-t_{\mathrm{stab}}}
         {1+t-t_{\mathrm{stab}}}
  \right)^{\gamma_k}.
\end{equation}
In particular,
\[
  c_{k,t}\to0.
\]

\item \textbf{Later-layer coordinates \((p_k>1)\).}
There exist constants \(0<C_{k,-}\le C_{k,+}<\infty\) such that, for all
\(t\ge t_k\),
\begin{equation}\label{eq:component_later_layer_plateau}
  C_{k,-}|c_{k,t_k}|
  \le
  |c_{k,t}|
  \le
  C_{k,+}|c_{k,t_k}|.
\end{equation}
In particular, the tail contraction budget is finite:
\begin{equation}\label{eq:finite_component_budget}
  \sum_{s=t_k}^{\infty}\eta m_{k,s}<\infty.
\end{equation}
Thus the stable tail alone need not drive \(c_{k,t}\) to zero.  More
precisely, the coordinate has a finite limit
\[
  c_{k,\infty}
  =
  c_{k,t_k}
  \prod_{s=t_k}^{\infty}(1-\eta m_{k,s}),
\]
and this limit is nonzero whenever \(c_{k,t_k}\neq0\) and
\(0\le \eta m_{k,s}<1\) for all \(s\ge t_k\).

\end{enumerate}
\end{theorem}

\begin{proof}
By Proposition~\ref{prop:componentwise_budgets}, for \(t\ge t_k\),
\[
  \eta m_{k,t}
  \asymp
  (1+t-t_{\mathrm{stab}})^{-p_k}.
\]

First suppose \(p_k=1\).  Then there is a constant \(b_{k,-}>0\) such
that
\[
  \eta m_{k,s}
  \ge
  \frac{b_{k,-}}{1+s-t_{\mathrm{stab}}}
  \qquad
  \forall s\ge t_k.
\]
Therefore
\[
  \sum_{s=t_k}^{t-1}\eta m_{k,s}
  \ge
  b_{k,-}
  \sum_{s=t_k}^{t-1}
  \frac{1}{1+s-t_{\mathrm{stab}}}
  \ge
  b_{k,-}
  \log\!\left(
    \frac{1+t-t_{\mathrm{stab}}}
         {1+t_k-t_{\mathrm{stab}}}
  \right).
\]
Substituting into \eqref{eq:component_product_bound} gives
\[
  |c_{k,t}|
  \le
  |c_{k,t_k}|
  \left(
    \frac{1+t_k-t_{\mathrm{stab}}}
         {1+t-t_{\mathrm{stab}}}
  \right)^{b_{k,-}}.
\]
Thus \eqref{eq:component_poly_decay} holds with
\(\gamma_k=b_{k,-}\), and \(c_{k,t}\to0\).

Now suppose \(p_k>1\).  Then
\[
  \sum_{s=t_k}^{\infty}
  (1+s-t_{\mathrm{stab}})^{-p_k}<\infty.
\]
Using the upper bound in \eqref{eq:mk_time_tail}, we get
\[
  \sum_{s=t_k}^{\infty}\eta m_{k,s}<\infty,
\]
which proves \eqref{eq:finite_component_budget}.  The upper bound in
\eqref{eq:component_later_layer_plateau} is immediate from nonexpansion:
\[
  |c_{k,t}|\le |c_{k,t_k}|.
\]
For the lower bound, increase \(t_k\) if necessary so that
\(0\le \eta m_{k,s}\le 1/2\) for all \(s\ge t_k\).  Then
\[
  \log(1-\eta m_{k,s})
  \ge
  -2\eta m_{k,s}.
\]
Thus, for \(t\ge t_k\),
\[
\begin{aligned}
  |c_{k,t}|
  &=
  |c_{k,t_k}|
  \prod_{s=t_k}^{t-1}(1-\eta m_{k,s})  \\
  &\ge
  |c_{k,t_k}|
  \exp\!\left(
    -2\sum_{s=t_k}^{t-1}\eta m_{k,s}
  \right)  \\
  &\ge
  |c_{k,t_k}|
  \exp\!\left(
    -2\sum_{s=t_k}^{\infty}\eta m_{k,s}
  \right).
\end{aligned}
\]
The final exponential is a strictly positive constant.  This gives
\eqref{eq:component_later_layer_plateau} with
\[
  C_{k,-}
  :=
  \exp\!\left(
    -2\sum_{s=t_k}^{\infty}\eta m_{k,s}
  \right),
  \qquad
  C_{k,+}:=1.
\]

Finally, since
\[
  \sum_{s=t_k}^{\infty}\eta m_{k,s}<\infty
  \quad\text{and}\quad
  0\le \eta m_{k,s}<1,
\]
the infinite product
\[
  \prod_{s=t_k}^{\infty}(1-\eta m_{k,s})
\]
converges to a strictly positive number.  Therefore
\[
  c_{k,t}\to
  c_{k,t_k}
  \prod_{s=t_k}^{\infty}(1-\eta m_{k,s}),
\]
which is nonzero whenever \(c_{k,t_k}\neq0\).
\end{proof}

\begin{cor}[Spanning versus non-spanning component behavior]
\label{cor:componentwise_spanning_nonspanning}
Under Assumption~\ref{asmp:simultaneous_diag}, the stable tail separates
the active coordinates into two classes:
\[
  U_{\mathrm{sv}}
  :=
  \Span\{e_k:\ j(k)=1\},
  \qquad
  U_{\mathrm{late}}
  :=
  \Span\{e_k:\ j(k)>1\}.
\]
Writing
\[
  \tho_t={\tho_t}^{\mathrm{sv}}+{\tho_t}^{\mathrm{late}},
\]
one has
\[
  {\tho_t}^{\mathrm{sv}}\to0,
\]
with coordinatewise polynomial rates.  By contrast, each nonzero component
of \({\tho_t}^{\mathrm{late}}\) generally converges to a nonzero finite
limit determined by the entry value and the finite post-stability
contraction budget.

In particular, if \(j_*=1\), then \(U_{\mathrm{late}}=\{0\}\), and the
whole active normal component decays.  If \(j_*>1\), then
\(U_{\mathrm{late}}\neq\{0\}\), and the stable tail cannot in general
remove the later-layer component.
\end{cor}

\begin{proof}
The subspace \(U_{\mathrm{sv}}\) consists exactly of the coordinates with
\(p_k=1\), so Theorem~\ref{thm:componentwise_tail_dichotomy}(i) gives
coordinatewise convergence to zero.  The subspace \(U_{\mathrm{late}}\)
consists exactly of the coordinates with \(p_k>1\), so
Theorem~\ref{thm:componentwise_tail_dichotomy}(ii) gives a finite
contraction budget and, generically, nonzero limiting coordinates.

If \(j_*=1\), then the first layer spans all of \(U\), so every coordinate
has \(j(k)=1\).  Hence \(U_{\mathrm{late}}=\{0\}\).  If \(j_*>1\), some
coordinate is not seen by the first layer, so \(U_{\mathrm{late}}\neq\{0\}\).
\end{proof}

\begin{rem}[Interpretation for the non-spanning experiments]
\label{rem:componentwise_experiment_interpretation}
In the non-spanning logistic experiment (Figures \ref{fig:nonspanning}, \ref{fig:nonspanning_nonmonotone}, and \ref{fig:nonspanning_suppression}), the support-vector direction
corresponds to coordinates with \(j(k)=1\), hence it receives a harmonic
tail contraction budget and decays polynomially.  The weak direction is
invisible to the support-vector layer and first appears at a later
exponent \(\alpha_{j(k)}>\alpha_1\).  Its contraction budget is summable,
so the stable phase can reduce it by only a finite multiplicative factor.
The final residual is therefore determined by the value handed off from
the unstable phase together with this finite tail budget.
\end{rem}

\section{Multiclass Cross-Entropy}\label{app:cross_entropy}
\subsection{Setup}\label{app:ce_setup}

The multiclass cross-entropy analysis mirrors the binary logistic case
of Appendix~\ref{app:logistic_verification} step for step. The main
difference is that the orthogonal dynamics now have two sources. As in
the binary high-dimensional model, shared off-axis components \(\vu_i\)
create data-induced active directions. In addition, even in the collinear
case \(\vu_i=0\), the multi-column structure of the softmax weights creates
nontrivial matrix-valued orthogonal directions inside the simplex span.

\myparagraph{Data model}
Fix \(K\ge 3\) and assume \(m\ge K-1\). Let
\(r_1,\dots,r_K\in\mathbb R^m\) be the unit-norm vertices of a centered
regular simplex:
\[
\langle \vr_c,\vr_d\rangle
=
\begin{cases}
1, & c=d,\\[1mm]
-\dfrac{1}{K-1}, & c\neq d,
\end{cases}
\qquad
\sum_{c=1}^K \vr_c=0.
\]
Let
\[
S:=\Span\{r_1,\dots,r_K\},
\qquad
\dim S=K-1.
\]
For each \(i=1,\dots,N\), choose a margin scale \(b_i>0\) and a shared
orthogonal vector
\[
\vu_i\in S^\perp.
\]
For each class \(c\in\{1,\dots,K\}\), define
\[
\vx_{c,i}:=b_i \vr_c+\vu_i,
\qquad
y_{c,i}:=c.
\]
Thus each index \(i\) defines a full \(K\)-class orbit: the
class-dependent component \(b_i \vr_c\) rotates across the simplex
directions, while the orthogonal component \(\vu_i\) is shared across
classes.

This is the multiclass analogue of the binary pair-symmetric construction
\[
\vx_i^+=a_i v+\vu_i,
\qquad
\vx_i^-=-a_i v+\vu_i.
\]
The scalar \(b_i\) plays the role of the binary margin scale \(a_i\), and
the shared vector \(\vu_i\in S^\perp\) plays the same role as the binary
off-axis component \(\vu_i\). The collinear simplex model is recovered as
the special case \(\vu_i=0\) for all \(i\).

\myparagraph{Parameterization and loss}
We parameterize the classifier by a weight matrix
$W=[w_1,\dots,w_K]\in\mathbb{R}^{m\times K}$, working on the centered
linear space
\[
\mathcal{C} := \Big\{\,W\in\mathbb{R}^{m\times K} : \textstyle\sum_{c=1}^K \vw_c = 0\,\Big\},
\]
equipped with the Frobenius inner product
$\langle A,B\rangle_F := \operatorname{tr}(A^\top B)$. Gradient descent
reads
\[
W_{t+1} = W_t - \eta\,\nabla\Loss(W_t),
\]
where the empirical cross-entropy loss is
\begin{equation}\label{eq:ce_loss}
\Loss(W) := \frac{1}{KN}\sum_{c=1}^K\sum_{i=1}^N
\left[
-\langle \vw_c, \vx_{c,i}\rangle
+ \log\sum_{d=1}^K e^{\langle \vw_d, \vx_{c,i}\rangle}
\right].
\end{equation}

\myparagraph{Active slice and inactive directions}
Let
\[
U_{\mathrm{data}}:=\Span\{u_1,\dots,u_N\}\subseteq S^\perp,
\]
and define the full data span
\[
E:=S\oplus U_{\mathrm{data}}.
\]
Since every data point has the form
\[
\vx_{c,i}=b_i \vr_c+\vu_i\in E,
\]
the loss depends on each weight vector \(\vw_d\) only through its projection
onto \(E\).

We decompose the centered parameter space \(\mathcal C\) orthogonally as
\[
\mathcal C=\mathcal A\oplus\mathcal I,
\]
where
\begin{equation}\label{eq:ce_AI_def_shared}
\mathcal A
:=
\{W\in\mathcal C:\ \vw_c\in E\ \forall c\},
\qquad
\mathcal I
:=
\{W\in\mathcal C:\ \vw_c\in E^\perp\ \forall c\}.
\end{equation}
The space \(\mathcal A\) is the active slice: it consists of centered
class-weight matrices whose columns lie in the span of the data. The space
\(\mathcal I\) contains the inactive directions, which are orthogonal to
all data vectors and hence invisible to the loss.

\begin{prop}[Inactive directions are frozen]
\label{prop:ce_inactive_frozen}
For every \(W\in\mathcal C\), one has
\[
\nabla\Loss(W)\in\mathcal A.
\]
Consequently, writing
\[
W_t=W_t^{\mathrm{act}}+W_t^{\mathrm{inact}},
\qquad
W_t^{\mathrm{act}}\in\mathcal A,\quad
W_t^{\mathrm{inact}}\in\mathcal I,
\]
gradient descent freezes the inactive component:
\[
W_{t+1}^{\mathrm{inact}}=W_t^{\mathrm{inact}}
\qquad\forall\,t\ge 0.
\]
\end{prop}

\begin{proof}
Fix a sample \((\vx_{c,i},c)\). For this sample, the cross-entropy term is
\[
-\langle \vw_c,\vx_{c,i}\rangle
+
\log\!\left(\sum_{e=1}^K e^{\langle w_e,\vx_{c,i}\rangle}\right).
\]
Its gradient with respect to the \(d\)-th weight column is
\[
\partial_{\vw_d}
=
\bigl(p_{d\mid c,i}(W)-\mathbf 1_{\{d=c\}}\bigr)\vx_{c,i},
\]
where
\[
p_{d\mid c,i}(W)
:=
\frac{e^{\langle \vw_d,\vx_{c,i}\rangle}}
{\sum_{e=1}^K e^{\langle w_e,\vx_{c,i}\rangle}}.
\]
Since \(\vx_{c,i}=b_i \vr_c+\vu_i\in E\), every column of this sample gradient
belongs to \(E\).

Moreover, summing the sample gradient over columns gives
\[
\sum_{d=1}^K
\bigl(p_{d\mid c,i}(W)-\mathbf 1_{\{d=c\}}\bigr)\vx_{c,i}
=
\left(\sum_{d=1}^K p_{d\mid c,i}(W)-1\right)\vx_{c,i}
=0.
\]
Thus the sample gradient is centered and has all columns in \(E\). Hence
it belongs to \(\mathcal A\). Averaging over \(c\) and \(i\) gives
\[
\nabla\Loss(W)\in\mathcal A.
\]
Therefore the \(\mathcal I\)-component of gradient descent is unchanged:
\[
W_{t+1}^{\mathrm{inact}}
=
\Pi_{\mathcal I}(W_t-\eta\nabla\Loss(W_t))
=
W_t^{\mathrm{inact}},
\]
because \(\nabla\Loss(W_t)\in\mathcal A\perp\mathcal I\).
\end{proof}

The loss depends only on the active component \(W^{\mathrm{act}}\). Indeed,
for every inactive column \(\vw_d^{\mathrm{inact}}\in E^\perp\) and every
data point \(\vx_{c,i}\in E\),
\[
\langle \vw_d^{\mathrm{inact}},\vx_{c,i}\rangle=0.
\]
Hence
\[
\langle \vw_d,\vx_{c,i}\rangle
=
\langle \vw_d^{\mathrm{act}},\vx_{c,i}\rangle.
\]
Combined with Proposition~\ref{prop:ce_inactive_frozen}, this means that
the inactive component is dynamically irrelevant: it is fixed by gradient
descent and never affects the loss. Thus, after subtracting the fixed inactive component, the nontrivial
dynamics are completely described by the active variable
\(W_t^{\mathrm{act}}\). We therefore analyze the reduced dynamics on
\(\mathcal A\), equivalently assuming \(W_0\in\mathcal A\).

\myparagraph{Valley line and active subspace} Inside \(\mathcal A\), we take the symmetric parallel  direction to be the
configuration in which each weight column points along its own class axis. Setting
\[
R:=[r_1,\dots,r_K]\in\mathbb R^{m\times K},
\]
we have \(R\in\mathcal A\), since each \(\vr_c\in S\subset E\) and
\(\sum_{c=1}^K \vr_c=0\). Moreover,
\[
\|R\|_F^2=\sum_{c=1}^K\|\vr_c\|^2=K.
\]
Thus the unit-norm parallel  generator is
\[
v:=\frac{R}{\sqrt K}\in\mathcal A,
\qquad
\|v\|_F=1.
\]
We define the parallel  line and active orthogonal space by
\[
\mathcal M:=\Span\{v\},
\qquad
\mathcal U:=\{Z\in\mathcal A:\langle Z,v\rangle_F=0\}.
\]
Then
\[
\mathcal A=\mathcal M\oplus\mathcal U
\]
orthogonally. Every active iterate admits the canonical decomposition
\begin{equation}\label{eq:ce_decomp}
W_t=\theta_t^\parallel v+Z_t,
\qquad
\theta_t^\parallel:=\langle W_t,v\rangle_F,
\qquad
Z_t\in\mathcal U.
\end{equation}
This is the multiclass analogue of the binary decomposition
\(\bth=\thp\vv+\tho\), with the matrix-valued active orthogonal space
\(\mathcal U\) playing the role of the active normal space.

Furthermore, we can show that this subspace corresponds to the Crammer---Singer hard-margin max-margin direction of our symmetric data configuration.

\begin{prop}[Crammer--Singer max-margin direction]
\label{prop:ce_cs_svm_direction}
Consider the multiclass data model of Appendix~\ref{app:ce_setup}, and let
\[
b_{\min}:=\min_{1\le i\le N} b_i .
\]
Consider the centered Crammer--Singer hard-margin SVM problem
\begin{equation}\label{eq:ce_cs_svm}
\min_{W\in\mathcal C}\ \frac12\|W\|_F^2
\qquad
\text{subject to}\qquad
\langle w_c-w_d,x_{c,i}\rangle \ge 1
\quad
\forall i,\ \forall c,\ \forall d\neq c .
\end{equation}
Then its unique solution is
\begin{equation}\label{eq:ce_cs_solution}
W_{\mathrm{SVM}}
=
\frac{K-1}{K b_{\min}}\,R .
\end{equation}
Equivalently, since \(v=R/\sqrt K\), the Crammer--Singer max-margin
direction is \(v\), and hence the max-margin line is precisely
\[
\mathcal M=\Span\{v\}.
\]
\end{prop}

\begin{proof}
Let
\[
\alpha_*:=\frac{K-1}{K b_{\min}} .
\]
We first show that \(W=\alpha_*R\) is feasible. For a sample
\(x_{c,i}=b_i r_c+u_i\) and a competing class \(d\neq c\),
\[
\langle \alpha_* r_c-\alpha_* r_d,\, b_i r_c+u_i\rangle
=
\alpha_* b_i \langle r_c-r_d,r_c\rangle
+
\alpha_*\langle r_c-r_d,u_i\rangle .
\]
Since \(u_i\in S^\perp\) and \(r_c-r_d\in S\), the second term vanishes.
Using the simplex identities,
\[
\langle r_c-r_d,r_c\rangle
=
1-\left(-\frac{1}{K-1}\right)
=
\frac{K}{K-1}.
\]
Therefore
\[
\langle \alpha_* r_c-\alpha_* r_d,\, x_{c,i}\rangle
=
\alpha_* b_i \frac{K}{K-1}
=
\frac{b_i}{b_{\min}}
\ge 1.
\]
Thus \(\alpha_*R\) is feasible.

It remains to prove optimality. Let \(W=[w_1,\dots,w_K]\in\mathcal C\)
be any feasible point. Choose an index \(i_*\) such that
\(b_{i_*}=b_{\min}\), and write \(u_*:=u_{i_*}\). For each fixed class
\(c\), feasibility gives
\[
\langle w_c-w_d,\, b_{\min} r_c+u_*\rangle\ge 1
\qquad
\forall d\neq c .
\]
Summing over all \(d\neq c\), we obtain
\[
\left\langle
\sum_{d\neq c}(w_c-w_d),\, b_{\min} r_c+u_*
\right\rangle
\ge K-1 .
\]
Since \(W\in\mathcal C\), we have \(\sum_{d=1}^K w_d=0\), and hence
\[
\sum_{d\neq c}(w_c-w_d)
=
K w_c .
\]
Thus
\[
K\langle w_c,\, b_{\min} r_c+u_*\rangle
\ge K-1,
\]
or equivalently
\[
\langle w_c,\, b_{\min} r_c+u_*\rangle
\ge \frac{K-1}{K}.
\]
Summing this inequality over \(c=1,\dots,K\) yields
\[
\sum_{c=1}^K
\langle w_c,\, b_{\min} r_c+u_*\rangle
\ge K-1 .
\]
The shared off-axis term cancels because \(W\in\mathcal C\):
\[
\sum_{c=1}^K \langle w_c,u_*\rangle
=
\left\langle \sum_{c=1}^K w_c,u_*\right\rangle
=0.
\]
Therefore
\[
b_{\min}\sum_{c=1}^K\langle w_c,r_c\rangle
=
b_{\min}\langle W,R\rangle_F
\ge K-1.
\]
Hence
\[
\langle W,R\rangle_F
\ge
\frac{K-1}{b_{\min}}.
\]
By Cauchy--Schwarz and \(\|R\|_F=\sqrt K\),
\[
\|W\|_F\sqrt K
=
\|W\|_F\|R\|_F
\ge
\langle W,R\rangle_F
\ge
\frac{K-1}{b_{\min}}.
\]
Thus every feasible \(W\) satisfies
\[
\|W\|_F
\ge
\frac{K-1}{\sqrt K\,b_{\min}}.
\]
On the other hand,
\[
\|\alpha_*R\|_F
=
\alpha_*\sqrt K
=
\frac{K-1}{K b_{\min}}\sqrt K
=
\frac{K-1}{\sqrt K\,b_{\min}}.
\]
So \(\alpha_*R\) attains the minimum norm.

Finally, equality in Cauchy--Schwarz forces \(W\) to be a nonnegative
scalar multiple of \(R\). The margin constraints then force the scalar to
be exactly \(\alpha_*\). Hence the centered hard-margin SVM solution is
unique and equals \(\alpha_*R\).
\end{proof}

\begin{rem}[Origin of active orthogonal dynamics]
\label{rem:ce_orthogonal_origin}
Let
\[
q:=\dim U_{\mathrm{data}},
\qquad
U_{\mathrm{data}}=\Span\{u_1,\dots,u_N\}\subseteq S^\perp,
\]
so that
\[
\dim E=\dim S+\dim U_{\mathrm{data}}=(K-1)+q.
\]
Because \(\mathcal A\) consists of \(K\) columns in \(E\) subject to one
centering constraint in \(E\), we have
\[
\dim\mathcal A=(K-1)\dim E=(K-1)(K-1+q).
\]
Consequently,
\[
\dim\mathcal U=(K-1)(K-1+q)-1.
\]

Thus the active orthogonal directions have two sources. First, even when
\(q=0\), there are matrix-structured perturbations of the symmetric
configuration \(\vw_c\propto \vr_c\) inside the simplex span \(S\). Second,
when \(q>0\), the shared orthogonal features \(\vu_i\) create additional
data-induced active directions, directly mirroring the role of
\(\Span\{\vu_i\}\) in the binary high-dimensional model.

In the special collinear case \(\vu_i=0\) for all \(i\), one has \(q=0\) and
\[
\dim\mathcal U=(K-1)^2-1=K(K-2).
\]
For \(K=2\), this gives \(\dim\mathcal U=0\), recovering the binary
symmetric model with no off-axis components.
\end{rem}

\myparagraph{Binary--multiclass dictionary}
The setup above maps onto the binary case as follows.

\begin{center}
\renewcommand{\arraystretch}{1.2}
\begin{tabular}{ll}
\toprule
\textbf{Binary symmetric} &
\textbf{Multiclass simplex with shared orthogonal features} \\
\midrule
Class directions \(\{\vv,-\vv\}\)
& Simplex class directions \(\{r_1,\dots,r_K\}\) \\
Margin scalars \(a_i>0\)
& Margin scalars \(b_i>0\) \\
Data \(\vx_i^\pm=\pm a_i\vv+\vu_i\)
& Data \(\vx_{c,i}=b_i \vr_c+\vu_i\) \\
Shared orthogonal vector \(\vu_i\in \vv^\perp\)
& Shared orthogonal vector \(\vu_i\in S^\perp\) \\
Active data span \(\Span\{\vv\}\oplus\Span\{\vu_i\}\)
& Active data span \(S\oplus\Span\{\vu_i\}\) \\
Subspace \(\Span\{\vv\}\)
& Subspace \(\Span\{R/\sqrt K\}\) \\
Orthogonal Subspace \(U=\Span\{\vu_i\}\)
& Orthogonal Subspace \(\mathcal U=\mathcal A\cap(R/\sqrt K)^\perp\) \\
\bottomrule
\end{tabular}
\end{center}

Thus the one-dimensional restricted loss has the same form as in the binary
case, with the effective margin scale
\[
\tilde a_i:=\frac{\kappa b_i}{\sqrt K}
\]
replacing \(a_i\) in the restricted loss. The next subsection computes
the corresponding active sharpness operator \(H(\theta^\parallel)\),
whose block structure reflects both the simplex-span matrix directions
and the shared orthogonal features \(\vu_i\).

\subsection{Closed-form subspace restricted objects}
\label{app:ce_valley_objects}

We now compute the subspace-restricted loss, parallel drift, and active
sharpness operator for the multiclass simplex model with shared orthogonal
features. Throughout this subsection, recall that
\[
v=\frac{R}{\sqrt K},
\qquad
R=[r_1,\dots,r_K],
\qquad
\kappa:=\frac{K}{K-1}.
\]
For \(s\in\mathbb R\), define the one-dimensional multiclass profile
\[
\phi_K(s):=\log\!\bigl(1+(K-1)e^{-s}\bigr),
\qquad
q_K(s):=\frac{K-1}{e^s+K-1}=-\phi_K'(s).
\]
We also write
\[
\tilde a_i:=\frac{\kappa b_i}{\sqrt K}.
\]

For \(c\in\{1,\dots,K\}\) and \(s\in\mathbb R\), define the softmax
probability vector \(p^{(c)}(s)\in\mathbb R^K\) by
\[
p^{(c)}_d(s)
:=
\begin{cases}
1-q_K(s), & d=c,\\[1mm]
\dfrac{q_K(s)}{K-1}, & d\neq c.
\end{cases}
\]
Equivalently, \(p^{(c)}(s)\) is the softmax distribution obtained when
the correct class logit exceeds each incorrect class logit by \(s\).

For a vector \(p\in\mathbb R^K\) with nonnegative entries summing to one,
let
\[
\Cov(p):=\operatorname{Diag}(p)-pp^\top.
\]
Thus for \(h\in\mathbb R^K\),
\[
h^\top\Cov(p)h
=
\sum_{d=1}^K p_d h_d^2
-
\left(\sum_{d=1}^K p_d h_d\right)^2
=
\operatorname{Var}_{d\sim p}(h_d).
\]

\begin{prop}[Closed-form restricted objects]
\label{prop:ce_valley_objects}
For the multiclass cross-entropy loss \eqref{eq:ce_loss}, restricted to
the active slice \(\mathcal A\), the following identities hold.

\begin{enumerate}[label=\textup{(\alph*)},leftmargin=2.4em]

\item \textbf{Valley-restricted loss.}
For every \(\theta^\parallel\in\mathbb R\),
\begin{equation}\label{eq:ce_ell_closed}
\ell(\theta^\parallel)
:=
\Loss(\theta^\parallel v)
=
\frac1N\sum_{i=1}^N
\phi_K\!\left(\tilde a_i\theta^\parallel\right).
\end{equation}

\item \textbf{Valley drift.}
For every \(\theta^\parallel\in\mathbb R\),
\begin{equation}\label{eq:ce_ellprime_closed}
-\ell'(\theta^\parallel)
=
\frac1N\sum_{i=1}^N
\tilde a_i\,
q_K\!\left(\tilde a_i\theta^\parallel\right).
\end{equation}
In particular, \(-\ell'(\theta^\parallel)>0\) for all
\(\theta^\parallel\).

\item \textbf{Gradient alignment along the parallel direction.}
For every \(\theta^\parallel\in\mathbb R\),
\begin{equation}\label{eq:ce_grad_valley_parallel}
\nabla\Loss(\theta^\parallel v)
=
\ell'(\theta^\parallel)v.
\end{equation}
Equivalently, the active orthogonal gradient vanishes along the parallel direction:
\[
\Pi_{\mathcal U}\nabla\Loss(\theta^\parallel v)=0.
\]

\item \textbf{Active sharpness operator.}
Let
\[
H(\theta^\parallel)
:=
\Pi_{\mathcal U}\nabla^2\Loss(\theta^\parallel v)\Pi_{\mathcal U}
\]
be the Hessian restricted to the active orthogonal space. Then for every
\(Z\in\mathcal U\),
\begin{equation}\label{eq:ce_H_quadratic_closed}
\langle Z,H(\theta^\parallel)Z\rangle_F
=
\frac{1}{KN}
\sum_{i=1}^N\sum_{c=1}^K
h_{c,i}(Z)^\top
\Cov\!\left(p^{(c)}(\tilde a_i\theta^\parallel)\right)
h_{c,i}(Z),
\end{equation}
where
\[
h_{c,i}(Z)
:=
\begin{bmatrix}
\langle z_1,\vx_{c,i}\rangle\\
\vdots\\
\langle z_K,\vx_{c,i}\rangle
\end{bmatrix}
\in\mathbb R^K,
\qquad
Z=[z_1,\dots,z_K].
\]
In particular, \(H(\theta^\parallel)\) is symmetric positive
semidefinite on \(\mathcal U\).

\item \textbf{Derivative of the active sharpness.}
For every \(Z\in\mathcal U\),
\begin{equation}\label{eq:ce_Hprime_quadratic_closed}
\langle Z,H'(\theta^\parallel)Z\rangle_F
=
\frac{1}{KN}
\sum_{i=1}^N\sum_{c=1}^K
\tilde a_i\,
h_{c,i}(Z)^\top
\Cov'\!\left(p^{(c)}(\tilde a_i\theta^\parallel)\right)
h_{c,i}(Z),
\end{equation}
where \(\Cov'(p^{(c)}(s))\) denotes the derivative with respect to the
scalar margin variable \(s\):
\[
\Cov'\!\left(p^{(c)}(s)\right)
=
\operatorname{Diag}\!\left((p^{(c)})'(s)\right)
-
(p^{(c)})'(s)\,p^{(c)}(s)^\top
-
p^{(c)}(s)\,(p^{(c)})'(s)^\top,
\]
with
\[
(p^{(c)}_d)'(s)
=
\begin{cases}
q_K(s)\bigl(1-q_K(s)\bigr), & d=c,\\[1mm]
-\dfrac{q_K(s)\bigl(1-q_K(s)\bigr)}{K-1}, & d\neq c.
\end{cases}
\]

\end{enumerate}
\end{prop}

\begin{proof}
Fix \(i\) and \(c\). Along the parallel direction,
\[
W=\theta^\parallel v=\frac{\theta^\parallel}{\sqrt K}R,
\qquad
\vw_d=\frac{\theta^\parallel}{\sqrt K}\vr_d.
\]
Since \(\vu_i\in S^\perp\) and \(\vr_d\in S\), we have
\[
\langle \vr_d,\vu_i\rangle=0.
\]
Therefore
\[
\langle \vw_d,\vx_{c,i}\rangle
=
\frac{\theta^\parallel}{\sqrt K}
\langle \vr_d,b_i \vr_c+\vu_i\rangle
=
\frac{\theta^\parallel b_i}{\sqrt K}\langle \vr_d,\vr_c\rangle.
\]
Using the simplex identities,
\[
\langle \vr_d,\vr_c\rangle
=
\begin{cases}
1, & d=c,\\[1mm]
-\dfrac{1}{K-1}, & d\neq c,
\end{cases}
\]
we obtain
\[
\langle \vw_c,\vx_{c,i}\rangle
=
\frac{\theta^\parallel b_i}{\sqrt K},
\qquad
\langle \vw_d,\vx_{c,i}\rangle
=
-\frac{\theta^\parallel b_i}{(K-1)\sqrt K}
\quad(d\neq c).
\]
Thus the correct-class logit exceeds each incorrect-class logit by
\[
\frac{\theta^\parallel b_i}{\sqrt K}
+
\frac{\theta^\parallel b_i}{(K-1)\sqrt K}
=
\frac{\kappa b_i}{\sqrt K}\theta^\parallel
=
\tilde a_i\theta^\parallel.
\]
The per-sample cross-entropy is therefore
\[
-\frac{\theta^\parallel b_i}{\sqrt K}
+
\log\!\left(
e^{\theta^\parallel b_i/\sqrt K}
+
(K-1)e^{-\theta^\parallel b_i/((K-1)\sqrt K)}
\right).
\]
Factoring out \(e^{\theta^\parallel b_i/\sqrt K}\) gives
\[
\log\!\left(
1+(K-1)e^{-\tilde a_i\theta^\parallel}
\right)
=
\phi_K(\tilde a_i\theta^\parallel).
\]
Averaging over \(c\) and \(i\) proves \eqref{eq:ce_ell_closed}.
Differentiating yields
\[
\ell'(\theta^\parallel)
=
\frac1N\sum_{i=1}^N
\tilde a_i\phi_K'(\tilde a_i\theta^\parallel)
=
-\frac1N\sum_{i=1}^N
\tilde a_i q_K(\tilde a_i\theta^\parallel),
\]
which proves \eqref{eq:ce_ellprime_closed}.

We next prove the parallel-gradient alignment. For a sample
\((\vx_{c,i},c)\), the gradient with respect to the \(d\)-th column is
\[
\bigl(p_{d\mid c,i}(W)-\mathbf 1_{\{d=c\}}\bigr)\vx_{c,i}.
\]
Along the parallel direction, the softmax probabilities are precisely
\[
p_{d\mid c,i}(\theta^\parallel v)
=
p^{(c)}_d(\tilde a_i\theta^\parallel).
\]
Fix \(i\) and average the sample gradient over the \(K\) classes \(c\).
The \(\vu_i\)-component of the \(d\)-th column is proportional to
\[
\sum_{c=1}^K
\left(p^{(c)}_d(\tilde a_i\theta^\parallel)-\mathbf 1_{\{d=c\}}\right).
\]
For \(c=d\), this coefficient is \((1-q_K)-1=-q_K\), while for each
\(c\neq d\) it is \(q_K/(K-1)\). Hence the sum is
\[
-q_K+(K-1)\frac{q_K}{K-1}=0.
\]
Thus all shared orthogonal components cancel on the full class orbit.

For the simplex component, again fixing \(d\), the \(d\)-th gradient
column contributed by radius \(b_i\) is
\[
\frac{b_i}{K}
\sum_{c=1}^K
\left(p^{(c)}_d(\tilde a_i\theta^\parallel)-\mathbf 1_{\{d=c\}}\right)
\vr_c.
\]
Using \(\sum_{c\neq d}\vr_c=-\vr_d\), this equals
\[
\frac{b_i}{K}
\left(
-q_K \vr_d
+
\frac{q_K}{K-1}\sum_{c\neq d}\vr_c
\right)
=
-\frac{b_i q_K}{K}
\left(1+\frac1{K-1}\right)\vr_d
=
-\frac{b_i q_K}{K-1}\vr_d.
\]
Averaging over \(i\), we obtain
\[
\nabla_{\vw_d}\Loss(\theta^\parallel v)
=
-\frac1N\sum_{i=1}^N
\frac{b_i q_K(\tilde a_i\theta^\parallel)}{K-1}\vr_d.
\]
Since
\[
v=\frac1{\sqrt K}[r_1,\dots,r_K],
\]
and
\[
\ell'(\theta^\parallel)
=
-\frac1N\sum_{i=1}^N
\frac{\kappa b_i}{\sqrt K}q_K(\tilde a_i\theta^\parallel)
=
-\frac1N\sum_{i=1}^N
\frac{\sqrt K\, b_i}{K-1}q_K(\tilde a_i\theta^\parallel),
\]
we have
\[
\ell'(\theta^\parallel)v
=
\left[
-\frac1N\sum_{i=1}^N
\frac{b_i q_K(\tilde a_i\theta^\parallel)}{K-1}r_1,
\dots,
-\frac1N\sum_{i=1}^N
\frac{b_i q_K(\tilde a_i\theta^\parallel)}{K-1}r_K
\right],
\]
which proves \eqref{eq:ce_grad_valley_parallel}. Since
\(\mathcal U\perp v\), this also gives
\(\Pi_{\mathcal U}\nabla\Loss(\theta^\parallel v)=0\).

It remains to compute the Hessian. For a perturbation
\(Z=[z_1,\dots,z_K]\), the logit perturbation vector for sample
\((\vx_{c,i},c)\) is
\[
h_{c,i}(Z)
=
\bigl(\langle z_1,\vx_{c,i}\rangle,\dots,
\langle z_K,\vx_{c,i}\rangle\bigr)^\top.
\]
The second directional derivative of the sample cross-entropy in the
direction \(Z\) is the softmax covariance of these logit perturbations:
\[
h_{c,i}(Z)^\top
\Cov\!\left(p^{(c)}(\tilde a_i\theta^\parallel)\right)
h_{c,i}(Z).
\]
Averaging over \(c\) and \(i\), and restricting to \(Z\in\mathcal U\),
gives \eqref{eq:ce_H_quadratic_closed}. This also shows that
\(H(\theta^\parallel)\) is symmetric positive semidefinite.

Finally, differentiating \eqref{eq:ce_H_quadratic_closed} with respect to
\(\theta^\parallel\) gives \eqref{eq:ce_Hprime_quadratic_closed}, because
the only \(\theta^\parallel\)-dependence is through
\(s_i=\tilde a_i\theta^\parallel\). The derivative of \(p^{(c)}\) follows
from
\[
q_K'(s)=-q_K(s)(1-q_K(s)).
\]
This completes the proof.
\end{proof}

\subsection{Verification of the abstract assumptions}
\label{app:ce_verify_assumptions}

We now verify the constants entering the abstract pre- and post-stability
theorems for the multiclass simplex model. The verification is completely
parallel to the binary logistic case, except that the active sharpness is
matrix-valued. Accordingly, the full  contraction statement in
the post-stability phase requires a tail conditioning assumption on the
active sharpness operator.

Recall the effective margin scales
\[
\tilde a_i:=\frac{\kappa b_i}{\sqrt K},
\qquad
\kappa:=\frac{K}{K-1},
\]
and define
\[
\tilde a_{\min}:=\min_{1\le i\le N}\tilde a_i,
\qquad
\tilde a_{\max}:=\max_{1\le i\le N}\tilde a_i,
\qquad
I_*:=\{i:\tilde a_i=\tilde a_{\min}\}.
\]
We also define the maximum data norm
\[
X_{\max}:=\max_{1\le i\le N,\ 1\le c\le K}\|\vx_{c,i}\|.
\]
Since \(\vx_{c,i}=b_i \vr_c+\vu_i\) with \(\vu_i\perp \vr_c\) and \(\|\vr_c\|=1\),
this is equivalently
\[
X_{\max}=\max_{1\le i\le N}\sqrt{b_i^2+\|\vu_i\|^2}.
\]

We use the multiclass profile
\[
\phi_K(s):=\log\!\bigl(1+(K-1)e^{-s}\bigr),
\qquad
q_K(s):=\frac{K-1}{e^s+K-1}.
\]
Its curvature satisfies
\[
\phi_K''(s)=\frac{(K-1)e^s}{(e^s+K-1)^2},
\]
and
\[
-\frac{d}{ds}\log \phi_K''(s)
=
\tanh\!\left(\frac{s-\log(K-1)}{2}\right).
\]
Thus the curvature is decreasing once \(s\ge \log(K-1)\).

For the oscillation slab, fix \(r>0\), an initial value
\(\theta_0^\parallel\), and the corrected stability anchor
\[
\theta_*^{(r)}
:=
\inf\Bigl\{\theta^\parallel\ge \theta_0^\parallel:\;
\eta(1+\delta_r)\lambda_1(\theta^\parallel)\le 2\Bigr\},
\]
where
\[
\lambda_1(\theta^\parallel)
:=
\lambda_{\max}\bigl(H(\theta^\parallel)|_{\mathcal U}\bigr).
\]
Assume throughout this verification that the oscillation slab lies to
the right of the multiclass curvature threshold:
\begin{equation}\label{eq:ce_threshold_condition}
\tilde a_{\min}\theta_0^\parallel\ge \log(K-1).
\end{equation}

\begin{prop}[Subspace alignment and parallel expansion]
\label{prop:ce_alignment_parallel_expansion}
Let
\[
G_{\mathcal U}(\theta^\parallel,Z)
:=
\Pi_{\mathcal U}\nabla\Loss(\theta^\parallel v+Z),
\qquad
g_\parallel(\theta^\parallel,Z)
:=
\langle \nabla\Loss(\theta^\parallel v+Z),v\rangle_F,
\]
for \(\theta^\parallel\in\mathbb R\) and \(Z\in\mathcal U\). Then the
following hold.

\begin{enumerate}[label=\textup{(\alph*)},leftmargin=2.4em]

\item \textbf{Valley alignment.}
For every \(\theta^\parallel\in\mathbb R\),
\[
\nabla\Loss(\theta^\parallel v)=\ell'(\theta^\parallel)v.
\]
Equivalently,
\[
G_{\mathcal U}(\theta^\parallel,0)=0.
\]

\item \textbf{Active orthogonal Jacobian along the parallel direction.}
For every \(\theta^\parallel\in\mathbb R\),
\[
D_ZG_{\mathcal U}(\theta^\parallel,0)=H(\theta^\parallel),
\]
where
\[
H(\theta^\parallel)
:=
\Pi_{\mathcal U}\nabla^2\Loss(\theta^\parallel v)\Pi_{\mathcal U}.
\]

\item \textbf{Parallel Taylor expansion.}
Fix a compact interval \(J\subset\mathbb R\) and \(r>0\). Then there exists
a finite constant \(c_{\parallel,r,J}<\infty\) such that, for every
\(\theta^\parallel\in J\) and every \(Z\in\mathcal U\) with
\(\|Z\|_F\le r\),
\begin{equation}\label{eq:ce_parallel_taylor}
g_\parallel(\theta^\parallel,Z)
=
\ell'(\theta^\parallel)
+
\frac12\langle Z,H'(\theta^\parallel)Z\rangle_F
+
R_\parallel(\theta^\parallel,Z),
\end{equation}
where
\begin{equation}\label{eq:ce_parallel_remainder}
|R_\parallel(\theta^\parallel,Z)|
\le
c_{\parallel,r,J}\|Z\|_F^3.
\end{equation}

\end{enumerate}
\end{prop}

\begin{proof}
Part \textup{(a)} is exactly the gradient-alignment identity from
Proposition~\ref{prop:ce_valley_objects}. In particular,
\[
\Pi_{\mathcal U}\nabla\Loss(\theta^\parallel v)
=
\Pi_{\mathcal U}\bigl(\ell'(\theta^\parallel)v\bigr)
=0,
\]
because \(\mathcal U\perp v\).

For part \textup{(b)}, differentiate
\[
G_{\mathcal U}(\theta^\parallel,Z)
=
\Pi_{\mathcal U}\nabla\Loss(\theta^\parallel v+Z)
\]
with respect to \(Z\) at \(Z=0\). Since \(\Pi_{\mathcal U}\) is constant
and \(Z\in\mathcal U\), this gives
\[
D_ZG_{\mathcal U}(\theta^\parallel,0)
=
\Pi_{\mathcal U}\nabla^2\Loss(\theta^\parallel v)\Pi_{\mathcal U}
=
H(\theta^\parallel).
\]

It remains to prove the parallel expansion. Fix
\(\theta^\parallel\in J\) and define
\[
F_{\theta^\parallel}(Z)
:=
g_\parallel(\theta^\parallel,Z)
=
\langle \nabla\Loss(\theta^\parallel v+Z),v\rangle_F,
\qquad Z\in\mathcal U.
\]
Since the cross-entropy loss is smooth, \(F_{\theta^\parallel}\) is smooth
in \(Z\). Taylor's theorem at \(Z=0\) gives
\[
F_{\theta^\parallel}(Z)
=
F_{\theta^\parallel}(0)
+
D_ZF_{\theta^\parallel}(0)[Z]
+
\frac12 D_Z^2F_{\theta^\parallel}(0)[Z,Z]
+
R_\parallel(\theta^\parallel,Z).
\]
The zeroth-order term is
\[
F_{\theta^\parallel}(0)
=
\langle \nabla\Loss(\theta^\parallel v),v\rangle_F
=
\ell'(\theta^\parallel).
\]
The first-order term vanishes. Indeed,
\[
D_ZF_{\theta^\parallel}(0)[Z]
=
\langle \nabla^2\Loss(\theta^\parallel v)Z,v\rangle_F
=
\langle Z,\nabla^2\Loss(\theta^\parallel v)v\rangle_F.
\]
But
\[
\nabla\Loss(\theta^\parallel v)=\ell'(\theta^\parallel)v.
\]
Differentiating this identity with respect to \(\theta^\parallel\) gives
\[
\nabla^2\Loss(\theta^\parallel v)v
=
\ell''(\theta^\parallel)v.
\]
Since \(Z\in\mathcal U\perp v\), it follows that
\[
D_ZF_{\theta^\parallel}(0)[Z]=0.
\]

For the quadratic term, note that
\[
D_Z^2F_{\theta^\parallel}(0)[Z,Z]
=
\left\langle D^3\Loss(\theta^\parallel v)[Z,Z],v\right\rangle_F.
\]
On the other hand,
\[
H(\theta^\parallel)
=
\Pi_{\mathcal U}\nabla^2\Loss(\theta^\parallel v)\Pi_{\mathcal U}.
\]
Therefore, for \(Z\in\mathcal U\),
\[
\langle Z,H(\theta^\parallel)Z\rangle_F
=
\langle Z,\nabla^2\Loss(\theta^\parallel v)Z\rangle_F.
\]
Differentiating with respect to \(\theta^\parallel\) yields
\[
\langle Z,H'(\theta^\parallel)Z\rangle_F
=
\left\langle D^3\Loss(\theta^\parallel v)[v,Z],Z\right\rangle_F
=
\left\langle D^3\Loss(\theta^\parallel v)[Z,Z],v\right\rangle_F,
\]
where the last equality uses symmetry of the third derivative tensor.
Thus
\[
D_Z^2F_{\theta^\parallel}(0)[Z,Z]
=
\langle Z,H'(\theta^\parallel)Z\rangle_F.
\]
This proves the expansion up to third order.

Finally, because \(\Loss\) is smooth and the set
\[
\{\theta^\parallel v+sZ:\ \theta^\parallel\in J,\ \|Z\|_F\le r,\ s\in[0,1]\}
\]
is compact, the third derivative of \(F_{\theta^\parallel}\) with respect
to \(Z\) is uniformly bounded there. Hence Taylor's theorem gives a
constant \(c_{\parallel,r,J}<\infty\) such that
\[
|R_\parallel(\theta^\parallel,Z)|
\le
c_{\parallel,r,J}\|Z\|_F^3.
\]
This completes the proof.
\end{proof}
\begin{prop}[Verification constants for multiclass cross-entropy]
\label{prop:ce_verify_assumptions}
Assume \eqref{eq:ce_threshold_condition}. Then the following constants
verify the abstract assumptions for the multiclass simplex model on the
tubular neighborhood
\[
\bigl\{
\theta^\parallel v+Z:\
\theta^\parallel\in[\theta_0^\parallel,\theta_*^{(r)}],
\ \|Z\|_F\le r
\bigr\}
\]
and on the corresponding tail half-tubular neighborhood
\[
\bigl\{
\theta^\parallel v+Z:\
\theta^\parallel\ge \theta_*^{(r)},\ \|Z\|_F\le r
\bigr\}.
\]

\begin{enumerate}[label=\textup{(\alph*)},leftmargin=2.5em]

\item \textbf{Relative curvature comparability.}
A sufficient RC tolerance is
\begin{equation}\label{eq:ce_delta_r}
\delta_r:=e^{\sqrt{2} X_{\max}r}-1.
\end{equation}
In particular, the admissibility condition \(\delta_r<1\) is implied by
\[
\sqrt{2} X_{\max}r<\log 2.
\]

\item \textbf{Matrix log-derivative lower bound.}
On the oscillation slab,
\[
-H'(\theta^\parallel)\succeq \beta H(\theta^\parallel)
\qquad\text{on }\: \mathcal U,
\]
where
\[
\beta
:=
\tilde a_{\min}
\tanh\!\left(
\frac{\tilde a_{\min}\theta_0^\parallel-\log(K-1)}{2}
\right).
\]
Under \eqref{eq:ce_threshold_condition}, this gives \(\beta\ge 0\), and it
is strictly positive if
\[
\tilde a_{\min}\theta_0^\parallel>\log(K-1).
\]

\item \textbf{Valley drift lower bound.}
The restricted loss satisfies
\[
-\ell'(\theta^\parallel)
=
\frac1N\sum_{i=1}^N
\tilde a_i q_K(\tilde a_i\theta^\parallel),
\]
so on the oscillation slab one may take
\begin{equation}\label{eq:ce_mell}
m_\ell
:=
\frac1N\sum_{i=1}^N
\tilde a_i q_K(\tilde a_i\theta_*^{(r)}).
\end{equation}

\item \textbf{Uniform upper bound on tubular neighborhood-level parallel drift.}
For every \(\theta^\parallel\in\mathbb R\) and every \(Z\in\mathcal U\),
\[
-\langle \nabla\Loss(\theta^\parallel v+Z),v\rangle_F
\le
G_\parallel,
\qquad
G_\parallel
:=
\frac1N\sum_{i=1}^N \tilde a_i.
\]

\item \textbf{Tail drift constants.}
Set $\beta_{\mathrm{tail}}:=\tilde a_{\min},$
and $\tau_r:=\sqrt{2}X_{\max}r.$
If $\tilde a_{\min}\theta_*^{(r)}-\tau_r
\ge
\log(K-1),$ then for every \(\theta^\parallel\ge\theta_*^{(r)}\), and every
\(Z\in\mathcal U\) with \(\|Z\|_F\le r\), the parallel drift satisfies
\[
A_-e^{-\beta_{\mathrm{tail}}(\theta^\parallel-\theta_*^{(r)})}
\le
-\langle \nabla\Loss(\theta^\parallel v+Z),v\rangle_F
\le
A_+e^{-\beta_{\mathrm{tail}}(\theta^\parallel-\theta_*^{(r)})},
\]
with
\begin{align}
A_-
&:=
e^{-\tau_r}
\frac{K-1}{2N}
\sum_{i\in I_*}
\tilde a_i e^{-\tilde a_i\theta_*^{(r)}},
\label{eq:ce_Aminus}
\\
A_+
&:=
e^{\tau_r}
\frac{K-1}{N}
\sum_{i=1}^N
\tilde a_i e^{-\tilde a_{\min}\theta_*^{(r)}}.
\label{eq:ce_Aplus}
\end{align}

\item \textbf{Tail sharpness constants.}
For \(i=1,\dots,N\), define the positive semidefinite operator
\(\mathsf B_i:\mathcal U\to\mathcal U\) by the quadratic form
\begin{equation}\label{eq:ce_Bi_def}
\langle Z,\mathsf B_i Z\rangle_F
:=
\frac1{KN}
\sum_{c=1}^K\sum_{d\neq c}
\left(
\langle z_d-z_c,\vx_{c,i}\rangle
\right)^2,
\qquad
Z=[z_1,\dots,z_K]\in\mathcal U.
\end{equation}
Let
\[
\mathsf B_*:=\sum_{i\in I_*}\mathsf B_i,
\qquad
\mathsf B_{\mathrm{all}}:=\sum_{i=1}^N\mathsf B_i.
\]
Then, for every \(\theta^\parallel\ge\theta_*^{(r)}\),
\[
\lambda_{1,-}
e^{-\beta_{\mathrm{tail}}(\theta^\parallel-\theta_*^{(r)})}
\le
\lambda_1(\theta^\parallel)
\le
\lambda_{1,+}
e^{-\beta_{\mathrm{tail}}(\theta^\parallel-\theta_*^{(r)})},
\]
where one may take
\begin{align}
\lambda_{1,-}
&:=
\frac14
e^{-\tilde a_{\min}\theta_*^{(r)}}
\lambda_{\max}\!\left(\mathsf B_*|_{\mathcal U}\right),
\label{eq:ce_lambda1minus}
\\
\lambda_{1,+}
&:=
2
e^{-\tilde a_{\min}\theta_*^{(r)}}
\lambda_{\max}\!\left(\mathsf B_{\mathrm{all}}|_{\mathcal U}\right).
\label{eq:ce_lambda1plus}
\end{align}

\item \textbf{Unstable phase conditioning, when needed.}
If
\[
H(\theta^\parallel)|_{\mathcal U}\succ 0
\qquad
\forall \theta^\parallel\in[\theta_0^\parallel,\theta_*^{(r)}],
\]
then compactness and continuity imply that
\[
\rho_{\mathrm{pre}}
:=
\inf_{\theta^\parallel\in[\theta_0^\parallel,\theta_*^{(r)}]}
\frac{\lambda_{\min}(H(\theta^\parallel)|_{\mathcal U})}
{\lambda_{\max}(H(\theta^\parallel)|_{\mathcal U})}
>0,
\]
and hence
\[
H(\theta^\parallel)
\succeq
\rho_{\mathrm{pre}}\lambda_1(\theta^\parallel)I_{\mathcal U}
\qquad
\forall \theta^\parallel\in[\theta_0^\parallel,\theta_*^{(r)}].
\]

\item \textbf{Tail conditioning, when applicable.}
If the slowest-layer operator is positive definite on the active
orthogonal space,
\begin{equation}\label{eq:ce_tail_spanning_operator}
\mathsf B_*|_{\mathcal U}\succ 0,
\end{equation}
then the active sharpness satisfies the tail conditioning bound
\[
H(\theta^\parallel)
\succeq
\rho_{\mathrm{tail}}\lambda_1(\theta^\parallel)I_{\mathcal U}
\qquad
\forall\,\theta^\parallel\ge\theta_*^{(r)},
\]
with
\begin{equation}\label{eq:ce_rhotail}
\rho_{\mathrm{tail}}
:=
\frac{
\lambda_{\min}\!\left(\mathsf B_*|_{\mathcal U}\right)
}{
8\,\lambda_{\max}\!\left(\mathsf B_{\mathrm{all}}|_{\mathcal U}\right)
}.
\end{equation}
Consequently, the full  three-phase post-stability theorem
applies under \eqref{eq:ce_tail_spanning_operator}.
\end{enumerate}
\end{prop}

\begin{proof}
We verify each item.

\smallskip\noindent
\emph{Relative curvature comparability.}
Fix a sample \((\vx_{c,i},c)\). For a perturbation
\(Z=[z_1,\dots,z_K]\), define the logit perturbations
\[
\Delta_d:=\langle z_d,\vx_{c,i}\rangle,\qquad d=1,\dots,K.
\]
Softmax probabilities and their covariance are invariant under adding the
same scalar to all logits. Hence only the oscillation of
\(\Delta=(\Delta_1,\dots,\Delta_K)\) matters. For every \(d,e\),
\[
|\Delta_d-\Delta_e|
=
|\langle z_d-z_e,\vx_{c,i}\rangle|
\le
\|z_d-z_e\|\,\|\vx_{c,i}\|
\le
\sqrt2\,\|Z\|_F X_{\max}
\le
\sqrt2 X_{\max}r.
\]
Thus, after subtracting a common scalar from the perturbed logits, every
logit perturbation lies in an interval of length at most
\(\sqrt2X_{\max}r\). Consequently the perturbed softmax distribution is
within density ratio
\[
C_r:=e^{\sqrt2X_{\max}r}
\]
of the parallel softmax distribution.

Next, we will need to invoke the following lemma (proof is one line using $h^\top\Cov(p)h
=
\min_{a\in\mathbb R}\sum_d p_d(h_d-a)^2.$:

\begin{lem}[Covariance comparability under bounded density ratios]
Let \(p,\widetilde p\in\Delta_K\) satisfy
\[
C^{-1}\le \frac{\widetilde p_d}{p_d}\le C
\qquad\forall d.
\]
Then for every \(h\in\mathbb R^K\),
\[
C^{-1}h^\top\Cov(p)h
\le
h^\top\Cov(\widetilde p)h
\le
C h^\top\Cov(p)h.
\]
\end{lem}

By covariance comparability under bounded density ratios,
\[
C_r^{-1}\,
h^\top\Cov(p)h
\le
h^\top\Cov(\widetilde p)h
\le
C_r\,
h^\top\Cov(p)h
\]
for every \(h\in\mathbb R^K\). Applying this samplewise and averaging over
\((c,i)\) gives
\[
e^{-\sqrt2X_{\max}r}H(\theta^\parallel)
\preceq
D_ZG_{\mathcal U}(\theta^\parallel,Z)
\preceq
e^{\sqrt2X_{\max}r}H(\theta^\parallel).
\]
Therefore the matrix RC condition holds with
\[
\delta_r:=e^{\sqrt2X_{\max}r}-1,
\]
since
\[
1-\delta_r=2-e^{\sqrt2X_{\max}r}\le e^{-\sqrt2X_{\max}r}.
\]

\smallskip\noindent
\emph{Log-derivative lower bound.}
We prove the matrix inequality
\[
-H'(\theta^\parallel)\succeq \beta H(\theta^\parallel)
\qquad\text{on }\mathcal U.
\]
Fix a class \(c\) and write \(q=q_K(s)\). Then
\[
p^{(c)}_c(s)=1-q,
\qquad
p^{(c)}_d(s)=\frac{q}{K-1}\quad(d\neq c).
\]
For every \(h\in\mathbb R^K\), the softmax covariance admits the pairwise
representation
\[
h^\top\Cov(p^{(c)}(s))h
=
\sum_{d<e}p^{(c)}_d(s)p^{(c)}_e(s)(h_d-h_e)^2.
\]
There are two types of pair weights. If one index is the correct class
\(c\), then
\[
p^{(c)}_c(s)p^{(c)}_d(s)
=
(1-q)\frac{q}{K-1}
\qquad(d\neq c),
\]
and therefore
\[
-\frac{d}{ds}
\log\!\left((1-q)q\right)
=
1-2q
=
\tanh\!\left(\frac{s-\log(K-1)}{2}\right).
\]
If both indices are incorrect, then
\[
p^{(c)}_d(s)p^{(c)}_e(s)
=
\frac{q^2}{(K-1)^2}
\qquad(d,e\neq c,\ d\neq e),
\]
and hence
\[
-\frac{d}{ds}\log(q^2)
=
2(1-q).
\]
On the region \(s\ge \log(K-1)\), we have \(q\le 1/2\), so
\[
2(1-q)\ge 1\ge 1-2q.
\]
Thus every pair weight in the covariance representation satisfies
\[
-\frac{d}{ds}\log\!\left(p^{(c)}_d(s)p^{(c)}_e(s)\right)
\ge
\tanh\!\left(\frac{s-\log(K-1)}{2}\right).
\]
Consequently, for every \(h\in\mathbb R^K\),
\[
-h^\top \frac{d}{ds}\Cov(p^{(c)}(s))h
\ge
\tanh\!\left(\frac{s-\log(K-1)}{2}\right)
h^\top\Cov(p^{(c)}(s))h.
\]

Now take \(s_i(\theta^\parallel)=\tilde a_i\theta^\parallel\). On the
oscillation slab, assumption
\[
\tilde a_{\min}\theta_0^\parallel\ge \log(K-1)
\]
implies \(s_i(\theta^\parallel)\ge \log(K-1)\) for all \(i\) and all
\(\theta^\parallel\in[\theta_0^\parallel,\theta_*^{(r)}]\). Hence
\[
-h^\top
\frac{d}{d\theta^\parallel}
\Cov(p^{(c)}(s_i(\theta^\parallel)))h
\ge
\tilde a_i
\tanh\!\left(
\frac{\tilde a_i\theta^\parallel-\log(K-1)}{2}
\right)
h^\top\Cov(p^{(c)}(s_i(\theta^\parallel)))h.
\]
Since the function
\[
a\mapsto
a\tanh\!\left(\frac{a\theta^\parallel-\log(K-1)}{2}\right)
\]
is increasing for \(a>0\) whenever \(a\theta^\parallel\ge \log(K-1)\), and
\(\theta^\parallel\ge\theta_0^\parallel\), the right-hand coefficient is
bounded below by
\[
\beta
:=
\tilde a_{\min}
\tanh\!\left(
\frac{\tilde a_{\min}\theta_0^\parallel-\log(K-1)}{2}
\right).
\]
Substituting into the quadratic-form expression for \(H\) and summing over
\(i\) and \(c\) gives
\[
-\langle Z,H'(\theta^\parallel)Z\rangle_F
\ge
\beta\langle Z,H(\theta^\parallel)Z\rangle_F
\qquad
\forall Z\in\mathcal U.
\]
Equivalently,
\[
-H'(\theta^\parallel)\succeq \beta H(\theta^\parallel)
\quad\text{on }\mathcal U.
\]
This proves the stated log-derivative lower bound.

\smallskip\noindent
\emph{Valley drift constants.}
By Proposition~\ref{prop:ce_valley_objects},
\[
-\ell'(\theta^\parallel)
=
\frac1N\sum_{i=1}^N
\tilde a_i q_K(\tilde a_i\theta^\parallel).
\]
Since \(q_K\) is positive and decreasing, the infimum on
\([\theta_0^\parallel,\theta_*^{(r)}]\) is attained at
\(\theta_*^{(r)}\), giving \eqref{eq:ce_mell}. The upper bound follows from \(q_K(s)\le1\).

\smallskip\noindent
\emph{Tail drift on the tubular neighborhood.}
Fix \(\theta^\parallel\ge\theta_*^{(r)}\) and
\(Z=[z_1,\dots,z_K]\in\mathcal U\) with \(\|Z\|_F\le r\). For a sample
\((\vx_{c,i},c)\), define the softmax probabilities at
\(W=\theta^\parallel v+Z\) by
\[
p_{d\mid c,i}(W)
:=
\frac{\exp(\langle \vw_d,\vx_{c,i}\rangle)}
{\sum_{e=1}^K\exp(\langle w_e,\vx_{c,i}\rangle)}.
\]
We first compute the contribution to the parallel drift. Since
\(v=R/\sqrt K\), one has
\[
-\langle \nabla\Loss(W),v\rangle_F
=
\frac{1}{KN}
\sum_{i=1}^N\sum_{c=1}^K
\frac{b_i}{\sqrt K}
\sum_{d=1}^K
\bigl(\mathbf 1_{\{d=c\}}-p_{d\mid c,i}(W)\bigr)
\langle \vr_d,\vr_c\rangle.
\]
Using the simplex identities,
\[
\sum_{d=1}^K
\bigl(\mathbf 1_{\{d=c\}}-p_{d\mid c,i}(W)\bigr)
\langle \vr_d,\vr_c\rangle
=
\frac{K}{K-1}\bigl(1-p_{c\mid c,i}(W)\bigr).
\]
Therefore
\begin{equation}\label{eq:ce_parallel_drift_error_prob}
-g_\parallel(\theta^\parallel,Z)
=
\frac{1}{KN}
\sum_{i=1}^N\sum_{c=1}^K
\tilde a_i
\bigl(1-p_{c\mid c,i}(\theta^\parallel v+Z)\bigr).
\end{equation}

It remains to bound the correct-class error probability. For each
incorrect class \(d\neq c\), the correct-vs-incorrect logit gap is
\[
\langle \vw_c-\vw_d,\vx_{c,i}\rangle
=
\tilde a_i\theta^\parallel
+
\langle z_c-z_d,\vx_{c,i}\rangle.
\]
Because \(\|Z\|_F\le r\),
\[
\|z_c-z_d\|
\le
\sqrt2\|Z\|_F
\le
\sqrt2 r,
\]
and hence
\[
|\langle z_c-z_d,\vx_{c,i}\rangle|
\le
\sqrt2 X_{\max}r
=
\tau_r.
\]
Thus every incorrect-class exponential ratio satisfies
\[
e^{-(\tilde a_i\theta^\parallel+\tau_r)}
\le
\exp\!\left(
-\langle \vw_c-\vw_d,\vx_{c,i}\rangle
\right)
\le
e^{-(\tilde a_i\theta^\parallel-\tau_r)}.
\]
Set
\[
S_{c,i}(W)
:=
\sum_{d\neq c}
\exp\!\left(
-\langle \vw_c-\vw_d,\vx_{c,i}\rangle
\right).
\]
Then
\[
(K-1)e^{-\tilde a_i\theta^\parallel-\tau_r}
\le
S_{c,i}(W)
\le
(K-1)e^{-\tilde a_i\theta^\parallel+\tau_r}.
\]
Moreover,
\[
1-p_{c\mid c,i}(W)
=
\frac{S_{c,i}(W)}{1+S_{c,i}(W)}.
\]
The tail tubular neighborhood margin condition
\[
\tilde a_{\min}\theta_*^{(r)}-\tau_r\ge\log(K-1)
\]
implies
\[
S_{c,i}(W)\le 1
\]
for all \(\theta^\parallel\ge\theta_*^{(r)}\), all \(i\), and all \(c\).
Therefore
\[
\frac12S_{c,i}(W)
\le
1-p_{c\mid c,i}(W)
\le
S_{c,i}(W).
\]
Consequently,
\begin{equation}\label{eq:ce_error_prob_tail_tubular neighborhood}
\frac{K-1}{2}
e^{-\tilde a_i\theta^\parallel-\tau_r}
\le
1-p_{c\mid c,i}(W)
\le
(K-1)e^{-\tilde a_i\theta^\parallel+\tau_r}.
\end{equation}

Substituting \eqref{eq:ce_error_prob_tail_tubular neighborhood} into
\eqref{eq:ce_parallel_drift_error_prob}, and restricting the lower bound
to \(i\in I_*\), gives
\[
-g_\parallel(\theta^\parallel,Z)
\ge
e^{-\tau_r}
\frac{K-1}{2N}
\sum_{i\in I_*}
\tilde a_i e^{-\tilde a_i\theta^\parallel}.
\]
Since \(\tilde a_i=\tilde a_{\min}\) for \(i\in I_*\), this equals
\[
A_-^{\mathrm{tubular neighborhood}}
e^{-\tilde a_{\min}(\theta^\parallel-\theta_*^{(r)})}.
\]
Similarly,
\[
-g_\parallel(\theta^\parallel,Z)
\le
e^{\tau_r}
\frac{K-1}{N}
\sum_{i=1}^N
\tilde a_i e^{-\tilde a_i\theta^\parallel}
\le
e^{\tau_r}
\frac{K-1}{N}
\sum_{i=1}^N
\tilde a_i e^{-\tilde a_{\min}\theta^\parallel},
\]
which is
\[
A_+^{\mathrm{tubular neighborhood}}
e^{-\tilde a_{\min}(\theta^\parallel-\theta_*^{(r)})}.
\]
This proves the tubular neighborhood-level tail drift bounds.

\smallskip\noindent
\emph{Tail sharpness.}
Let \(p^{(c)}(s)\) be the parallel softmax distribution for class \(c\),
with correct-class error probability \(q_K(s)\). For any vector
\(h\in\mathbb R^K\), the softmax covariance admits the pairwise
representation
\[
h^\top\Cov(p^{(c)}(s))h
=
\sum_{d<e}p^{(c)}_d(s)p^{(c)}_e(s)(h_d-h_e)^2.
\]
When \(s\ge\log(K-1)\), we have \(q_K(s)\le1/2\). Hence
\[
h^\top\Cov(p^{(c)}(s))h
\ge
\frac{q_K(s)}{2(K-1)}
\sum_{d\neq c}(h_d-h_c)^2,
\]
and also
\[
h^\top\Cov(p^{(c)}(s))h
\le
\frac{2q_K(s)}{K-1}
\sum_{d\neq c}(h_d-h_c)^2.
\]
Combining these inequalities with
\[
\frac{K-1}{2}e^{-s}\le q_K(s)\le (K-1)e^{-s}
\]
gives
\[
\frac14 e^{-s}
\sum_{d\neq c}(h_d-h_c)^2
\le
h^\top\Cov(p^{(c)}(s))h
\le
2e^{-s}
\sum_{d\neq c}(h_d-h_c)^2.
\]
Substituting
\[
h_d=\langle z_d,\vx_{c,i}\rangle,
\qquad
s=\tilde a_i\theta^\parallel,
\]
and averaging over \(c\) and \(i\), we obtain the operator inequalities
\[
\frac14\sum_{i=1}^N e^{-\tilde a_i\theta^\parallel}\mathsf B_i
\preceq
H(\theta^\parallel)
\preceq
2\sum_{i=1}^N e^{-\tilde a_i\theta^\parallel}\mathsf B_i.
\]
The lower bound restricted to \(i\in I_*\) gives
\[
H(\theta^\parallel)
\succeq
\frac14 e^{-\tilde a_{\min}\theta^\parallel}\mathsf B_*,
\]
and the upper bound gives
\[
H(\theta^\parallel)
\preceq
2 e^{-\tilde a_{\min}\theta^\parallel}\mathsf B_{\mathrm{all}}.
\]
Taking top eigenvalues yields \eqref{eq:ce_lambda1minus} and
\eqref{eq:ce_lambda1plus}.

\smallskip\noindent
\emph{Tail conditioning.}
If \(\mathsf B_*|_{\mathcal U}\succ0\), then
\[
H(\theta^\parallel)
\succeq
\frac14 e^{-\tilde a_{\min}\theta^\parallel}
\lambda_{\min}(\mathsf B_*|_{\mathcal U})I_{\mathcal U}.
\]
Meanwhile,
\[
\lambda_1(\theta^\parallel)
\le
2 e^{-\tilde a_{\min}\theta^\parallel}
\lambda_{\max}(\mathsf B_{\mathrm{all}}|_{\mathcal U}).
\]
Combining the two inequalities gives
\[
H(\theta^\parallel)
\succeq
\frac{
\lambda_{\min}(\mathsf B_*|_{\mathcal U})
}{
8\lambda_{\max}(\mathsf B_{\mathrm{all}}|_{\mathcal U})
}
\lambda_1(\theta^\parallel)I_{\mathcal U},
\]
which is \eqref{eq:ce_rhotail}. This proves the proposition.
\end{proof}

\subsection{Transient Implicit Bias for Multiclass Cross-Entropy}

\begin{theorem}[Spanning versus non-spanning geometry for multiclass cross-entropy]
\label{thm:ce_spanning_nonspanning}
Consider the multiclass cross-entropy model of
Appendix~\ref{app:cross_entropy}. Let
\[
\tilde a_i:=\frac{\kappa b_i}{\sqrt K},
\qquad
\kappa:=\frac{K}{K-1},
\]
and let
\[
\tilde\alpha_1<\tilde\alpha_2<\cdots<\tilde\alpha_L
\]
be the distinct values among \(\tilde a_1,\dots,\tilde a_N\). Define
\[
I_j:=\{i:\tilde a_i=\tilde\alpha_j\},
\qquad
\mathsf B_j:=\sum_{i\in I_j}\mathsf B_i,
\qquad
\mathsf B_{\le j}:=\sum_{k=1}^j \mathsf B_k,
\]
where \(\mathsf B_i:\mathcal U\to\mathcal U\) is the positive
semidefinite operator defined in~\eqref{eq:ce_Bi_def}. Assume the active
nondegeneracy condition
\[
\mathsf B_{\le L}|_{\mathcal U}\succ0.
\]
Define the first spanning layer index
\[
j_*^{\mathrm{CE}}
:=
\min\{j:\mathsf B_{\le j}|_{\mathcal U}\succ0\}.
\]
Then, for all sufficiently large \(\theta^\parallel\),
\[
\lambda_{\max}\!\left(H(\theta^\parallel)|_{\mathcal U}\right)
\asymp
e^{-\tilde\alpha_1\theta^\parallel},
\]
and
\[
\lambda_{\min}\!\left(H(\theta^\parallel)|_{\mathcal U}\right)
\asymp
e^{-\tilde\alpha_{j_*^{\mathrm{CE}}}\theta^\parallel}.
\]
Consequently, the tail conditioning ratio
\[
\rho_{\mathrm{CE}}(\theta^\parallel)
:=
\frac{
\lambda_{\min}(H(\theta^\parallel)|_{\mathcal U})
}{
\lambda_{\max}(H(\theta^\parallel)|_{\mathcal U})
}
\]
satisfies
\[
\rho_{\mathrm{CE}}(\theta^\parallel)
\asymp
e^{-(\tilde\alpha_{j_*^{\mathrm{CE}}}-\tilde\alpha_1)\theta^\parallel}.
\]

Along the stable-tail GD trajectory, since
\[
e^{\tilde\alpha_1(\theta_t^\parallel-\theta_*^{(r)})}
\asymp t,
\]
one has
\[
\rho_{\mathrm{CE}}(\theta_t^\parallel)
\asymp
t^{-(\tilde\alpha_{j_*^{\mathrm{CE}}}-\tilde\alpha_1)/\tilde\alpha_1}.
\]

In particular:
\begin{enumerate}[label=\textup{(\roman*)},leftmargin=2.2em]
\item If \(j_*^{\mathrm{CE}}=1\), equivalently
\[
\mathsf B_1|_{\mathcal U}=\mathsf B_*|_{\mathcal U}\succ0,
\]
then \(\rho_{\mathrm{CE}}(\theta^\parallel)\asymp1\), so the tail
conditioning hypothesis of Theorem~\ref{thm:stable_phase} holds and the
full three-phase contraction theorem applies.

\item If \(j_*^{\mathrm{CE}}>1\), then
\[
\rho_{\mathrm{CE}}(\theta_t^\parallel)
\asymp
t^{-(\tilde\alpha_{j_*^{\mathrm{CE}}}-\tilde\alpha_1)/\tilde\alpha_1}
\to0.
\]
Thus the top sharpness still decays on the \(t^{-1}\) scale, but it no
longer controls all active orthogonal directions uniformly.
\end{enumerate}
\end{theorem}

\begin{proof}
By the tail sharpness estimates in Proposition~\ref{prop:ce_verify_assumptions},
for all sufficiently large \(\theta^\parallel\) there exist constants
\(0<c_-\le c_+<\infty\) such that, in Loewner order on \(\mathcal U\),
\[
c_-\sum_{j=1}^L e^{-\tilde\alpha_j\theta^\parallel}\mathsf B_j
\preceq
H(\theta^\parallel)
\preceq
c_+\sum_{j=1}^L e^{-\tilde\alpha_j\theta^\parallel}\mathsf B_j.
\]
For the top eigenvalue, the upper bound follows from
\[
H(\theta^\parallel)
\preceq
c_+e^{-\tilde\alpha_1\theta^\parallel}
\sum_{j=1}^L\mathsf B_j,
\]
so
\[
\lambda_{\max}(H(\theta^\parallel)|_{\mathcal U})
\lesssim
e^{-\tilde\alpha_1\theta^\parallel}.
\]
For the lower bound, since \(\mathsf B_1\neq0\) on the active directions
seen by the slowest layer,
\[
H(\theta^\parallel)
\succeq
c_-e^{-\tilde\alpha_1\theta^\parallel}\mathsf B_1,
\]
and hence
\[
\lambda_{\max}(H(\theta^\parallel)|_{\mathcal U})
\gtrsim
e^{-\tilde\alpha_1\theta^\parallel}.
\]
Thus
\[
\lambda_{\max}(H(\theta^\parallel)|_{\mathcal U})
\asymp
e^{-\tilde\alpha_1\theta^\parallel}.
\]

For the smallest eigenvalue, let
\[
j_*^{\mathrm{CE}}
=
\min\{j:\mathsf B_{\le j}|_{\mathcal U}\succ0\}.
\]
Since \(\mathsf B_{\le j_*^{\mathrm{CE}}}|_{\mathcal U}\succ0\),
we have
\[
\lambda_{\min}(\mathsf B_{\le j_*^{\mathrm{CE}}}|_{\mathcal U})>0.
\]
Using \(\tilde\alpha_k\le \tilde\alpha_{j_*^{\mathrm{CE}}}\) for
\(k\le j_*^{\mathrm{CE}}\), the lower Loewner bound gives
\[
H(\theta^\parallel)
\succeq
c_- e^{-\tilde\alpha_{j_*^{\mathrm{CE}}}\theta^\parallel}
\mathsf B_{\le j_*^{\mathrm{CE}}},
\]
and therefore
\[
\lambda_{\min}(H(\theta^\parallel)|_{\mathcal U})
\gtrsim
e^{-\tilde\alpha_{j_*^{\mathrm{CE}}}\theta^\parallel}.
\]

For the matching upper bound, if \(j_*^{\mathrm{CE}}>1\), minimality of
\(j_*^{\mathrm{CE}}\) implies that
\(\mathsf B_{\le j_*^{\mathrm{CE}}-1}|_{\mathcal U}\) is not positive
definite. Hence there exists a unit vector
\[
Z\in\mathcal U
\quad\text{such that}\quad
\langle Z,\mathsf B_{\le j_*^{\mathrm{CE}}-1}Z\rangle_F=0.
\]
Since each \(\mathsf B_j\succeq0\), this implies
\[
\langle Z,\mathsf B_j Z\rangle_F=0
\qquad
\forall j<j_*^{\mathrm{CE}}.
\]
Therefore
\[
\langle Z,H(\theta^\parallel)Z\rangle_F
\le
c_+\sum_{j=j_*^{\mathrm{CE}}}^L
e^{-\tilde\alpha_j\theta^\parallel}
\langle Z,\mathsf B_j Z\rangle_F
\lesssim
e^{-\tilde\alpha_{j_*^{\mathrm{CE}}}\theta^\parallel}.
\]
Taking the minimum over unit \(Z\in\mathcal U\) gives
\[
\lambda_{\min}(H(\theta^\parallel)|_{\mathcal U})
\lesssim
e^{-\tilde\alpha_{j_*^{\mathrm{CE}}}\theta^\parallel}.
\]
The case \(j_*^{\mathrm{CE}}=1\) follows from
\[
\lambda_{\min}(H)\le\lambda_{\max}(H)
\asymp e^{-\tilde\alpha_1\theta^\parallel}
\]
together with the lower bound already proved.

Dividing the smallest-eigenvalue estimate by the top-eigenvalue estimate
gives
\[
\rho_{\mathrm{CE}}(\theta^\parallel)
\asymp
e^{-(\tilde\alpha_{j_*^{\mathrm{CE}}}-\tilde\alpha_1)\theta^\parallel}.
\]
Finally, the stable-tail schedule gives
\[
e^{\tilde\alpha_1(\theta_t^\parallel-\theta_*^{(r)})}\asymp t,
\]
so
\[
e^{-(\tilde\alpha_{j_*^{\mathrm{CE}}}-\tilde\alpha_1)\theta_t^\parallel}
\asymp
t^{-(\tilde\alpha_{j_*^{\mathrm{CE}}}-\tilde\alpha_1)/\tilde\alpha_1}.
\]
This proves the theorem.
\end{proof}
\newpage

\section{Experimental Details}\label{app:additional_experiments}

\myparagraph{Compute resources}
All experiments are low-dimensional synthetic simulations run with full-batch
gradient descent. They require only standard CPU computation, no GPU or
specialized accelerator, and use negligible memory; each run completes in
under a few seconds on a typical laptop.

\subsection{Experimental Setup}\label{app:experiment_setup}

\myparagraph{Spanning experiment (Section~\ref{sec:experiments},
Figure~\ref{fig:rate_verification})}
We use pair-symmetric logistic regression in \(\RR^3\), with
\(\vv=e_1\) and active orthogonal space
\(U=\Span\{e_2,e_3\}\).   The dataset has \(N=24\) reflected pairs.  The slowest layer has
\(N_{\rm slow}=10\) pairs with \(a_i=a_{\min}=0.20\), and the faster
layers have \(N_{\rm fast}=14\) pairs with
\[
  a_i\in\{0.35,0.45,0.55\}.
\]
The slowest layer is constructed to span \(U\): two slow vectors are fixed
as
\[
  u_1=(1.8,0),
  \qquad
  u_2=(0,0.5),
\]
and the remaining slow vectors, together with the faster-layer vectors,
are sampled using random seed \(1\).  Specifically, their directions are
uniform on \([0,2\pi]\), their radii are uniform on \([0.5,1.8]\), and
the faster-layer \(a_i\)'s are sampled uniformly from
\(\{0.35,0.45,0.55\}\).

We use step size \(\eta=15\), initialize at
\[
  \theta_0^\parallel=0,
  \qquad
  \theta_0^\perp=(0.01,0.006),
\]
and run full-batch gradient descent for \(T=100\) iterations.  The
prescribed tubular neighborhood radius is \(r_{\rm tubular neighborhood}=0.15\).  The RC tolerance is
computed as
\[
  \delta_r=\cosh(U_{\max}r_{\rm tubular neighborhood})-1,
  \qquad
  U_{\max}:=\max_i\|u_i\|,
\]
and the tail-conditioning constant is
\[
  \rho_{\rm tail}
  =
  \frac{\lambda_{\min}(\Sigma_{\rm slow})}
       {\lambda_{\max}(\Sigma_U)},
  \qquad
  \Sigma_U=\frac1N\sum_{i=1}^N u_i u_i^\top,
  \qquad
  \Sigma_{\rm slow}
  =
  \frac1N\sum_{i:a_i=a_{\min}}u_i u_i^\top.
\]
For this instance,
\(\delta_r=0.037\), \(\rho_{\rm tail}=0.200\),
\(\varepsilon_*=0.186\), and the observed maximum tubular neighborhood radius is
\(r_{\rm obs}=0.13<r_{\rm tubular neighborhood}\).  The phase boundaries are computed
from the corrected stability product
\[
  \eta(1+\delta_r)\lambda_1(\theta_t^\parallel),
\]
with \(t_{\rm stab}\) the first time this product is at most \(2\),
\(t_{\rm lin}\) the first time it is at most \(2-\varepsilon_*\), and
\(t_{\rm poly}\) the first time it is at most \(1\).

\myparagraph{Non-spanning experiment (Section~\ref{sec:experiments},
Figure~\ref{fig:nonspanning})}
We use pair-symmetric logistic regression in \(\RR^3\), with signed-feature
coordinates \(z=yx\) and empirical loss
\[
  \Loss(\bth)
  =
  \frac1M\sum_{j=1}^M
  \log\!\left(1+\exp(-\langle \bth,z_j\rangle)\right).
\]
The max-margin direction is \(v=e_1\), and the active orthogonal space is
\(U=\Span\{e_2,e_3\}\).  The training set consists of reflected
signed-feature pairs \(z_{i,\pm}=a_i e_1\pm u_i\).  We use \(5\) support
pairs with \(a_i=\alpha_1=0.1\) and \(u_i=r_i e_2\), and \(5\) weak-layer
pairs with \(a_i=\alpha_2=1.0\) and \(u_i=r_i e_3\), for \(M=20\) signed
features in total.  The radii are generated with random seed \(3\) by
\[
  r_i = 2(0.5+0.5\,\xi_i),
  \qquad
  \xi_i\sim\mathrm{Unif}[0,1],
\]
independently for each reflected pair.  Thus the orthogonal components of the support vectors span only the \(e_2\)-direction, while the weak \(e_3\)-direction appears only in the
faster-decaying layer.

We initialize at
\[
  \bth_0=(0.8,\,0.20,\,0.05)
\]
and run full-batch gradient descent for \(T=150\) iterations.  The baseline
step size is
\[
  \eta_c
  :=
  \frac{2}{\lambda_{\max}(\nabla^2\Loss(\bth_0)|_U)},
\]
where the Hessian is restricted to the active orthogonal coordinates
\((e_2,e_3)\).  We sweep
\[
  \eta/\eta_c\in\{1,1.5,2,2.5,3\}.
\]
At each time \(t\), we record the loss, the parallel coordinate
\(\theta_t^\parallel=\langle\bth_t,e_1\rangle\), the support-vector
component \(c_{\mathrm{sv},t}=\langle\bth_t,e_2\rangle\), and the weak
component \(c_{\mathrm{wk},t}=\langle\bth_t,e_3\rangle\).  The restricted
Hessian is evaluated at \((\theta_t^\parallel,0,0)\), and the conditioning
ratio is computed as
\[
  \rho_t
  =
  \frac{\lambda_{\min}(H(\theta_t^\parallel)|_U)}
       {\lambda_{\max}(H(\theta_t^\parallel)|_U)}.
\]
The angle to the max-margin line is
\[
  \arctan\!\left(
  \frac{\sqrt{c_{\mathrm{sv},t}^2+c_{\mathrm{wk},t}^2}}
       {|\theta_t^\parallel|}
  \right),
\]
reported in degrees.

\subsection{Geometry-Dependent Step-Size Effects in Non-Spanning Configurations}
\label{app:additional_nonspanning}

\begin{figure}
    \centering
    \includegraphics[width=0.95\linewidth]{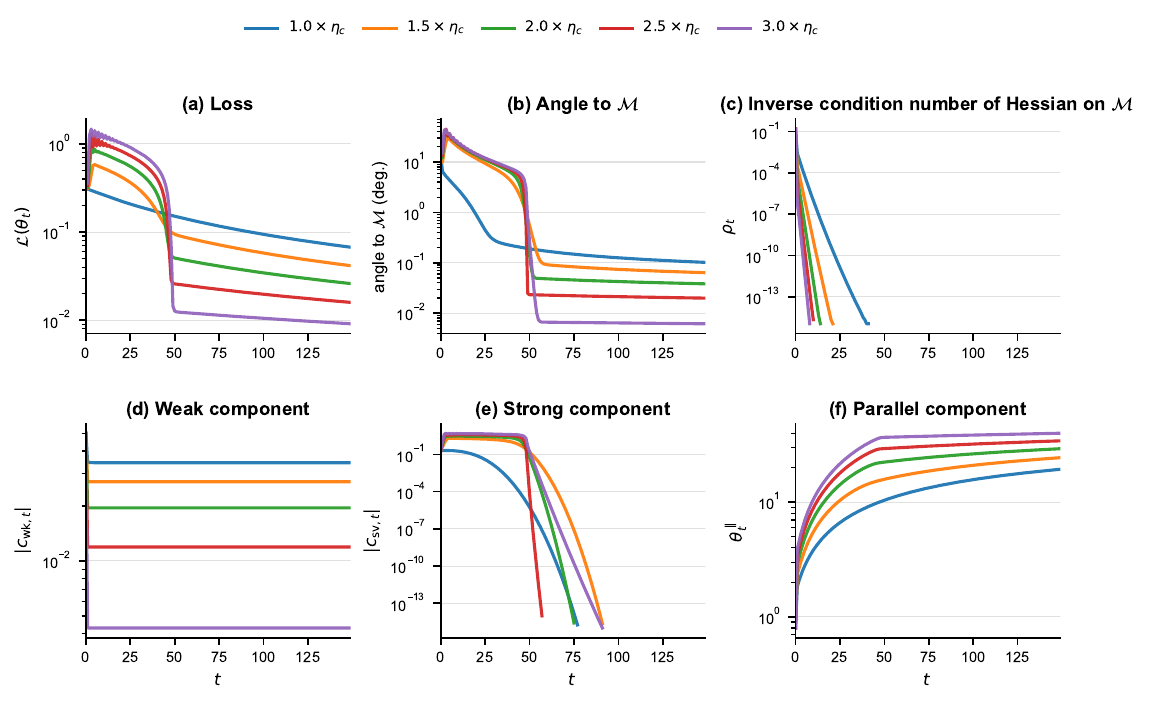}
    \caption{Non-spanning configuration with $\alpha_2=4$.}
    \label{fig:nonspanning_suppression}
\end{figure}

\begin{figure}
    \centering
    \includegraphics[width=0.95\linewidth]{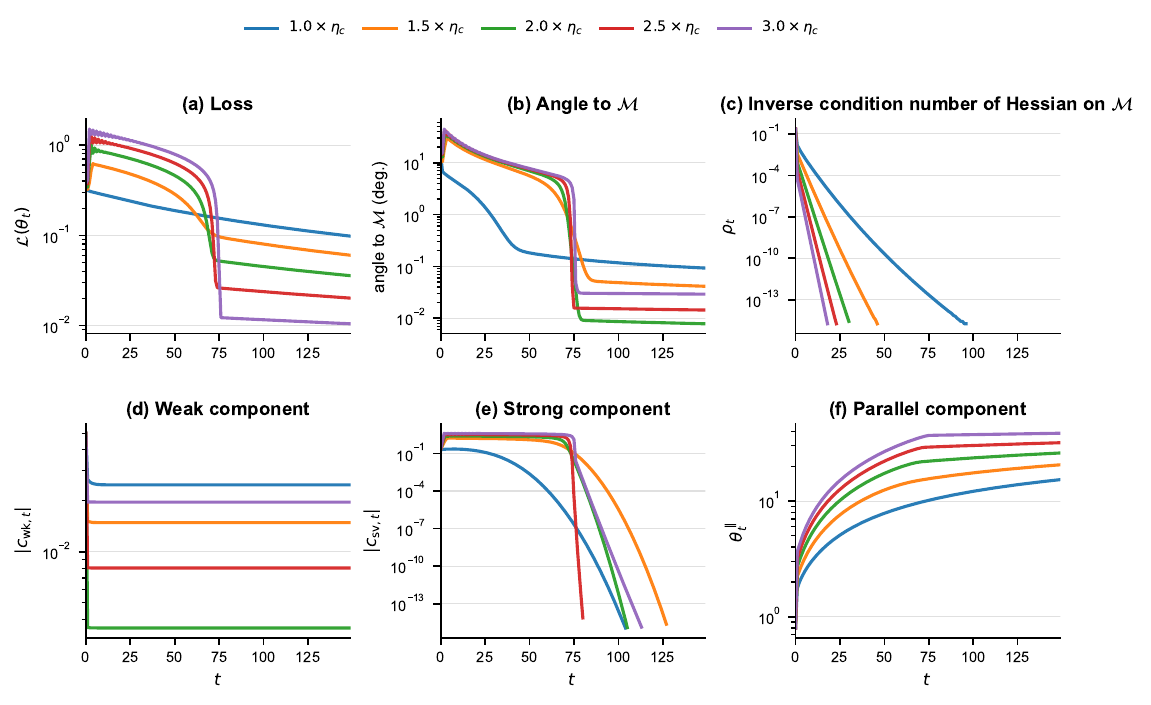}
    \caption{Non-spanning configuration with $\alpha_2=3$.}
    \label{fig:nonspanning_nonmonotone}
\end{figure}

Section~\ref{sec:experiments} illustrates one regime of step-size behavior in non-spanning geometries: larger $\eta$ amplifies the persistent orthogonal  residual measured at a finite horizon $T$. This appendix complements that experiment with two additional regimes, both arising in non-spanning  configurations: residual \emph{suppression} by larger $\eta$ (Figure~\ref{fig:nonspanning_suppression}) and \emph{non-monotone} dependence with an interior minimum (Figure~\ref{fig:nonspanning_nonmonotone}). Together with the main-text experiment, these three panels demonstrate that the step-size dependence of the residual is governed by the data geometry rather than by a single universal law. All three experiments use the same pair-symmetric logistic regression  template as in Section~\ref{sec:experiments}: $d = 3$, $N = 20$ data  points, and identical initialization protocol. The configurations differ  only in the values of $\alpha_2$. Specific parameter values are listed in the figure captions. Across all panels we sweep five step sizes spanning the unstable range; the smallest is in the classically stable regime and the  largest is well above $2/\lambda_{\max}(\bth_0)$.

\subsection{Downstream effect of the orthogonal residual}\label{app:robustness_experiment}

\begin{figure}
    \centering
    \includegraphics[width=\linewidth]{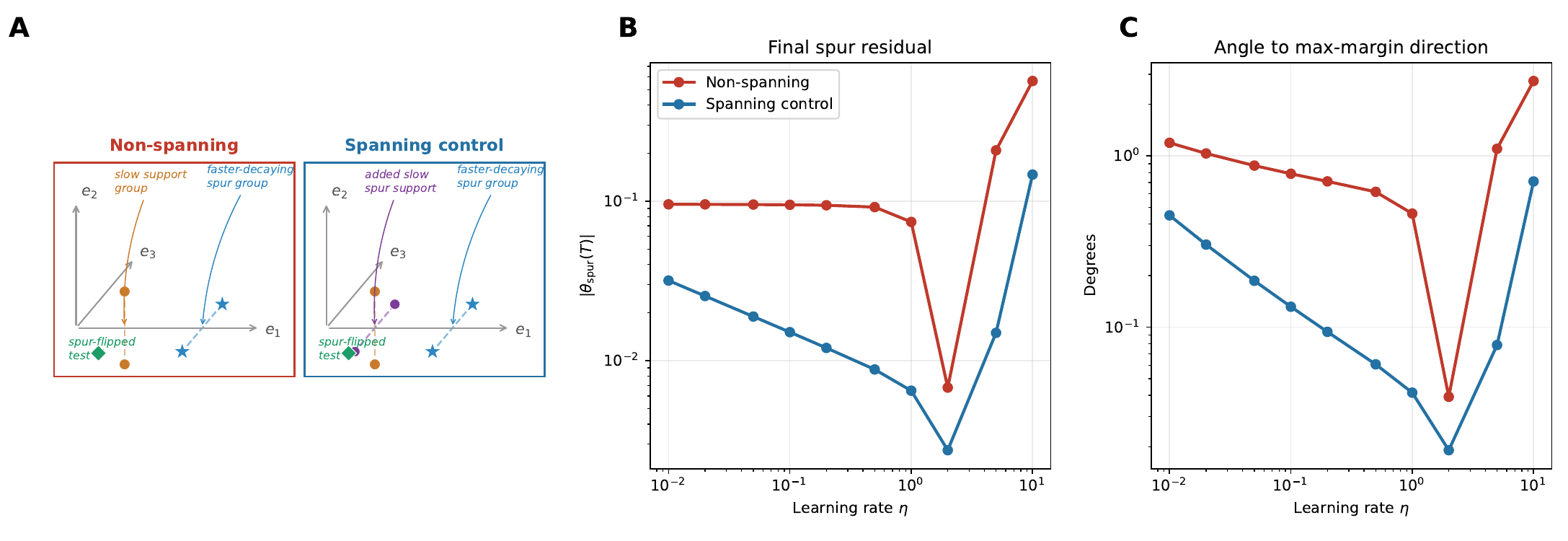}
    \caption{Non-spanning support geometry produces learning-rate-dependent
  residuals in pair-symmetric logistic regression ($\RR^3$,
  $T = 20{,}000$ steps).
  \textbf{(A)}~Signed-feature schematics. Reflected pairs
  $z_{i,\pm} = a_i e_1 \pm u_i$ are connected by dashed lines.
  The slow support group (orange, $a = 1$) spans only~$e_2$;
  the fast spur group (blue, $a = 4$) spans~$e_3$.
  In the spanning control, an additional slow group (purple)
  adds support in~$e_3$.
  Both geometries share the same max-margin line~$e_1$.
  \textbf{(B)}~Final spur residual
  $|\langle \bth_T, e_3 \rangle|$ versus learning rate.
  The non-spanning geometry retains a systematically larger
  residual, with the gap widening at large~$\eta$ where
  unstable amplification is strongest.
  \textbf{(C)}~Angle between $\bth_T$ and the max-margin
  direction.  The non-spanning residual translates into a
  larger finite-time deviation from the asymptotic classifier,
  confirming that top-eigenvalue stabilization alone does not
  guarantee uniform contraction of all orthogonal components.}
    \label{fig:robustness}
\end{figure}

The non-spanning experiments above show that weakly controlled orthogonal components can persist at any finite horizon. To illustrate that this residual can have a downstream effect, we construct a simple distribution-shift test on the same non-spanning logistic geometry used in Section~\ref{sec:experiments}. We use the pair-symmetric model in $\RR^3$ with support vectors spanning only $\ve_2$ (margin scale $\alpha_1$) and non-support vectors along $\ve_3$ (margin scale $\alpha_2 > \alpha_1$), so that $\ve_3$ is the weakly controlled ``spur'' direction. An informal analogy: $\ve_1$ encodes a robust class feature (e.g., animal shape in a cats-vs-dogs classifier), while $\ve_3$ encodes a spurious cue (e.g., background texture) that correlates with the label in training but not at test time. We evaluate the trained classifier on a shifted test distribution in which the sign of the $\ve_3$ component is flipped relative to training: a classifier with spur residual $\theta_{\mathrm{spur}} = \langle \bth_T, \ve_3 \rangle$ sees its effective margin reduced by $2|\theta_{\mathrm{spur}}|$ on this shifted distribution. Figure~\ref{fig:robustness} shows that the non-spanning geometry leaves a larger spur residual and a larger finite-time angle to the max-margin direction, while the spanning control suppresses this residual. This illustrates, in a controlled low-dimensional example, how a finite-time spur residual can affect downstream test margins under a shift that reverses the spur coordinate. We emphasize that this is an illustrative low-dimensional example; understanding the scope of this phenomenon in more realistic settings is left to future work.

\paragraph{Implementation details.}
We use pair-symmetric logistic regression in \(\RR^3\), with
\(\vv=e_1\) as the max-margin direction and
\(U=\Span\{e_2,e_3\}\) as the active orthogonal space.  We work in
signed-feature coordinates \(z=yx\), so the logistic loss is
\[
  \Loss(\bth)
  =
  \frac1M\sum_{j=1}^M \phi(\langle \bth,z_j\rangle),
  \qquad
  \phi(s)=\log(1+e^{-s}).
\]
The training set consists of reflected signed-feature pairs
\(z_{i,\pm}=a_i e_1\pm u_i\).  In the non-spanning geometry, the slow
support group has \(a_1=1\) and \(u_1=e_2\), with \(40\) reflected pairs
(\(80\) signed features), while the fast spur group has \(a_2=4\) and
\(u_2=2e_3\), also with \(40\) reflected pairs (\(80\) signed features).
Thus the non-spanning training set has \(M=160\) signed features.  The
spanning control appends a third group at the slow decay rate \(a_3=1\)
with \(u_3=0.8e_3\), using \(40\) reflected pairs, bringing the total to
\(M=240\) signed features.

In both geometries, the max-margin direction is \(e_1\) by symmetry.  The
two settings differ only in whether the slowest-decay support group spans
the spur direction \(e_3\).  We initialize at
\[
  \bth_0=(0.1,\,0.15,\,0.15)
\]
and run full-batch gradient descent for \(T=20{,}000\) steps at each of
ten learning rates
\[
  \eta\in\{0.01,0.02,0.05,0.1,0.2,0.5,1,2,5,10\}.
\]

For the downstream test, we evaluate two near-boundary signed-feature
clouds.  The aligned cloud consists of \(600\) signed features sampled
from a Gaussian with mean \((1,0,+2)\) and coordinatewise standard
deviation \(0.02\).  The shifted cloud is sampled analogously with mean
\((1,0,-2)\), so its spur coordinate has the opposite sign.  We report
the final spur residual
\[
  |\theta_{\mathrm{spur}}(T)|
  =
  |\langle \bth_T,e_3\rangle|,
\]
the angle to the max-margin line
\[
  \arctan\!\left(
    \frac{\|\Pi_U\bth_T\|}{|\langle \bth_T,e_1\rangle|}
  \right),
\]
and the empirical logistic loss on the aligned and shifted test clouds.

\newpage

\section{Approximate Symmetry and Extensions}\label{app:approximate_symmetry}

\begin{figure}
    \centering
    \includegraphics[width=0.9\linewidth]{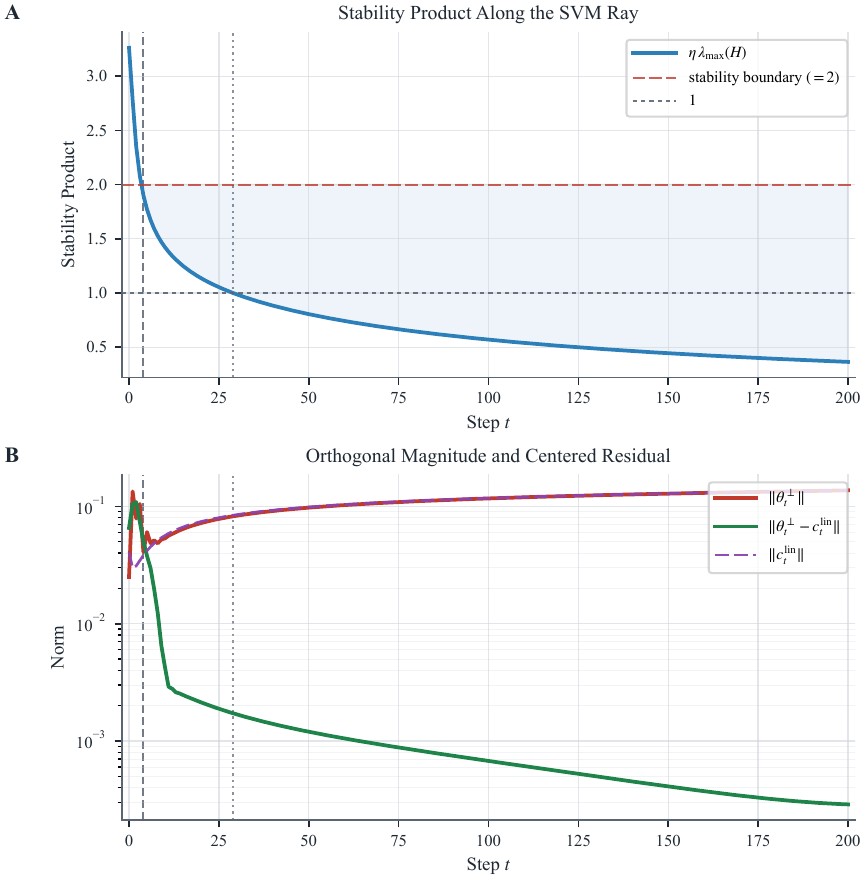}
    \caption{Small asymmetric perturbation of the pair-symmetric logistic model.}
    \label{fig:asymmetric_sanity}
\end{figure}

Exact pair-symmetry yields the clean multiplicative recursion $\tho_{t+1}=(I-\eta\widetilde H_t)\tho_t$ by making the max-margin direction coincide with the subspace $\Span\{\vv\}$ and freezing the inactive subspace.  For a general separable dataset, one instead decomposes along the hard-margin SVM
direction $\vv_{\mathrm{svm}}$, producing a perturbed recursion
\[
  \tho_{t+1}
  =
  \bigl(I-\eta H(\thp_t)\bigr)\tho_t
  +\mathrm{Err}_t,
\]
where $\mathrm{Err}_t$ measures the failure of exact
block-alignment, controlled by the off-diagonal quantities
$\|\Pi_{\vv_{\mathrm{svm}}^\perp}
    \nabla\Loss(\thp\vv_{\mathrm{svm}})\|$
and
$\|\Pi_{\vv_{\mathrm{svm}}}
    \nabla^2\Loss(\thp\vv_{\mathrm{svm}})
    \Pi_{\vv_{\mathrm{svm}}^\perp}\|$.
If these remain small on the relevant tubular neighborhood, the same analytical strategy applies, and the orthogonal component converges to an
asymmetry-dependent residual rather than to zero.
The symmetric setting is thus the exactly aligned limit of a
broader perturbative theory whose development we leave to
future work. To test whether this mechanism
is brittle, we also consider a small asymmetric perturbation of the
pair-symmetric logistic construction.  We compute the hard-margin SVM
direction \(v_{\rm svm}\) of the perturbed dataset and decompose
\[
    \theta_t=\theta_t^\parallel v_{\rm svm}+\theta_t^\perp,
    \qquad
    \theta_t^\perp\perp v_{\rm svm}.
\]
In this setting the SVM line is no longer exactly invariant.  Therefore
the orthogonal component need not contract to zero.  Instead, the
linearized orthogonal dynamics have the instantaneous center
\[
    c_t^{\rm lin}
    =
    -H_\perp(\theta_t^\parallel)^{-1}
    \Pi_{v_{\rm svm}^\perp}
    \nabla L(\theta_t^\parallel v_{\rm svm}),
\]
where \(H_\perp(\theta^\parallel)\) is the Hessian restricted to
\(v_{\rm svm}^\perp\).  Figure~\ref{fig:asymmetric_sanity} shows that
the stability product still self-stabilizes, while the centered residual
\(\|\theta_t^\perp-c_t^{\rm lin}\|\) contracts.  Thus the exact symmetric
analysis appears to describe the aligned limit of a perturbed recursion,
where stability controls deviations from a moving center rather than the
raw orthogonal norm itself.  This experiment is illustrative and is not
covered by the theorem.

\paragraph{Experimental setup.}
We study a controlled synthetic aligned-feature logistic-regression problem in three dimensions, where each example is represented in aligned coordinates as $Z_i = y_i x_i \in \mathbb{R}^3$. The base dataset is pair-symmetric around the $e_1$ direction. It contains a slow layer of six anchor points with parallel coordinate $a_{\mathrm{slow}}=0.08$ and transverse coordinates $(\pm1,0)$, $(0,\pm1)$, and $(\pm0.7,\pm0.7)$, together with a faster layer of eight reflected pairs with parallel coordinate $a_{\mathrm{fast}}=0.28$. For each fast pair, the orthogonal direction is sampled from a normalized Gaussian draw and its radius is sampled uniformly from $[0.25,1.10]$. This produces a dataset of $N=22$ aligned examples.

To move away from the exactly symmetric regime, we introduce a controlled asymmetry parameter $\varepsilon_{\mathrm{asym}}=0.04$. Specifically, we perturb both the parallel coordinate and the transverse coordinates of every example by Gaussian noise scaled by $\varepsilon_{\mathrm{asym}}$, while enforcing that all parallel coordinates remain positive so that the dataset stays separable. We then compute an approximate hard-margin direction $v_{\mathrm{svm}}$ directly in aligned coordinates and decompose the iterate as
\[
\theta_t = \theta_t^\parallel v_{\mathrm{svm}} + B\theta_t^\perp,
\]
where the columns of $B$ form an orthonormal basis of $v_{\mathrm{svm}}^\perp$.

We optimize the mean logistic loss with full-batch gradient descent using a fixed step size $\eta=35$ for $T=100$ iterations. Initialization is set to zero in the parallel direction and to a random orthogonal component of norm $0.025$. Unless otherwise stated, all experiments use random seed $3$.

\end{document}